\documentclass[a4paper,11pt]{article}
\usepackage[paper=a4paper,left=25mm,right=25mm,top=25mm,bottom=25mm]{geometry} 

 \usepackage{setspace}
\usepackage{authblk}

\usepackage{fullpage}
\usepackage{parskip}
\usepackage{titlesec}
\usepackage[section]{placeins}
\usepackage[dvipsnames]{xcolor}

\usepackage{breakcites}
\usepackage{lineno}
\usepackage{hyphenat}
\usepackage[all]{nowidow}

\PassOptionsToPackage{hyphens}{url}
\usepackage[colorlinks = true]{hyperref} 

\hypersetup{
  linkcolor  = RedViolet,
  citecolor = Blue!90!black,
  urlcolor   = Aquamarine!80!black,
  colorlinks = true
}

\usepackage{etoolbox}
\usepackage[utf8]{inputenc}
\usepackage[natbib=true, backend=biber, sorting=nyt, style=apa, apamaxprtauth=20]{biblatex}
\renewenvironment{abstract}
  {{\bfseries\noindent{\abstractname}\par\nobreak}\footnotesize}
  {\bigskip}
\titlespacing{\section}{0pt}{*3}{*1}
\titlespacing{\subsection}{0pt}{*2}{*0.5}
\titlespacing{\subsubsection}{0pt}{*1.5}{0pt}

\usepackage{graphicx}
\usepackage[space]{grffile}
\usepackage{latexsym}
\usepackage{textcomp}
\usepackage{longtable}
\usepackage{tabularx}
\usepackage{booktabs,array,multirow}
\usepackage{amsfonts,amsmath,amssymb}
\usepackage{subcaption}
\providecommand\citet{\cite}
\providecommand\citep{\cite}

\newif\iflatexml\latexmlfalse
\usepackage{rotating}
\usepackage{color}
\usepackage{tabularray}

\AtBeginDocument{\DeclareGraphicsExtensions{.pdf,.PDF,.eps,.EPS,.png,.PNG,.tif,.TIF,.jpg,.JPG,.jpeg,.JPEG}}

\usepackage[english]{babel}
\usepackage{float}

\usepackage{orcidlink}
\usepackage{csquotes}
\newcommand{\beginsupplement}{
        \setcounter{table}{0}
        \renewcommand{\thetable}{S\arabic{table}}
        \setcounter{figure}{0}
        \renewcommand{\thefigure}{S\arabic{figure}}
     }

\newcommand{\bs}[1]{\boldsymbol{#1}} 
\newcommand{\pdistr}{\mathds{P}_{xy}} 
\newcommand{\lp}{\mathcal{I}} 
\newcommand{\lpi}{\mathcal{I}_{\hpi}} 
\newcommand{\lpii}{\mathcal{I}_{\hpii}} 
\newcommand{\ds}{\mathcal{D}} 
\newcommand{\dtrain}{\mathcal{D}_{\text{train}}} 
\newcommand{\dtraine}{\mathcal{D}^\prime_{\text{train}}} 
\newcommand{\ntrain}{n^{}_{\text{train}}} 
\newcommand{\ntraine}{n^{\prime}_{\text{train}}} 
\newcommand{\ntest}{n^{}_{\text{test}}} 
\newcommand{\dtest}{\mathcal{D}_{\text{test}}}
\newcommand{\dtrainee}{\mathcal{D}^{\prime\prime}_{\text{train}}} 
\newcommand{\dteste}{\mathcal{D}^\prime_{\text{test}}} 
\newcommand{\fdltrain}{\hat{f}^{\dtrain}_{\mathcal{I}}}
\newcommand{\fdlitrain}{\hat{f}^{\dtrain}_{\mathcal{I}_{\hpi}}}
\newcommand{\fdlitraine}{\hat{f}^{\dtraine}_{\mathcal{I}_{\hpi}}}
\newcommand{\fdlhptrain}{\hat{f}^{\dtrain}_{\mathcal{I}_{\bs{\lambda}}}}
\newcommand{\fdliitrain}{\hat{f}^{\dtrain}_{\mathcal{I}_{\hpii}}}
\newcommand{\fdliitraine}{\hat{f}^{\dtraine}_{\mathcal{I}_{\hpii}}}
\newcommand{\thetadl}{\bs{\hat{\theta}}^{\dtrain}_{\lp}} 
\newcommand{\hp}{\bs{\lambda}}
\newcommand{\hpp}{\bs{\lambda}_P}  % preprocesssing HPs
\newcommand{\hpa}{\bs{\lambda}_A}  % algorithm HPs
\newcommand{\hpindsmall}{j} 
\newcommand{\hpindlarge}{J} 
\newcommand{\hpji}{\lambda_j^\mathrm{I}}
\newcommand{\hpjii}{\lambda_j^\mathrm{II}}
\newcommand{\hpi}{\bs{\lambda}^\mathrm{I}}
\newcommand{\hpii}{\bs{\lambda}^\mathrm{II}}
\newcommand{\hpstar}{\bs{\lambda}^\ast}
\newcommand{\hpstartilde}{\bs{\tilde{\lambda}}^\ast} 
\newcommand{\searchspace}{\bs{\tilde{\Lambda}}}
\newcommand{\searchspacej}{\tilde{\Lambda}_\hpindsmall}
\newcommand{\hpaii}{\bs{\lambda}_A^\mathrm{II}}
\newcommand{\hppii}{\bs{\lambda}_P^\mathrm{II}}
\newcommand{\hpcand}{\hp^{(c)}}
\newcommand{\hpiie}{\bs{\lambda}^{\prime\mathrm{II}}}
\newcommand{\hpoutlier}{\lambda_{outlier}}
\newcommand{\hpca}{\lambda_{ca}}
\newcommand{\hpipos}{\lambda_{ipos}}
\newcommand{\hpage}{\lambda_{age}}
\newcommand{\hpakps}{\lambda_{akps}}
\newcommand{\hpalpha}{\lambda_{\alpha}}
\newcommand{\hpcp}{\lambda_{cp}}
\newcommand{\hpminb}{\lambda_{minbucket}}
\newcommand{\pdefault}{I-no tuning}
\newcommand{\pM}{II-manual-P}
\newcommand{\pA}{II-automated-A}
\newcommand{\pC}{II-combined-PA}
\newcommand{\pAii}{II-automated-PA}
\newcommand{\dnew}{\mathcal{D_{\text{new}}}} \newcommand{\ped}{\widehat{\mathrm{PE}}_{\text{train}}}
\newcommand{\pednew}{\widehat{\mathrm{PE}}_{\text{new}}}
\usepackage{dsfont}
\usepackage{makecell}
\usepackage{mathtools}
\usepackage{enumitem}
\newlist{tabitemize}{itemize}{1}
\setlist[tabitemize]{label=\textbullet,leftmargin=*,topsep=1ex,
parsep=0pt,
                  after=\vspace{-\baselineskip},
                  before=\vspace{-0.75\baselineskip}
                  }  
\newcolumntype{L}[1]{>{\raggedright\arraybackslash}p{#1}}
\usepackage{eurosym}
\usepackage{siunitx}
\usepackage{colortbl}
\usepackage{hhline}
\begin{document}
\title{Beyond algorithm hyperparameters: on preprocessing hyperparameters and associated pitfalls in machine learning applications}

\def\correspondingauthor{\footnote{Corresponding author, e-mail: \href{mailto:theresa.ullmann@meduniwien.ac.at}{theresa.ullmann@meduniwien.ac.at}}}
\author[1,2]{Christina Sauer\orcidlink{0000-0003-2425-7858}}
\author[1,2]{Anne-Laure Boulesteix\orcidlink{0000-0002-2729-0947}} %\email{boulesteix@ibe.med.uni-muenchen.de}
\author[1]{Luzia Hanßum}%\email{iiauthor@gmail.com}
\author[3]{Farina Hodiamont}%\email{farina.hodiamont@med.uni-muenchen.de}
\author[3]{Claudia Bausewein}%\email{claudia.bausewein@med.uni-muenchen.de}
\author[4]{Theresa Ullmann\correspondingauthor{}\orcidlink{0000-0003-1215-8561}}%\email{theresa.ullmann@meduniwien.ac.at}

\affil[1]{Institute for Medical Information Processing, Biometry and Epidemiology, Faculty of Medicine, LMU Munich, Munich, Germany}
\affil[2]{Munich Center for Machine Learning (MCML), Munich, Germany}
\affil[3]{Department of Palliative Medicine, University Hospital, LMU Munich, Munich, Germany}
\affil[4]{Institute of Clinical Biometrics, Center for Medical Data Science, Medical University of Vienna, Vienna, Austria}

\vspace{-1em}
  \date{\today}
\maketitle

\begin{abstract}
Adequately generating and evaluating prediction models based on supervised machine learning (ML) is often challenging, especially for less experienced users in applied research areas. Special attention is required in settings where the model generation process involves hyperparameter tuning, i.e.\ data-driven optimization of different types of hyperparameters to improve the predictive performance of the resulting model. Discussions about tuning typically focus on the hyperparameters of the ML algorithm (e.g., the minimum number of observations in each terminal node for a tree-based algorithm). In this context, it is often neglected that hyperparameters also exist for the preprocessing steps that are applied to the data before it is provided to the algorithm (e.g., how to handle missing feature values in the data). As a consequence, users experimenting with different preprocessing options to improve model performance may be unaware that this constitutes a form of hyperparameter tuning, albeit informal and unsystematic, and thus may fail to report or account for this optimization. To illuminate this issue, this paper reviews and empirically illustrates different procedures for generating and evaluating prediction models, explicitly addressing the different ways algorithm and preprocessing hyperparameters are typically handled by applied ML users. By highlighting potential pitfalls, especially those that may lead to exaggerated performance claims, this review aims to further improve the quality of predictive modeling in ML applications.

\end{abstract}

\vspace{-1em}
\sloppy

\section{Introduction}
Many applied research areas have recently seen an increase in the development of prediction models based on supervised machine learning (ML) algorithms.
However, after initially generating widespread enthusiasm---partly due to the availability of user-friendly software that enables model development without requiring extensive expertise---ML-based prediction models are now undergoing critical reexamination \citep{Ball2023,Kapoor2023leakage,Pfob2022}. 
Among other concerns, such as insufficient reporting of relevant aspects of the model development process, it has been found that the claimed predictive performance of many models is considerably exaggerated \citep{Kapoor2023leakage,Dhiman2022_methconduct,Dhiman2022riskofbias,Navarro2021riskofbias}. 
While some of the pitfalls leading to such optimistically biased performance claims (e.g., using the exact same observations for model generation and evaluation) typically occur only among very inexperienced applied ML users and are well known within the ML research community,  others arise more subtly \citep{Domingos2012, Kapoor2023leakage,Poldrack2020,Hofman2023}.\\
This is particularly true when the model generation process involves data-driven hyperparameter optimization, which is also referred to as hyperparameter tuning and is commonly employed in ML applications.
The most prominent type of hyperparameters (HPs) are those associated with the learning algorithm, which specify its configuration (e.g., the minimum number of observations in each terminal node for tree-based algorithms). If selected by an adequate (and ideally automated) tuning procedure, HPs can substantially enhance the performance of the resulting prediction model. However, HP tuning also complicates model evaluation, as common procedures such as simple $k$-fold cross-validation no longer guarantee an unbiased assessment \citep{Bischl2023hpo,Hosseini2020}. 
\\
An additional challenge comes from the fact that, beyond algorithm HPs, there are also preprocessing HPs, which specify the steps applied to the data before it is fed into the learning algorithm (e.g., selecting the set of features for prediction or determining how missing feature values are handled; \citealp{mlr3book_7sequential,Bischl2023hpo}). While the tuning of algorithm HPs is rightfully considered important for model performance, the relevance of tuning preprocessing HPs should not be overlooked. Preprocessing steps can make or break a model's predictive performance, and solely relying on user expertise to specify these steps (which is the alternative to tuning) is often impractical and may result in arbitrary decisions \citep{Kuhn2013}. Despite this, reports of tuning preprocessing HPs aside from feature selection are relatively rare. This could be because integrating preprocessing HPs into automated tuning workflows typically requires advanced programming expertise, which not all applied ML users have, or because this possibility is not widely recognized. Importantly, the limited use of automated tuning procedures for preprocessing HPs does not mean that these HPs are not being tuned at all. In fact, it appears fairly common for applied ML users to experiment informally with different preprocessing options \citep{Hosseini2020,Hofman2023,Lones2024}, often without realizing that this constitutes a form of (manual) HP tuning.
If this type of tuning is indeed conducted subconsciously, it will also remain unaccounted for during model evaluation, thereby increasing the risk of drawing overly optimistic conclusions about the model’s performance.\\
To avoid such issues, it is essential to educate users in applied settings about the different types of HPs, the different forms of HP tuning, and how tuning can impact both the true and estimated performance of prediction models. Although valuable literature already exists describing the concept of HP tuning and various automated procedures \citep[e.g.,][]{Bischl2023hpo,Bartz2023,Feurer2019}, this research primarily adopts the perspective of ML methods researchers who are concerned with evaluating the overall performance of ML algorithms used to generate prediction models. This focus does not align with the perspective of applied ML users, who are more interested in the performance of a specific prediction model. Although this literature is still useful for them---since the general principles described there essentially hold for all types of audiences---applied ML users additionally need specific guidance for developing their \enquote{final model} (a notion that does not exist in the methodological context).  Moreover, they may find it challenging to extract the relevant insights from literature aimed at a different audience with partly different needs. In contrast, literature explicitly directed toward applied ML users tends to either focus on general guidelines for ML-based predictive modeling, lacking detailed coverage of HP tuning \citep[e.g.,][]{Kuhn2013,Pfob2022,Lones2024,Kapoor2024reforms,Poldrack2020,Collins2024,VanRoyen2023}, or addresses HP tuning only within specific
%domain
research areas
\citep[e.g.,][]{Hosseini2020,Dunias2024}. Additionally, much of the existing HP tuning literature does not consider preprocessing HPs. Exceptions include the review by \cite{Bischl2023hpo}, which, however, touches on this topic only briefly.  This lack of detail is reasonable, given that preprocessing HPs can, in principle, be tuned using the same automated procedures as algorithm HPs. However, this perspective overlooks that preprocessing HPs are often tuned manually in applied settings, which carries implications different from those associated with automated tuning.\\
This paper aims to complement the existing literature by reviewing the implications and pitfalls of HP tuning in the generation and evaluation of prediction models from the perspective of applied ML users with varying levels of expertise. It explicitly distinguishes between preprocessing and algorithm HPs, as well as the different procedures commonly used to tune them in practice. A particular focus is placed on the potential for optimistically biased performance estimation, which is also illustrated using a real-world prediction problem from palliative care medicine.\\
The paper is structured as follows. Section~\ref{sec:prelim} introduces the key concepts related to predictive modeling using ML, including the two types of HPs. In the next two sections, the challenges and pitfalls that arise in the generation and evaluation of prediction models are described, differentiating between the setting where all HPs are pre-specified (Section~\ref{sec:setting1}) and the setting where one or more HPs are selected through tuning (Section~\ref{sec:setting2}). Section~\ref{sec:illustration} empirically illustrates the impact of different tuning and evaluation procedures on the estimated model performance. Section~\ref{sec:discussion} summarizes the key insights, discusses the limitations of the empirical study, and outlines future research directions.

\section{General concepts of predictive modeling using supervised ML}
\label{sec:prelim} 
\subsection{Terminology and notation}\label{sec:prelim_start}
The following terminology and notation is adapted from \citet{Bischl2023hpo}. 
Let $\dtrain$ be a labeled data set with $\ntrain$ observations. Accordingly, each observation $i$ ($i=1,\dots,\ntrain$) consists of an outcome $y^{(i)}$ (i.e.\ the variable to be predicted, also referred to as label or target) and a $p$-dimensional feature vector $\bs{x}^{(i)}$ (i.e.\ the $p$ variables used to predict $y^{(i)}$, also referred to as predictors), where $y^{(i)}$ and $\bs{x}^{(i)}$ can take any value from the outcome space $\mathcal{Y}$ and feature space $\mathcal{X}$, respectively. Two common types of prediction problems are regression, for which $y^{(i)}$ can be any real number (i.e.\ $\mathcal{Y} = \mathds{R}$), and classification,  for which $y^{(i)}$ can be one of $g$ classes (i.e.\ $\mathcal{Y}$ is finite and categorical with $|\mathcal{Y}| = g$). We assume that the observations in $\dtrain$ are independent and have been sampled from the same (unknown) probability distribution $\pdistr$.\\
The general aim of supervised ML is to \enquote{learn} a  model from the data set $\dtrain$ that is able to predict the outcome values of new observations. Essentially, a prediction model is a function $\hat{f}: \mathcal{X} \rightarrow \mathds{R}^g$ that maps any observed feature vector $\bs{x}$ to a prediction vector $\hat{f}(\bs{x})$ in $\mathds{R}^g$. The prediction vector $\hat{f}(\bs{x})$ either directly corresponds to the predicted outcome value (e.g., for regression, where $g=1$) or can be transformed accordingly (e.g., for classification, where  $\hat{f}(\bs{x})$ corresponds to predicted probabilities for each class and the predicted class could be the class with the highest probability).  The prediction model results from a learning pipeline $\lp$, which uses the data set $\dtrain$ to find the function $\hat{f}$ that yields the best predictions for the true outcome values in $\dtrain$. To stress that a prediction model $\hat{f}$ is based on learning pipeline $\lp$ and data set $\dtrain$, we write $\fdltrain$. The prediction model $\fdltrain$ can usually be parameterized, meaning that it is defined by a set of parameters $\thetadl$ (simply denoted as $\bs{\hat{\theta}}$  when data set and learning pipeline are clear from context and $\bs{\theta}$ when referring to the parameters prior to estimation).\\
There are two key processes associated with $\lp$ and $\fdltrain$, which we will explore in more detail throughout the paper: (i) the training process, in which the learning pipeline $\lp$ is applied to $\dtrain$ and estimates the parameters $\thetadl$ and thus the prediction model $\fdltrain$, and (ii) the prediction process, in which $\fdltrain$ is used to make predictions for an observation (whether from $\dtrain$ or from a new data set) with feature vector $\bs{x}$, resulting in $\fdltrain(\bs{x})$. 
Note that to make predictions on a new data set, the outcome does not need to be observed (it would only be necessary for evaluating those predictions). The training and prediction processes serve as the foundation for more complex processes related to the development of prediction models, which we will address in Section~\ref{sec:prelim_processes}. 
 
\subsection{Learning pipeline}\label{sec:prelim_lp}
Each learning pipeline $\lp$ contains a learning algorithm as a central component but can also include several preprocessing steps that are performed before the algorithm is applied to the data. Since preprocessing steps are a particular focus of this paper, we use the term ``learning pipeline'' instead of the more common term ``learner'' to emphasize that $\lp$ can consist of several components. Note that for now, we consider all components of $\lp$ as fixed, but we will discuss the case in which they can be modified in Section~\ref{sec:prelim_hp}. 

\subsubsection{Learning algorithm}\label{sec:prelim_lp_algo}
The choice of learning algorithm usually depends on the specific prediction problem. For example, if the desired prediction model is a decision tree (which is the case for the real-world prediction problem considered in Section~\ref{sec:illustration}), a possible algorithm choice is the well-known Classification and Regression Tree algorithm (CART), which partitions the feature space $\mathcal{X}$ by a sequence of binary splits into terminal nodes and assigns a prediction value to each terminal node \citep{Breiman1984}. 
In this case, the parameters of the learning algorithm contained in $\thetadl$ are the splitting rules that generate the tree structure (i.e.\ which features are used with which threshold value) and the prediction values at each terminal node.  The learning algorithm can also consist of multiple individual algorithms that are combined into one overall algorithm (e.g., random forests). These types of algorithms are referred to as ensemble methods, but will not be discussed further in this paper.  In general, the choice of algorithm has a large impact on the hypothesis space of the learning pipeline, i.e.\ the set of prediction models the learning pipeline can generate. For example, selecting a standard linear regression as algorithm (with $\thetadl$ containing the regression coefficients) would imply that the corresponding learning pipeline would not be able to learn prediction models that do not correspond to linear combinations of the features (e.g., polynomials).

\subsubsection{Preprocessing}\label{sec:prelim_lp_preproc}
\begin{figure}[t]
    \centering
    \includegraphics[width=1\linewidth]{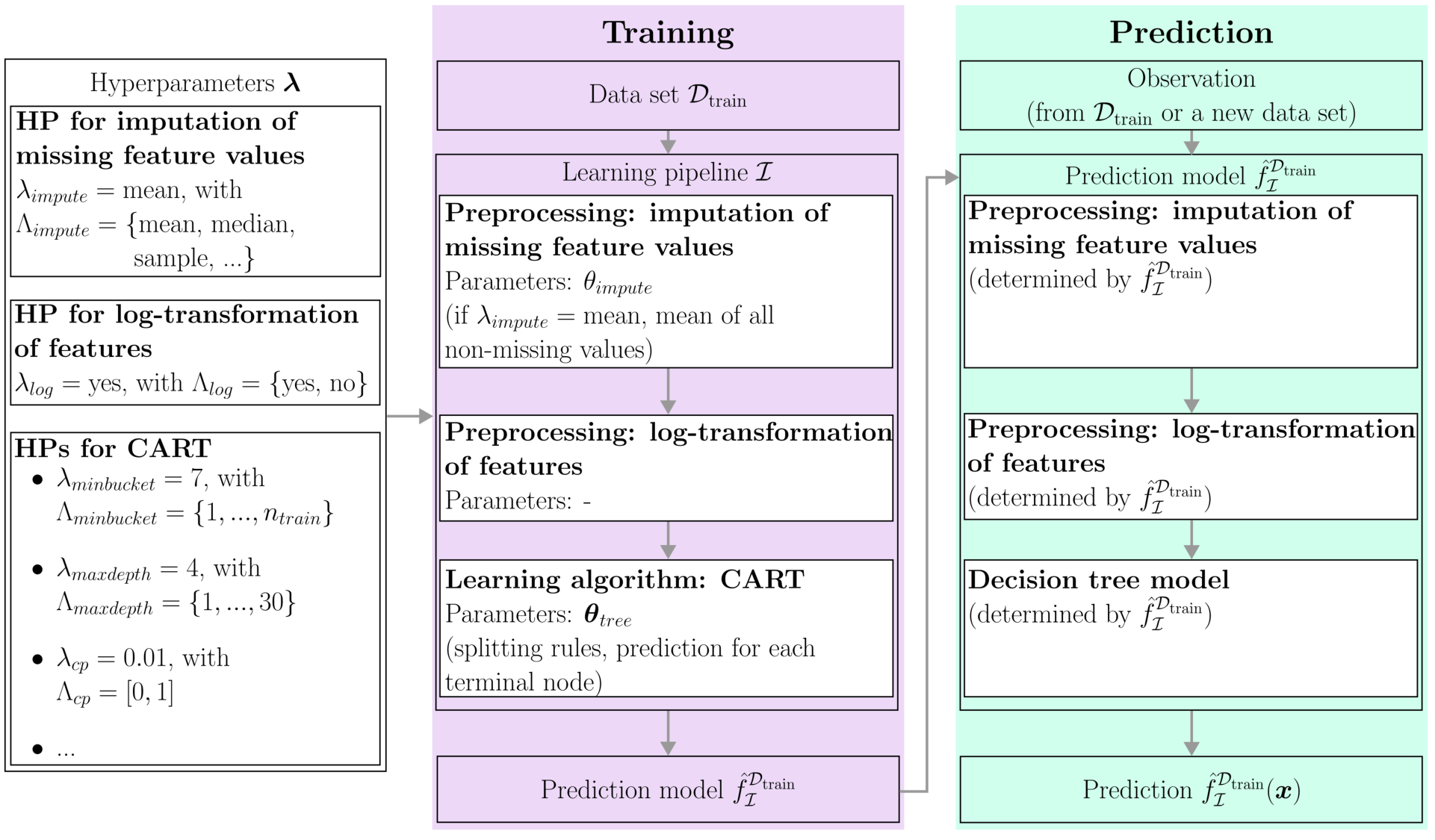}
    \caption{Example of a learning pipeline $\lp$ consisting of two preprocessing steps and one learning algorithm. Left panel: HPs of  the learning pipeline, with each HP set to an example value. Middle panel: Training process, where the learning pipeline is applied to the data set $\dtrain$ to generate the prediction model $\fdltrain$. Right panel: Prediction process, where a prediction for an observation with feature vector $\bs{x}$ is obtained by reapplying all preprocessing steps, followed by the prediction model resulting from the learning algorithm (here: a decision tree).} 
    \label{fig:prelim}
\end{figure}

While a data set can, in theory, be fed directly into the algorithm (i.e.\ the algorithm is the only component of the learning pipeline), it typically undergoes some modification first. This process can be referred to as data preprocessing and encompasses all the steps taken to transform the data set from its rawest available form into the final form provided as input to the learning algorithm \citep{Kapoor2024reforms}. Data preprocessing steps are usually performed to improve the performance of the resulting prediction model, to enable the data to be (better) handled by the learning algorithm \citep{mlr3book_9preprocessing}, or to improve the 
interpretability of the resulting prediction model. To better illustrate the different characteristics of preprocessing steps and their implications on the training and prediction process, we consider a simple learning pipeline as an example, which is also depicted in Figure~\ref{fig:prelim} (middle panel).
It consists of two preprocessing steps, which are followed by the CART algorithm. The first preprocessing step is the replacement of missing feature values using mean imputation, and the second preprocessing step is the log-transformation of features.

\paragraph{Parameterized vs.\ parameterless steps} Based on this example learning pipeline, we can make a first distinction between preprocessing steps. This distinction concerns whether the steps have parameters estimated from $\dtrain$ (with these parameters included in $\bs{\theta}$) or whether they are parameterless and are carried out independently for each observation  \citep{Kapoor2024reforms,mlr3book_7sequential}. 
In the example, the replacement of missing feature values is a parameterized preprocessing step, as it involves the parameter $\theta_{impute}$, representing the mean of all non-missing values estimated from $\dtrain$. In contrast, the log-transformation of features does not involve any parameters. Other examples of preprocessing steps with parameters include centering or scaling of features, where parameters such as the mean or standard deviation are estimated from $\dtrain$. On the other hand, creating a new feature by summing multiple features serves as another example of a parameterless preprocessing step.

\paragraph{Application during prediction vs.\ training only}
The second key distinction in preprocessing steps concerns whether they are applied only during the training process as part of the learning pipeline or also during the prediction process. This distinction is closely related to whether a preprocessing step modifies only the feature distribution or also affects the outcome distribution. More formally, let $\bs{y}$ denote the outcome vector in $\dtrain$. If, after applying all preprocessing steps in the learning pipeline during training, $\bs{y}$ remains unchanged, we classify the step as affecting only the feature distribution. Otherwise, the step affects the outcome distribution, for example, by removing or adding observations or transforming outcome values.\\
We first consider preprocessing steps that affect only the feature distribution. These comprise all preprocessing steps mentioned above, including those in the example learning pipeline. Additional examples are dimensionality reduction techniques (e.g., principal component analysis), feature selection, or data cleaning steps that do not alter the outcome distribution (e.g., correction of errors in features) \citep{Kuhn2013, mlr3book_9preprocessing}. Preprocessing steps of this type must be applied not only during training but also during prediction, in the same sequence as in the learning pipeline. This ensures that the model produced by the learning algorithm receives the data in the same format during prediction as it did during training, preserving the validity of the model \citep{mlr3book_7sequential}. This requirement implies that these steps are not only components of the learning pipeline $\lp$ but also part of the resulting prediction model $\fdltrain$. Consequently, if a learning pipeline $\lp$ includes $h$ preprocessing steps that only affect the feature distribution, the prediction model $\fdltrain$ is not a single function but a function composition of $h+1$ functions (omitting $\dtrain$ and $\lp$ for simplicity of notation):
\begin{equation}
\hat{f}_{h+1}(\hat{f}_h(\dots(\hat{f}_{1}(\bs{x})))),  
\end{equation}
where $\hat{f}_{h+1}$ corresponds to the model resulting from the learning algorithm, and $\hat{f}_{h},\dots,\hat{f}_1$ reflect the $h$ preprocessing steps.  Accordingly, a more accurate name for a prediction model would be prediction model \textit{pipeline}, but for brevity, we will continue to use the former. Returning to the example learning pipeline, the resulting prediction model is a  composition of three functions,  $\hat{f}_3(\hat{f}_2(\hat{f}_1(\bs{x})))$, where $\hat{f}_1$, $\hat{f}_2$, and $\hat{f}_3$ correspond to the imputation step, the log-transformation step, and the decision tree model, respectively. When making a prediction for one or more observations, all three functions must be applied (see  Figure~\ref{fig:prelim}, right panel). Importantly, if any functions constituting the prediction model %(except for identity functions representing preprocessing steps only applied during training)
are omitted during the prediction process, or if any preprocessing or algorithm parameters  are re-estimated on a new data set for which predictions are to be made, the validity of the prediction model may be compromised. However, in practice, this pitfall is often unavoidable for users who wish to apply a model but were not involved in its development, as studies introducing new prediction models frequently fail to report the preprocessing steps performed prior to applying the learning algorithm \citep{Kapoor2024reforms}. \\
In contrast to preprocessing steps that only affect the feature distribution, preprocessing steps that modify the outcome distribution are not necessarily applied during prediction. Here, we must distinguish between steps aimed at improving compatibility with the learning algorithm and those intended to alter the scope or interpretation of the prediction model. An example of the first type is (invertible) transformations applied to the outcome during training, such as a log-transformation to reduce skewness. To ensure predictions are returned on the correct scale, these transformations must be reversed during prediction \citep{mlr3book_9preprocessing}. For instance, if the outcome was log-transformed during training, the model will output $\mathrm{log}(\fdltrain(\bs{x}))$, which must then be exponentiated to restore the prediction to its original scale. Note that some other compatibility-focused steps are not applied at all during prediction. In the context of classification problems, this includes class-balancing steps such as oversampling, where observations from the least prevalent class are randomly resampled to overcome class imbalance effects during the training process (see, e.g., \citealp{Kuhn2013}, for more details). In the notation of the prediction model as a function composition introduced above, 
preprocessing steps that are applied only in their inverted form or not at all during prediction are represented as inversion function or identity function, respectively. \\
In contrast, preprocessing steps that modify the outcome to alter the scope or interpretation of the prediction model should be consistently applied during prediction. For example, if a continuous outcome is discretized to convert a regression problem into a classification problem \citep{Hofman2023}, this (irreversible) transformation must also be applied to the true outcome during prediction in order to enable a meaningful comparison between the predictions and the actual outcome values. Such transformations of the outcome are not part of the prediction model itself (which maps $\bs{x}$ to predictions, not $y$), but must be performed alongside the prediction process. Moreover, since the outcome values are generally unknown when making predictions for observations from a new data set that does not correspond to $\dtrain$, these transformations are typically not actual steps executed when making predictions but instead determine how the predictions are interpreted.

 \subsection{Hyperparameters}\label{sec:prelim_hp}
Until now, we have assumed that the learning pipeline $\lp$ is fixed.  However,  individual components of $\lp$ usually have several hyperparameters (HPs), which determine their specific configuration and thus substantially influence the resulting prediction model.
This also applies to the learning pipeline example considered in the previous section, for which possible HPs are shown in the left panel of Figure~\ref{fig:prelim} (see below for further explanation). In contrast to the parameters $\bs{\theta}$, which are estimated as outputs of the learning pipeline, the HPs serve as inputs. This means that they must be specified before the learning pipeline is applied to the data set \citep{Bischl2023hpo}. 

\subsubsection{Additional notation for HPs}\label{sec:prelim_hp_notation}
The following notation is based on \citet{Feurer2019}. We denote the $\hpindsmall$th HP of a learning pipeline as $\lambda_\hpindsmall$, which is selected from its domain $\Lambda_\hpindsmall$ (i.e.\ $\lambda_\hpindsmall \in \Lambda_\hpindsmall$). The domain of $\lambda_\hpindsmall$ can generally be real-valued, integer-valued, binary, or categorical, as we will see in the examples given below. All $\hpindlarge$ HPs of a learning pipeline can be summarized as a vector $\bs{\lambda} = (\lambda_1,\ldots,\lambda_\hpindlarge)$ and their overall configuration space as $\bs{\Lambda} = \Lambda_1 \times \Lambda_2 \dots \times \Lambda_\hpindlarge$ (with $\bs{\lambda} \in \bs{\Lambda}$). Note that  $\bs{\Lambda}$ may contain conditionality, meaning that some HPs might only be relevant when one or more other HPs are set to a certain value (see below for examples). \\
As described in Section~\ref{sec:prelim_lp}, the learning pipeline consists of several preprocessing steps and a learning algorithm. We can consequently differentiate between preprocessing and algorithm HPs, which we denote as $\hpp$ and $\hpa$ (i.e.\ $\bs{\lambda} = (\hpp, \hpa))$.

\subsubsection{Algorithm HPs}\label{sec:prelim_hp_algo}
Each learning algorithm usually has several HPs, which are specified by the software package used and can have a large impact on its complexity, speed, and other important properties of the algorithm \citep{Bischl2023hpo}. 
For example, the HPs of the CART algorithm include the minimum number of observations in any terminal node ($\lambda_{minbucket})$, the maximum tree depth, with the root node counted as depth 0 ($\lambda_{maxdepth}$),  and the factor by which a split needs to decrease the overall lack of fit to be attempted ($\lambda_{cp}$) \citep{rpart_package}. In the CART implementation of the \texttt{R} package \texttt{mlr3} \citep{mlr3_package}, the respective HP domains are $\Lambda_{minbucket} = \{1,\ldots,\ntrain\}$, $\Lambda_{maxdepth} = \{1,\ldots,30\}$ (both being integer-valued domains), and $\Lambda_{cp} = [0,1]$ (real-valued domain). 
Most algorithm HPs have default values that are specified by the software in which they are implemented 
 (e.g., in \texttt{mlr3}, $\lambda_{minbucket} = 7$ per default). \\
Note that since there is usually more than one algorithm suitable for a given prediction problem, the choice of algorithm can also be seen as an HP of the learning pipeline (with the HPs associated with each algorithm representing conditional HPs that are only relevant when the respective algorithm is used; \citealp{Bischl2023hpo}).  This creates an even more flexible but also complex learning pipeline, which is why, in this paper, we assume that the algorithm has already been selected.

\subsubsection{Preprocessing HPs}\label{sec:prelim_hp_preproc}
As mentioned above, it is not only possible to specify learning algorithm HPs but also preprocessing HPs \citep{mlr3book_7sequential,Bischl2023hpo}. 
In principle, whenever multiple options exist for performing a preprocessing step, these options can be considered as different HP values of the respective preprocessing step. \\
First, the choice of whether a preprocessing step $PS$ is applied at all can be considered as a binary HP $\lambda_{PS}$ with $\Lambda_{PS} =  \{\mathrm{yes, no}\}$ (e.g., whether features should be log-transformed or not). 
Second, there is often more than one possible option for performing a preprocessing step. 
For example, the influence of outliers in features can be reduced by replacing all values that are outside the range $[x_{min}, x_{max}]$ by $x_{min}$ and $x_{max}$, respectively (``winsorizing''; \citealp{Steyerberg2019}). There are different options to specify $x_{min}$ and $x_{max}$, which means that $\lambda_{x_{min}}$ and $\lambda_{x_{max}}$ are HPs of the winsorizing preprocessing step (e.g., \citealp{Steyerberg2019}, suggests percentiles such as $\lambda_{x_{min}}= 1\text{st percentile}$ and $\lambda_{x_{max}}= 99\text{th percentile}$).\\
Several possible options also exist for the imputation of missing feature values. For example, imputation can be based on the feature's mean or median, or on a sampled value from its empirical distribution    \citep[as illustrated in][]{mlr3book_9preprocessing}. This constitutes a (categorical) preprocessing HP 
$\lambda_{impute}$ with $\Lambda_{impute} = \{\text{mean, median, sample},\ldots\}$. \\
Another typical example of a preprocessing step with many possible options is feature selection. To define HPs in this context, we have to differentiate between filter and wrapper methods (the following explanations are based on \citealp{mlr3book_6featuresel}, who also provides more  details and additional examples). Filter methods are preprocessing steps that assign a numeric score to each feature (e.g., the correlation coefficient $\rho$ between each feature and the outcome) and select a set of features according to this score (e.g., all features with $\rho > 0.2$). Consequently, the set of selected features is the parameter of the filter (i.e.\ $\theta_{filter}$, with, e.g., $\hat{\theta}_{filter} = \{x_6, x_{8}, x_{21}, x_{25}\}$), while its specific configuration can be modified by its HPs. For example, there are different options to define the score ($\lambda_{filter_1}$, with $\Lambda_{filter_1} =  \{\text{correlation, variance, importance score},\ldots\}$) and to select the features based on their score ($\lambda_{filter_2}$, with $\Lambda_{filter_2}  = \{\text{top $r$ features, all features with a score} \geq \tau,\dots\}$, where $r$ and $\tau$ themselves are HPs that are conditional on $\lambda_{filter_2}$). Instead of using filter methods, it is also possible to directly specify the set of features that should be selected. 
In this case, the set of selected features is an input rather than an output of the learning pipeline and is therefore the HP ($\lambda_{features}$) of the feature selection step. For example, if only the features $x_6, x_{9}$, and $x_{21}$ should be used by the learning algorithm, then $\lambda_{features} = \{x_6, x_{9}, x_{21}\}$. 
In many applications, $\lambda_{features}$ is not specified once by the user, but different values of $\lambda_{features}$ are tried and evaluated on $\dtrain$. This process is referred to as a wrapper method but is, in fact, a special case of HP tuning, which will be discussed in Section~\ref{sec:setting2_gen}. \\
Note that the individual HP values can also be application-specific. For example, in the real-world prediction problem considered in Section~\ref{sec:illustration}, several options for aggregating 17 individual features covering physical symptoms, psycho-social burden, family needs, and practical problems of palliative care patients to a sum score are reasonable (see Section~\ref{sec:setup_lp}).\\ 
In addition to specifying the preprocessing steps, the order in which they appear in the learning pipeline can technically be considered an HP as well. For instance, in the learning pipeline shown in Figure~\ref{fig:prelim}, the log-transformation step could also be applied before the imputation step, resulting in a different $\hat{\theta}_{impute}$ and, therefore, potentially a different prediction model. However, we will not consider this type of preprocessing HP further in the remainder of this paper.\\
As already indicated by the examples above, many preprocessing HPs are conditional on other preprocessing HPs (e.g., the winsorizing HPs $\lambda_{x_{min}}$ and $\lambda_{x_{max}}$ are only relevant when winsorizing is the chosen method to reduce the influence of feature outliers, which could also be implemented by transforming the features instead). Moreover, in contrast to algorithm HPs, preprocessing HPs often cannot be set by a single software function argument (for example, all HPs of the CART algorithm named in the previous section can be specified within a single \texttt{R} function, using, e.g., the argument \texttt{minbucket} for $\hpminb$); instead, in many cases, the different options for a specific preprocessing step are implemented by different software packages. Consequently, there is often no formal HP domain, and defining the domain such that it contains all possible HP values may not even be feasible  (e.g., for $\lambda_{impute}$, defining $\Lambda_{impute}$ would require collecting all available methods for imputing missing values).  Moreover, many preprocessing HPs do not have a formal default value, although the option of not applying a preprocessing step (if applicable and not leading to an error) seems to be a reasonable default value that we will adopt in the following.\\
In contrast to algorithm HPs, it seems that preprocessing HPs---apart from those related to feature selection---are rarely discussed or referred to as such in ML applications %studies presenting new prediction models 
(see, e.g., the systematic reviews of \citealp{Dhiman2022_methconduct}, and \citealp{Navarro2023_systemreview}, where such terms were not mentioned).  ML methods research usually also focuses on algorithm HPs rather than preprocessing HPs. An exception is the benchmark study by \citet{Stueber2023}, which, among other factors, examines the impact of using principal component analysis in radiomics-based survival analysis.

 \subsubsection{Selection of HPs}\label{sec:prelim_hp_sel} 
 While it is usually possible to leave all HPs at their respective default value, it is common to modify them in an attempt to optimize the prediction model generated by the learning pipeline. This can also be necessary if there is no specified default value.
The term ``optimization'' here often refers to the predictive performance of the model but can also take into account other criteria such as simplicity, interpretability, or runtime to generate the model \citep{Domingos2012, Pfob2022, DeHond2022, Bischl2023hpo}. Note that the selection of HPs can be considered a \enquote{researcher degree of freedom} \citep{Simmons2011}, as it is one of many choices that users must make throughout the model development process (other choices are, e.g., how predictive performance is assessed; \citealp{Hofman2017,Klau2020,Hosseini2020}). \\
We can distinguish between two primary types of HP selection: data-independent and data-dependent procedures. Data-independent HP selection does not make use of the data set $\dtrain$ and is ideally based on the user's knowledge about the data set and learning algorithm. 
For example, sensible algorithm HPs can be selected when users are experienced with the learning algorithm or when corresponding recommendations from the literature (e.g., previous benchmark studies) are available \citep{Bartz2023,Bischl2023hpo}.
Similarly, some preprocessing HPs may be inferred from substantive knowledge about the data set (e.g., which set of features should be selected) or knowledge about how the learning algorithm is affected by certain data set characteristics (e.g., whether the algorithm is sensitive to outliers in features, which requires some form of transformation; \citealp{Kuhn2013}).  
 An example of data-independent HP selection on the basis of model simplicity is the specification of the maximum tree depth in the real-world prediction problem considered in Section~\ref{sec:illustration}, where the project team set the HP to $\lambda_{maxdepth} =4$ to ensure that the resulting decision tree can be implemented in clinical practice.\\
 In cases where users have insufficient knowledge about the data and learning algorithm to ensure a reasonable HP selection but wish to avoid arbitrary or default HP values, it is possible to use the data set $\dtrain$ to select optimal HP values.
 This process corresponds to a data-dependent HP selection, but terms such as HP tuning and (data-driven) HP optimization are more common \citep[e.g.,][]{Bischl2023hpo,Probst2019,Bartz2023}. We will accordingly use the term HP tuning in the remainder of this paper. 
 Note that HP tuning implies that not only the parameters $\bs{\theta}$ are estimated from the data set $\dtrain$ but also one or more HPs in $\bs{\lambda}$. 
 HP tuning thus generally complicates model generation and evaluation, which will be described in more detail in Section~\ref{sec:setting2}. \\
 Importantly, there are HPs that should not be selected through tuning. 
For learning algorithms, this includes, for example, the number of trees ($\lambda_{num.trees}$) in the random forest algorithm for classification problems: Due to the monotonous relation between $\lambda_{num.trees}$ and model performance in most cases, the largest computationally feasible number of trees should be chosen \citep{Probst2018}. Regarding preprocessing HPs, this typically applies to those associated with steps that alter the scope or interpretation of the prediction model (see Section~\ref{sec:prelim_lp_preproc}).  As such steps require careful specification, the corresponding HPs should be set based on user expertise (i.e.\ data-independently) rather than determined through tuning. \\
To indicate how the value of a HP $\lambda_\hpindsmall$ has been specified, we write $\hpji$ if the value is left at default value or selected independently of the data, and $\hpjii$ if the value was chosen through tuning.

\subsection{Model development processes} \label{sec:prelim_processes} 
The development of ML-based prediction models generally involves two key processes: (i) the generation of the prediction model $\fdltrain$ (model generation) and (ii) the evaluation of its predictive performance (model evaluation). Given our focus on HPs and their selection, we distinguish between two settings in the remainder of this paper. In Setting I, all HPs of the learning pipeline are pre-specified (i.e.\ either set to default values or selected independently of the data). In Setting II, one or more HPs are %not pre-specified and are 
selected through tuning.\\
Before explaining the principles and potential pitfalls of model generation and evaluation for both settings in Sections~\ref{sec:setting1} and \ref{sec:setting2}, we first clarify their general concepts.

\subsubsection{Model generation}
We refer to the model generation process as the set of processes required to obtain the final prediction model $\fdltrain$. 
In Setting I, the model generation process consists of a single training process, where the parameters that define the final prediction model are estimated from $\dtrain$ using the learning pipeline $\lp$ with pre-specified HPs. 
In Setting II, where one or more HPs are selected through tuning, the model generation process consists of a tuning process conducted on $\dtrain$ (which yields the tuned HPs), followed by a training process, where, similar to Setting I, the parameters of the final prediction model are estimated from $\dtrain$ using the learning pipeline $\lp$ with tuned HPs. 

\subsubsection{Model evaluation}\label{sec:prelim_processes_eval}
Once the final prediction model $\fdltrain$ has been generated, the next important step is its evaluation.
Since many algorithms yield black-box models that cannot be easily interpreted, and are thus difficult to assess for plausibility without additional tools (see, e.g., \citealp{Molnar2022}), a key quantity in the evaluation of a model is its prediction error.
In the context of this work, we will accordingly use the term \enquote{model evaluation} synonymously with determining a model’s prediction error. The prediction error indicates how well a model performs on new observations that are independently drawn from the same distribution as the observations in $\dtrain$ (i.e.\ from $\pdistr$). It is specified with respect to a loss function $L$, which assesses the discrepancy between true outcomes and predictions and constitutes the performance measure. Formally, the prediction error of $\fdltrain$ can be defined as 
\begin{equation}\label{eq:pe}
    \mathrm{PE}(\fdltrain) = \mathrm{E}_{(\bs{x},y) \sim \pdistr}[L(\fdltrain(\bs{x}), y)]
\end{equation}
\citep{Hastie2009,Boulesteix2015framework,Bischl2023hpo}.
The loss function $L$ %%, or possibly multiple loss functions, 
can be chosen according to the prediction problem being addressed.
For instance, a common choice for $L$ in regression problems is the squared loss. In this case, the prediction error reflects %corresponds to 
the well-known mean squared error (MSE). Note that in equation~\eqref{eq:pe}, we assume for simplicity that $L$ corresponds to a point-wise loss function, although many commonly used performance measures (e.g., the area under
the receiver operating characteristic curve, AUC) would necessitate a more general definition  (provided in \citealp{Bischl2023hpo}). 
Nonetheless, all following statements regarding the prediction error hold regardless of this simplified (and more common) representation. \\
An estimate of the prediction error in equation~\eqref{eq:pe} can be obtained by using $\fdltrain$ to make predictions for an additional data set with new observations drawn from $\pdistr$ (referred to as test data set $\dtest$). The prediction error can then be estimated by evaluating the loss function $L$ for each observation and calculating the average across all observations (again, assuming a point-wise loss; \citealp{Bischl2023hpo,Hastie2009}). The resulting prediction error estimate for $\fdltrain$ can be denoted as $\widehat{\mathrm{PE}}(\fdltrain, \dtest)$. Note that the outcome values for $\dtest$ must be observed; otherwise, the loss function $L$ cannot be evaluated.  \\
The requirement for an additional data set, $\dtest$, for model evaluation can be challenging in applications where data resources are limited. Denoting $\ds$ as the only available data set at the time of model generation and evaluation, there are two general approaches for defining $\dtrain$ and $\dtest$: (i) all available data are used for model generation, in which case $\dtest$ is inevitably a subset of $\dtrain$ (i.e.\ $\dtrain = \ds$ and $\dtest \subseteq \dtrain$), or (ii) the model is generated on a (proper) subset of the available data, with the remaining subset held back for model evaluation (i.e.\ $\dtrain \subset \ds$ and $\dtest = \ds \setminus \dtrain$). For the first approach, there are several ways to define $\dtest$, each leading to a different evaluation procedure, which will be detailed in Section~\ref{sec:setting1_eval} (Setting I) and Section~\ref{sec:setting2_eval} (Setting II). \\
Depending on the chosen evaluation procedure, a potential issue can be data leakage, which occurs whenever information about the designated $\dtest$ is improperly available during the generation of the model to be evaluated
\citep{Kapoor2023leakage, Kapoor2024reforms, Kaufman2012,Rosenblatt2024,Hornung2023}. Since, in this case, the observations in $\dtest$ no longer truly represent new observations to which the model will be applied, and the model thus has an unfair advantage when predicting these observations, the resulting prediction error estimate can be optimistically biased. 
\cite{Kapoor2023leakage}  identify three general types of data leakage, which may arise from: (i) overlap between the data used for model generation and evaluation, (ii) violation of the assumption that all observations are independently drawn from the same distribution, or (iii) use of illegitimate features. In this paper, we will focus on overlap-induced data leakage but provide additional information on the other two types in Supplementary Section~A. Furthermore, we encounter an example of one of the other types in our empirical illustration in Section~\ref{sec:illustration}.\\
Finally, note that in some applications of ML (e.g., in the context of healthcare research), the process of assessing a model's performance on observations from $\pdistr$ is referred to as internal validation. 
This is in contrast to external validation, which evaluates how well the model predicts observations from different distributions (e.g., different time points or healthcare settings; \citealp{DeHond2022,Collins2024,VanRoyen2023,VanCalster2023}). 
As external validation is recommended to be performed in subsequent research only after successful internal validation \citep{Collins2024}, we will focus on internal validation in this paper. 
Note that, in general, the term ``evaluation'' should be preferred over ``validation'' as the latter suggests that a ``validated model'' has a low prediction error, which is not necessarily the case \citep{Collins2024}.

\section{Setting I: Pre-specified HPs} \label{sec:setting1}
In this section, we describe the model generation and evaluation process for Setting I. We accordingly assume that the learning pipeline $\lp$ is configured by HP values that are either set to their default values or selected independently of the data, i.e.\ $\hp=\hpi$. This aspect is emphasized by denoting the learning pipeline as $\lpi$. 
\subsection{Model generation} \label{sec:setting1_gen}
As stated in Section~\ref{sec:prelim_processes}, the model generation process in Setting I consists of a single training process. Moreover, as already outlined, \enquote{training} refers to the learning pipeline estimating the parameters $\bs{\theta}$ (which constitute the prediction model) from $\dtrain$. For brevity, we will also refer to this process as ``training the prediction model'' although it is the learning pipeline that is being trained and subsequently yields the prediction model. \\
Importantly, all parameters in $\bs{\theta}$ must be estimated, including those from preprocessing steps. The estimation of preprocessing parameters follows the sequence of their corresponding steps in the learning pipeline $\lpi$. This process is specified by the respective preprocessing step. For example, in the case of mean imputation, the corresponding parameter estimate is found by calculating the mean of all non-missing observations of the corresponding feature. \\
The parameters of the learning algorithm are usually estimated based on a loss function $l$ that measures the discrepancy between the true outcome and a prediction vector for each observation $i$, i.e.\ $l(y^{(i)}, f(\bs{x}^{(i)}))$. 
The algorithm parameters are then found by minimizing $\sum_{i=1}^{\ntrain} l(y^{(i)}, f(\bs{x}^{(i)}))$  (see, e.g., \citealp{Bischl2023hpo}, or \citealp{Bartz2023}, for more details). For example, in a regression problem where the learning algorithm corresponds to the CART algorithm, the splitting rules are found by minimizing the sum of squared errors and the prediction value for each terminal node corresponds to the mean of all outcome values in the respective node \citep{Breiman1984}. Note that the loss function $l$ may, but does not necessarily have to, align with the loss function $L$ from Section~\ref{sec:prelim_processes_eval}, which is used to estimate the prediction error.
\\
When estimating the parameters, the learning pipeline may not only capture the signal in $\dtrain$  which represents the true underlying data-generating mechanism
%% or : the "true" associations
$\pdistr$, but it may also erroneously learn the specific pattern of noise (i.e.\ unexplained variation) in $\dtrain$. The resulting prediction model is too adapted to $\dtrain$ and will perform worse on new observations (drawn from $\pdistr$) than on the observations in $\dtrain$. This is a well-known problem in prediction model training and is commonly referred to as overfitting \citep[e.g.,][]{Hastie2009,Kuhn2013, Steyerberg2019,Bischl2023hpo,Poldrack2020,DeHond2022}. The risk of obtaining an overfitted prediction model depends on both the data set $\dtrain$ (specifically on its signal-to-noise ratio, which tends to decrease as the number of observations decreases) and on the learning pipeline $\lpi$ used to train the model \citep{Poldrack2020, Lones2024}. 
The association between the characteristics of a learning pipeline and its tendency to overfit is not straightforward, but it is related to factors such as the size of its hypothesis space (i.e.\ the number of prediction models that can be trained by $\lpi$) and the procedure by which the model is chosen from the hypothesis space (e.g., whether the hypothesis space is searched exhaustively; \citealp{Domingos2012}). 
These factors can vary greatly between learning pipelines, especially depending on the type of learning algorithm and the chosen HP values.  
Note that the learning pipeline may also suffer from underfitting rather than overfitting, which occurs if it is not flexible enough to adequately model the underlying data-generating mechanism \citep{Hastie2009}.  \\
As mentioned above, after training the learning pipeline once (and only once) on $\dtrain$, the generation of the final prediction model is completed. 
This implies that if the model is found to have a poor predictive performance in the subsequent evaluation (e.g., due to over- or underfitting), the result either has to be accepted or the HPs of the learning pipeline have to be modified based on the evaluation result.
However, users should be aware that the latter approach corresponds to Setting II, which has different implications for model evaluation (Section~\ref{sec:setting2}). We denote the final prediction model as $\fdlitrain$ to emphasize that it is the result of training a learning pipeline configured with HP values $\hpi$.

\subsection{Model evaluation} 
\label{sec:setting1_eval}
\begin{figure}[!ht]
    \centering
    \includegraphics[trim={0 0cm 0 0},clip, width=0.9\linewidth]{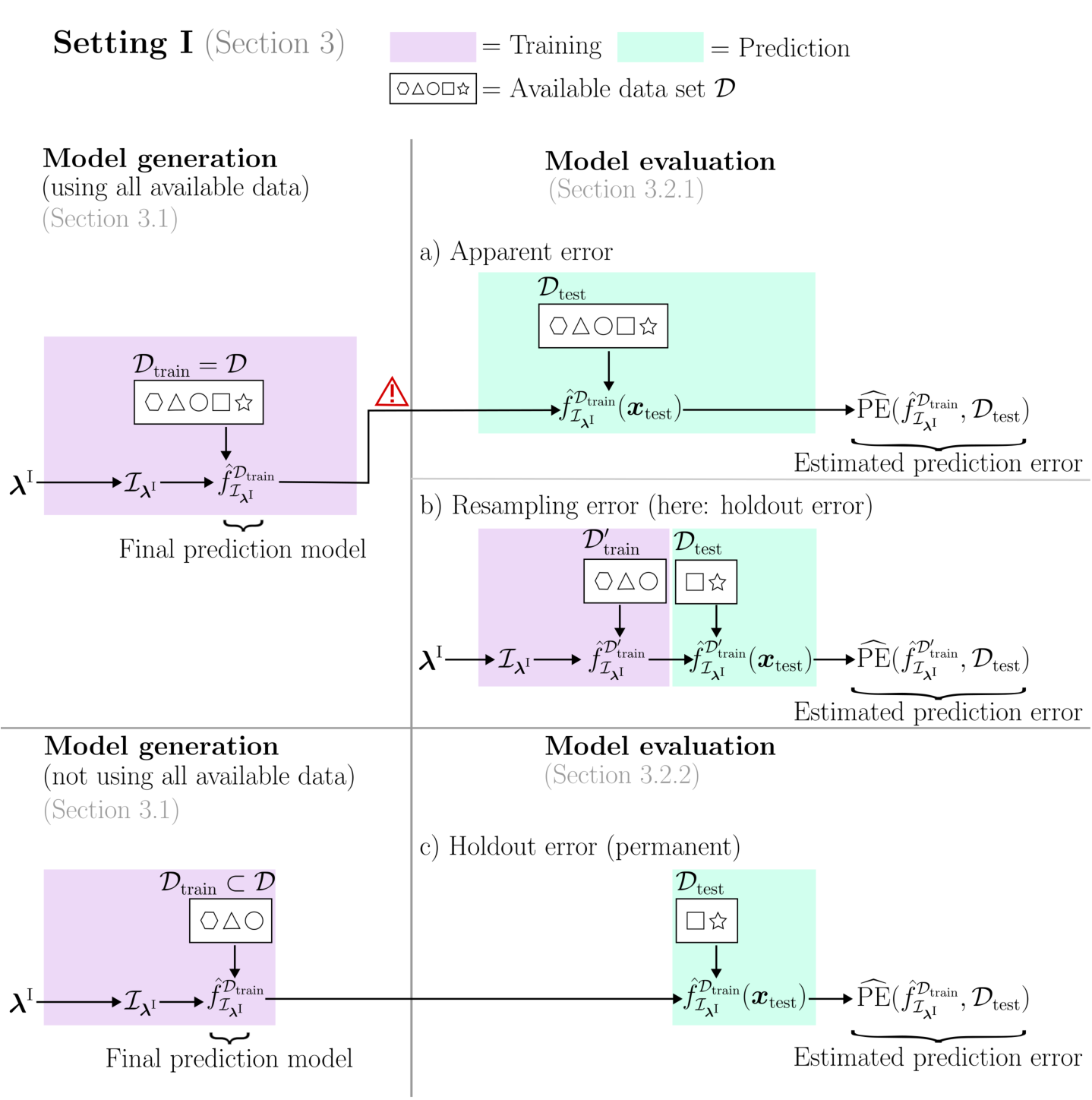}
    \caption{Overview of different model evaluation procedures and their relation to the model generation process if all HPs are pre-specified.  Data leakage is present if any subset of $\dtest$ used for prediction error estimation has also been employed to generate the evaluated prediction model (which is not necessarily the final model). In the figure, the point at which data \enquote{leaks} into the model evaluation is marked by the red caution symbol.}
    \label{fig:setting1}
\end{figure}

As outlined in Section~\ref{sec:prelim_processes_eval}, evaluating the prediction model $\fdlitrain$ requires a test data set $\dtest$, which is used to estimate the model's prediction error. In that section, it was also stated that evaluation procedures can be differentiated based on whether model generation (which corresponds to model training in Setting I) has been performed on all available data (with $\dtrain = \ds$ and $\dtest \subseteq \dtrain$)  or only on a (proper) subset of the available data (with $\dtrain \subset \ds$ and $\dtest = \ds \setminus \dtrain$). In the following sections, we examine the implications for model evaluation in more detail for both approaches. An additional graphical overview is provided in Figure~\ref{fig:setting1}.

\subsubsection{Evaluation of a model generated on all available data}\label{sec:setting1_eval_all}

\paragraph{Apparent error} A straightforward way to evaluate a prediction model trained on all available data is to estimate its prediction error using the same data set, i.e.\ $\dtrain = \dtest = \ds$. The resulting prediction error estimate is referred to as apparent error (see Figure~\ref{fig:setting1}, model evaluation a). 
As explained in Section~\ref{sec:prelim_processes_eval}, data leakage is present when information about the designated $\dtest$ is present during model generation.
For the apparent error, this is clearly the case, as $\dtest$ is equal to $\dtrain$. As a consequence, the apparent error is not able to detect any overfitting of the model (since the specific pattern of noise in $\dtrain$ exactly corresponds to that in $\dtest$) and will therefore be affected by a (possibly substantial) optimistic bias. 
Although this evaluation procedure is well-known to be flawed and has been frequently warned against in literature  \citep[e.g.,][]{Efron1986,Kuhn2013,Hastie2009,Poldrack2020,Collins2024}, it is often still the only prediction error estimate that is reported in studies presenting new prediction models \citep{Kapoor2023leakage, Poldrack2020}.

\paragraph{Resampling error} To avoid the optimistic bias caused by the overlap between $\dtrain$ and $\dtest$, several procedures exist that partition $\dtrain$ one or multiple times into two subsets for evaluation purposes while still training the final prediction model on the full data set.  These procedures can be referred to as resampling methods and the resulting estimate as the resampling error (see  Figure~\ref{fig:setting1}, model evaluation b). The following description is based on \cite{Simon2007}, \cite{Kuhn2013}, \cite{Bischl2023hpo}, and \cite{mlr3book_3evalbench}; see their work for more details.\\
The simplest resampling method is the holdout or split-sample method, where $\dtrain$ is randomly split into two subsets with different purposes: One subset, denoted as $\dtraine$, is used to retrain the same learning pipeline $\lpi$ that has been used to obtain the final prediction model. This results in an additional prediction model $\fdlitraine$, whose prediction error is then estimated on the second subset, which serves as $\dtest$. The holdout method essentially has two drawbacks, whose impact on the prediction error varies according to the split ratio and the absolute number of observations in $\dtraine$ and $\dtest$ (denoted as $\ntraine$ and $\ntest$).
First, while the holdout method ensures a clean separation between $\dtraine$ and $\dtest$, it does not evaluate the actual prediction model trained on $\dtrain$ but the additional prediction model trained on $\dtraine$, which does not necessarily coincide with the former. 
Since the additional prediction model is trained on fewer observations (i.e.\ $\ntraine < \ntrain$), estimating its prediction error on $\dtest$ yields a pessimistically biased estimate for the prediction error of $\fdlitrain$.
Second, the smaller $\ntest$, the more the prediction error estimate varies depending on which observations are assigned to $\dtest$ (i.e.\ the higher the variance of the holdout estimator). 
As a consequence, specifying the split ratio for the holdout method requires a careful trade-off between bias and variance. \\
A commonly used variation of holdout is $k$-fold cross-validation (CV), where $\dtrain$ is randomly split into $k$ subsets (or folds) of approximately the same size, with 5 or 10 being typical choices for $k$.
Based on the $k$ splits, the procedure described for the holdout method is repeated $k$ times: In each repetition (in this context also referred to as resampling iteration), the learning pipeline is trained on $k-1$ subsets of $\dtrain$ (constituting $\dtraine$), and the prediction error of the resulting model is estimated on the remaining subset (constituting $\dtest$).
The final prediction error estimate is obtained by averaging the $k$ prediction error estimates, which leads to the CV estimator having a smaller variance than a holdout estimator with the same split ratio. However, the prediction error estimate resulting from CV is also pessimistically biased because the evaluated prediction models are again trained on less than $\ntrain$ observations, although this bias decreases with increasing $k$ ($\ntraine=\frac{k-1}{k}\cdot \ntrain$). \\
Other common resampling methods include repeated versions of holdout and CV (to reduce the variance of the corresponding estimator) and bootstrapping.  
Repeated holdout and bootstrapping are similar in their execution, except that for repeated holdout, the observations constituting $\dtraine$ in each resampling iteration are drawn without replacement, while they are drawn with replacement for bootstrapping.\\
As stated above, all resampling methods require the learning pipeline to be retrained on one or multiple subsets $\dtraine$, each of which is a (proper) subset of $\dtrain$ (i.e.\ $\dtraine \subset \dtrain$). In this context, a flawed evaluation procedure would be to apply all preprocessing steps on the full data set $\dtrain$ and retrain only the learning algorithm on $\dtraine$ during resampling. This ``incomplete resampling'' \citep{Simon2003incomplete} results in another form of data leakage, as in each resampling iteration, the observations in the respective $\dtest$ subset have already been used to train part of the learning pipeline (i.e.\ the preprocessing steps). Incomplete resampling has been frequently warned against in the literature  \citep[e.g.,][]{Kapoor2024reforms,Poldrack2020,Pfob2022,DeHond2022,Hofman2023}, and the resulting optimistic bias has been demonstrated by illustrations on real data \citep[e.g.,][]{Hornung2015,Rosenblatt2024} and corrected reanalyses of published studies \citep[e.g.,][]{Neunhoeffer2019, Kapoor2023leakage}. Yet, it still seems to be a common pitfall in the evaluation of prediction models (see \citealp{Kapoor2023leakage}, and references therein), which is probably caused by a lack of understanding of its implications. In addition,  if the learning pipeline is not implemented as a single object that can be trained with a single function call such as \mbox{\texttt{train(learning\_pipeline)}}
%learning pipeline cannot be implemented as such in the software framework used 
(e.g., this is possible in \texttt{R} with the \texttt{mlr3} or \texttt{recipes} package by \citealp{mlr3_package}, and \citealp{recipes_package}), each preprocessing step must be manually repeated in every resampling iteration. In such cases, users may consider incomplete resampling a time-saving shortcut, without realizing that it introduces data leakage. To avoid incomplete resampling, every component of the learning pipeline, including the preprocessing steps, must be retrained in each resampling iteration. The only preprocessing steps that can be safely applied to the full data set prior to resampling are those that are both parameterless and precede the first parameterized preprocessing step in the learning pipeline.

\subsubsection{Evaluation of a model generated on a subset of the available data}\label{sec:setting1_eval_subset}
If the final prediction model has been trained on a subset of the available data (i.e.\ $\dtrain \subset \ds$), its prediction error can be estimated using the remaining observations as $\dtest$ (see Figure~\ref{fig:setting1}, model evaluation c). This means that the training process does not need to be repeated, as there is no need to use resampling methods. Note that this procedure is technically equivalent to the holdout method introduced above, except that the model trained on $\dtrain$, which corresponds to $\dtraine$ in the holdout method above, is the final prediction model and has not only been trained for evaluation purposes.  Accordingly, the procedure is referred to as holdout or split-sample method as well, which can make it difficult to infer which procedure was used when the evaluation result of a model is reported. We use the terms temporary holdout (described in Section~\ref{sec:setting1_eval_all}) and permanent holdout (described here) to distinguish the two procedures. \\
In principle, most points discussed in the previous section affecting temporary holdout (including data leakage due to incomplete resampling) also apply to permanent holdout. Again, the only difference is that, for the temporary holdout, the model trained on a subset of the available data is used solely for evaluation purposes, whereas it serves as the final prediction model for the permanent holdout. Consequently, the prediction error estimate derived from the permanent holdout is not pessimistically biased; instead, it is an unbiased estimate of a prediction error that is indeed higher (i.e.\ worse) than that of a model using all available data. Since not using all available data for training the prediction model essentially corresponds to a loss of important information, the permanent holdout method is only recommended if the number of observations in $\ds$ is sufficiently large or if repeating the training process is computationally expensive or infeasible \citep{Collins2024}.

\section{Setting II: HPs selected through tuning} \label{sec:setting2}
In this section, we review the model generation and evaluation process for Setting II, where one or more HPs are selected through tuning.

\subsection{Model generation} \label{sec:setting2_gen}
\subsubsection{Overview}\label{sec:setting2_gen_princ}
HP tuning generally aims to improve the predictive performance of a model \citep{Bischl2023hpo,Probst2019}. Using the terminology introduced in Section~\ref{sec:prelim_processes_eval}, this corresponds to finding the HP configuration that minimizes the model's prediction error. To simplify notation, we will assume for now that all HPs are to be tuned, but will revisit the scenario where this does not apply later in this section. Under this assumption, the HP tuning problem can be formalized as:
\begin{equation}\label{eq:tuning}
    \hpstar = \underset{\hp \in \bs{\Lambda}}{\mathrm{argmin}}   \,\mathrm{PE}(\fdlhptrain),
\end{equation}
where $\fdlhptrain$ is the final prediction model resulting from training the learning pipeline $\lp$ configured with HPs $\hp$, and $\hpstar$ denotes the theoretical optimum \citep{Bischl2023hpo}. The lowest prediction error (i.e.\ the best performance) that can be achieved using $\hpstar$ as HP configuration depends on several factors, such as the HPs to be tuned, the selected learning algorithm, the performance measure, and the prediction problem in general \citep{Probst2019}. Note that in the following, we refer to the prediction error of a model that results from training a learning pipeline determined by a candidate HP configuration $\hpcand$, i.e.\ $\hat{f}^{\dtrain}_{\mathcal{I}_{\bs{\lambda}^{(c)}}}$, simply as the prediction error of $\hpcand$ for brevity.  It should also be noted that equation~\eqref{eq:tuning} represents the standard case of single-objective HP tuning, i.e.\ the optimization is performed with respect to one performance measure. However, HP tuning can also be conducted based on multiple performance measures or additional criteria such as model simplicity \citep{Bischl2023hpo, Dunias2024}. 
Since such multi-objective HP tuning poses further challenges, we will only consider single-objective tuning in this paper. \\
While there exist different tuning procedures, the general model generation process involving tuning can be described as follows: 
Given a set of $C$ candidate HP configurations  (selected before or during the tuning process), each HP configuration $\hpcand$ ($c = 1,\dots, C$) is evaluated on $\dtrain$ by employing one of the model evaluation procedures introduced in Section~\ref{sec:setting1_eval_all}. Accordingly, $\dtrain$ is split into $\dtraine$ and $\dtest$ (either once or multiple times), which are then used for training ($\dtraine$) and prediction error estimation ($\dtest$). In other words, the model evaluation that is performed once with $\hp = \hpi$ in Setting I to assess the prediction error of the final prediction model is performed multiple times for each candidate configuration (i.e.\ with $\hp = \hpcand$) in the tuning process of Setting II. After having evaluated all candidate HP configurations, the HP configuration with the lowest (i.e.\ best) prediction error estimate is used as the final HP configuration. Following the notation introduced in Section~\ref{sec:prelim_hp_sel}, we refer to this configuration as $\hpii$. Note that $\hpii$ is also commonly denoted as $\hat{\hp}$, since it is an estimate of $\hpstar$ \citep{Bischl2023hpo}. However, we adhere to $\hpii$ to clearly distinguish it from Setting I, where $\hp = \hpi$. After setting $\hp = \hpii$, the learning pipeline $\lpii$ undergoes a final training on  $\dtrain$, which yields the final prediction model $\fdliitrain$.\\
Note that while the tuning process already results in a prediction error estimate for the final prediction model (the estimate based on which $\hpii$  was selected during tuning), this value is not necessarily adopted as the final model evaluation result, as we will discuss in Section~\ref{sec:setting2_eval}. In fact, it is also possible to use different performance measures for the prediction error estimation performed during tuning and the evaluation of the final model, but, for the sake of simplicity, we will assume that they are the same.\\
To summarize, during the model generation in Setting II, both the HPs $\hp$ and the parameters $\bs{\theta}$ of the final prediction model are optimized using the data set $\dtrain$. However, the optimization is not performed jointly: first, the HPs $\hp$ are optimized in the tuning process. Second, the parameters $\bs{\theta}$ are optimized in one (final) training process. Note that HPs are still an input of the learning pipeline but can be seen as an output of the tuning process.\\
If only a subset of the HPs $\hp$  are to be tuned, the tuning process described above is applied exclusively to those HPs, while the pre-specified HPs remain fixed throughout the process.
For example, assume that from all $\hpindlarge$ HPs in $\hp$, the HPs $\hp_{1:\hpindsmall} = \lambda_1, .. \lambda_\hpindsmall$  are pre-specified and the HPs $\hp_{\hpindsmall+1:\hpindlarge} = \lambda_{\hpindsmall+1},\dots,\lambda_\hpindlarge$ are to be tuned. 
In this case, the tuning process yields a HP configuration  $\hp_{\hpindsmall+1:\hpindlarge}^\mathrm{II}$, and the final prediction model is trained with $\hp_{1:\hpindsmall} = \hp_{1:\hpindsmall}^\mathrm{I}$ and $\hp_{\hpindsmall+1:\hpindlarge}=\hp_{\hpindsmall+1:\hpindlarge}^\mathrm{II}$. Since the tuning process is conceptually the same when not all HPs are optimized---untuned HPs are simply kept fixed---we will continue to assume that all HPs are tuned to maintain notational simplicity.\\
When choosing a tuning procedure, it is important to consider that the tuning process is limited in terms of both data availability and computation time: 
First, as outlined above, each candidate HP configuration, $\hpcand$, is evaluated using one of the evaluation procedures described in Section~\ref{sec:setting1_eval_all} for Setting I. As explained there, the specified $\dtraine$ and $\dtest$ subsets contain a limited number of observations (i.e.\ $\ntraine$ and $\ntest$ $\leq \ntrain$) and could overlap, potentially leading to unreliable prediction error estimates for each $\hpcand$. Second, the computational budget available for the tuning process is typically limited, which restricts both the number of evaluated HP configurations and the time spent evaluating each configuration (i.e.\ estimating its prediction error). Due to these limitations and the resulting trade-offs (discussed in more detail in Section~\ref{sec:setting2_gen_procedures}), choosing an adequate tuning procedure is often non-trivial. Yet, guidance is still lacking, and many of the existing recommendations are based on rules of thumb rather than empirical benchmarks (see \citealp{Bischl2023hpo}, for an overview). Inadequate tuning procedures can result in a $\hpii$ that yields a final prediction model with worse prediction error than $\hpstar$ (potentially even worse than setting all HPs to their default values) and/or an overly time-consuming tuning process (i.e.\ a more efficient tuning procedure could have achieved the same prediction error in less time).

\subsubsection{Automated vs.\ manual tuning}\label{sec:setting2_gen_am}
Before describing different tuning procedures in more detail, we note that their specification generally depends on whether the tuning process is fully automated or performed manually. We consider the tuning process as automated if the relevant tuning components only need to be specified as a function argument, which is possible in several ML software frameworks (see \citealp{Bischl2023hpo}, for an overview). In contrast, we refer to the tuning process as manual if the candidate HP configurations are evaluated by repeatedly calling the same function(s), altering only the argument that specifies the HP configuration. \\
Compared to automated tuning, manual tuning is more time-consuming, error-prone, and less reproducible, as it is usually an informal and unsystematic process. On the other hand, automated tuning is usually more difficult to implement and requires more programming expertise than manual tuning. As a consequence, although manual tuning is generally advised against \citep[e.g.,][]{Bartz2023, Bischl2023hpo}, it is likely still a common yet often unreported approach in many ML applications \citep{Hosseini2020,Hofman2023,Lones2024}. Note that this may be particularly true for the tuning of preprocessing HPs $\hpp$: As discussed in Section~\ref{sec:prelim_hp_preproc}, preprocessing HPs are often not identified as HPs. Consequently, users trying out different preprocessing options might not be aware that this corresponds to (manual) HP tuning and could be automated. Moreover, if the HPs to be tuned include application-specific preprocessing HPs, the barrier to using automated tuning is further increased, as these HPs may not yet be integrated into the corresponding software and require custom implementation. \\
As a consequence, given the potentially different characteristics of the tuned HPs (especially preprocessing HPs $\hpp$ vs.\ algorithm HPs $\hpa$), we cannot rule out that in practice, they are selected by a combination of automated and manual tuning (see Section~\ref{sec:setup_procedures} for a concrete example). 

\subsubsection{Tuning procedures}\label{sec:setting2_gen_procedures}
As stated above, the selected tuning procedure will affect both the duration of the tuning process and the prediction error of the final prediction model. In the following, we will review the individual components that characterize each tuning procedure and describe how they impact the tuning process. 

\paragraph{Search space} When tuning an HP $\lambda_\hpindsmall$, it is often not reasonable to consider all possible HP values (i.e.\ all values in $\Lambda_\hpindsmall$). For example, this applies if certain values of $\lambda_\hpindsmall$ are already known to cause overfitting or convergence issues. Moreover, when $\lambda_\hpindsmall$ is a preprocessing HP, $\Lambda_\hpindsmall$ may not even be formally specified (see Section~\ref{sec:prelim_hp_preproc}). 
To perform HP tuning, it is thus essential to specify a search space
$\searchspacej$ for each HP, where $\searchspacej$ is a bounded subset of $\Lambda_\hpindsmall$ and determines the HP values that are considered for tuning \citep{Bischl2023hpo}. 
For example, if the HPs of the CART algorithm, $\lambda_{cp}$ and $\lambda_{minsplit}$ with $\Lambda_{cp} = [0,1]$ and $\Lambda_{minbucket} = \{1,\dots,\ntrain\}$, are tuned, their search spaces could be defined as $\tilde{\Lambda}_{cp} = [0.001, 0.1]$ and $\tilde{\Lambda}_{minbucket} =  \{5,\dots,25\}$. The (overall) search space of all $J$ HPs is denoted as $\searchspace = \tilde{\Lambda}_{1}\times\dots\times\tilde{\Lambda}_{J}$.\\
It is important to consider that defining a search space $\searchspace$ restricts the tuning process to finding the optimal HP configuration within $\searchspace$, denoted as $\hpstartilde$, and not within $\bs{\Lambda}$, i.e.\ $\hpstar$. Given a search space $\searchspace$, the tuning problem specified in equation~\eqref{eq:tuning} thus updates to  
\begin{equation}\label{eq:tuning_searchspace}
    \hpstartilde =  \underset{\hp \in \searchspace}{\mathrm{argmin}}   \,\mathrm{PE}(\fdlhptrain).
\end{equation}
Choosing a search space involves the following trade-off: If the search space is too small, the prediction error achieved by $\hpstartilde$ and $\hpstar$ may differ greatly. 
On the other hand, if the search space is too large, this decreases the chance of finding $\hpstartilde$ (or a HP configuration that leads to a comparable prediction error) within a given computational budget \citep{Bischl2023hpo}. \\
Note that in contrast to automated tuning, the search space is usually not formally specified when performing manual tuning and may be extended during the tuning process (e.g., when the user initially planned to try two preprocessing options but then comes up with an additional option during tuning).

\paragraph{Termination criterion} Unless the specified search space $\searchspace$ is very small, such as when only a few categorical HPs are tuned, evaluating all HP configurations in the search space can be computationally challenging or even infeasible. For example, even if $\lambda_{cp}$ and $\lambda_{minbucket}$ are the only HPs being tuned, with the search spaces as specified above and $\tilde{\Lambda}_{cp}$ being searched in increments of 0.001, $C = 100 \times 21 = 2{,}100$ candidate HP configurations would need to be evaluated. Accordingly, one or several criteria must be specified to terminate the tuning process once it is met. 
The trade-off to consider when choosing a termination criterion is that the tuning process should neither stop before finding $\hpstartilde$ nor should it continue longer than necessary, which would result in an inefficient use of resources and, as we will discuss below, increase the risk of overtuning \citep{Bischl2023hpo}. \\
In automated tuning procedures, commonly used criteria are based on the number of evaluations or the runtime. However, additional criteria such as reaching a certain performance level or stagnation of performance might also be reasonable \citep{Bartz2023, Bischl2023hpo}. 
Similar termination criteria, though often more intuitive than formally specified, may also exist for manual tuning when, for example, the user stops searching when satisfied by the reached performance level or gives up searching after a certain amount of time.

\paragraph{Search strategy} Since, in many cases, only a subset of all HP configurations in the search space can be evaluated before the tuning process is terminated, the way in which the sequence of evaluations is determined, also called search strategy or HPO algorithm \citep{Elsken2019, Bischl2023hpo}, is another important component of the tuning procedure. 
Search strategies can be characterized by several aspects, such as the amount of time they spend inferring new candidate HP configurations from already evaluated ones (known as the inference vs.\ search trade-off; \citealp{Bischl2023hpo}). 
For example, search strategies such as evolutionary algorithms and Bayesian optimization consider the distribution and results of previously evaluated HP configurations to propose new configurations. In contrast, the commonly used random search strategy simply draws HP configurations from a predefined, typically uniform, distribution without taking into account past evaluations (see, e.g., \citealp{Feurer2019}, \citealp{Bischl2023hpo}, or \citealp{Bartz2023}, for more details and other search strategies).
In the special case where only the set of selected features is tuned, a well-known automated search strategy is backward or forward feature selection (see, e.g., \citealp{Hastie2009}).\\
Note that the described search strategies are formally used only in automated tuning, as there is usually no specified search strategy when tuning is conducted manually. However, the results of previous evaluations may still be considered in manual tuning when selecting new HP configurations to evaluate.

\paragraph{Joint vs.\ sequential tuning} In automated tuning procedures, all HPs are usually tuned jointly, i.e.\ each evaluated HP configuration potentially considers different values of each HP. However, the HPs could also be tuned sequentially, i.e.\ the complete tuning procedure is repeated for each HP \citep{Probst2019, Waldron2011}. For example, in a setting with three HPs (i.e.\ $\hp= (\lambda_1, \lambda_2, \lambda_3)$), $\lambda_1$ would be tuned first with $\lambda_2$ and $\lambda_3$ set to default, which yields $\lambda_1^{\mathrm{II}}$. Then, $\lambda_2$ is tuned with $\lambda_1 = \lambda_1^{\mathrm{II}}$ and $\lambda_3$ set to its default. Finally, $\lambda_3$ is tuned with $\lambda_1 = \lambda_1^{\mathrm{II}}$ and $\lambda_2 = \lambda_2^{\mathrm{II}}$, yielding $\lambda_3^{\mathrm{II}}$. 
As sequential tuning does not consider any interaction effects between the HPs, it is generally less likely to yield a $\hpii$ comparable to $\hpstartilde$ than joint tuning. 
On the other hand, sequential tuning demands less time, with the maximum number of evaluations increasing linearly rather than exponentially with the number of HPs to tune, as is the case with joint tuning. Hence, it could be a realistic approach for manual tuning.

\paragraph{Prediction error estimation} As outlined above, the prediction error of each HP configuration considered for tuning can be estimated using one of the evaluation procedures described in Section~\ref{sec:setting1_eval_all}. 
In principle, all issues discussed there also apply to the tuning context. However, instead of leading to potentially invalid performance claims about the final prediction model (which was the case in Section~\ref{sec:setting1_eval_all}), using an inadequate evaluation procedure for HP tuning initially only increases the risk of failing to select a $\hpii$ with a (true) prediction error that is comparable to the prediction error of $\hpstartilde$. In other words, if the prediction error of each candidate HP configuration is not estimated adequately, this will initially only affect the model generation process, but not (yet) the evaluation of the final prediction model. Still, the consequences can be detrimental.\\
For example, if each HP configuration is evaluated based on its apparent error (i.e.\ for each $\hpcand$, a model is trained and evaluated on  $\dtrain$, which also serves as $\dtest$), the tuning procedure will, due to the optimistically biased prediction error estimation, typically select the HP configuration that results in the model with the highest degree of overfitting. Although this approach should clearly be avoided, it might still be common practice in manual tuning as it is time-efficient (only one model per HP configuration needs to be trained, which in this case also corresponds to the final model) and may seem intuitive to inexperienced users. \\
Due to the optimistic bias of the apparent error, the standard approach for automated HP tuning is to employ a resampling method. 
In the case of $k$-fold CV, which is a common choice for HP tuning \citep{Bischl2023hpo}, this means that for each candidate HP configuration $\hpcand$, $k$ models are trained and evaluated on different subsets of $\dtrain$.\\ 
While resampling methods provide an improvement over using the apparent error, the corresponding estimators also exhibit a certain degree of pessimistic bias and variance (with the degree of bias and variance depending on the resampling method used, as discussed in Section~\ref{sec:setting1_eval_all}). A potential pitfall arising from the variance is that the winning HP configuration, $\hpii$, may have been selected simply because the trained prediction model(s) using $\hpii$ performed particularly well by chance on the specified test data set(s) $\dtest$, which are the same for each evaluated HP configuration. This means that the HP selection has essentially been overfitted to the respective test data set(s) $\dtest$, which in this context is also referred to as overtuning, overhyping, or oversearching \citep{Hosseini2020, Quinlan1995,Ng1997,Feurer2019,Bischl2023hpo,Cawley2010}. If the true prediction error of $\hpii$ is still comparable to the prediction error of $\hpstartilde$, overtuning effects are negligible. However, there might also be scenarios in which the \textit{true} prediction error of $\hpii$ is no better, or even worse, than that of the default HP configuration, but its \textit{estimated} prediction error is drastically deflated (i.e.\ over-optimistic), as the corresponding prediction model(s) that were trained during resampling incidentally fit very well to the specific noise pattern in the respective test data set(s) $\dtest$. This has been demonstrated in several experiments where tuning was conducted on null data (i.e.\ data without any true signal), yet the prediction error estimate of the selected HP configuration $\hpii$ was substantially smaller (i.e.\ better) than its true prediction error indicating random prediction \citep{Hosseini2020,Varma2006,Boulesteix2009,Bischl2023hpo}. \\
Note that since the HPs are overfitted to the test data set(s) $\dtest$, which are not seen during training on the corresponding $\dtraine$, overtuning occurs on a higher level than overfitting of the model parameters (see Section~\ref{sec:setting1_gen}). Accordingly, overtuning effects may only be visible after evaluating a large number of HP configurations \citep{Bischl2023hpo}. However, literature suggests that the risk of overtuning does not only depend on the number of evaluated HP configurations but also, for example, on the search strategy, the type of tuned HP, and the number of observations in $\dtrain$ \citep{Hosseini2020, Wainer2021,Cawley2010}. In general, overtuning is considered an open problem of HP tuning, and although strategies have been suggested to avoid it (e.g., using different splits for each evaluation, \citealp{Nagler2024}), there are no commonly agreed-upon solutions \citep{Feurer2019}.\\
Importantly, when overtuning is addressed in the literature, it is typically assumed that the prediction error estimation is performed through resampling methods. However, as discussed above, this estimation can alternatively be based on the apparent error. In cases where an inadequate HP configuration is selected due to the use of the apparent error for prediction error estimation, this can be considered a more extreme and direct form of overtuning since the test data set(s) $\dtest$ are seen during model training. We will refer to the two types of overtuning as resampling-induced and apparent error-induced overtuning.

\subsection{Model evaluation} \label{sec:setting2_eval}
\begin{figure}
    \centering
    \includegraphics[width=0.99\linewidth]{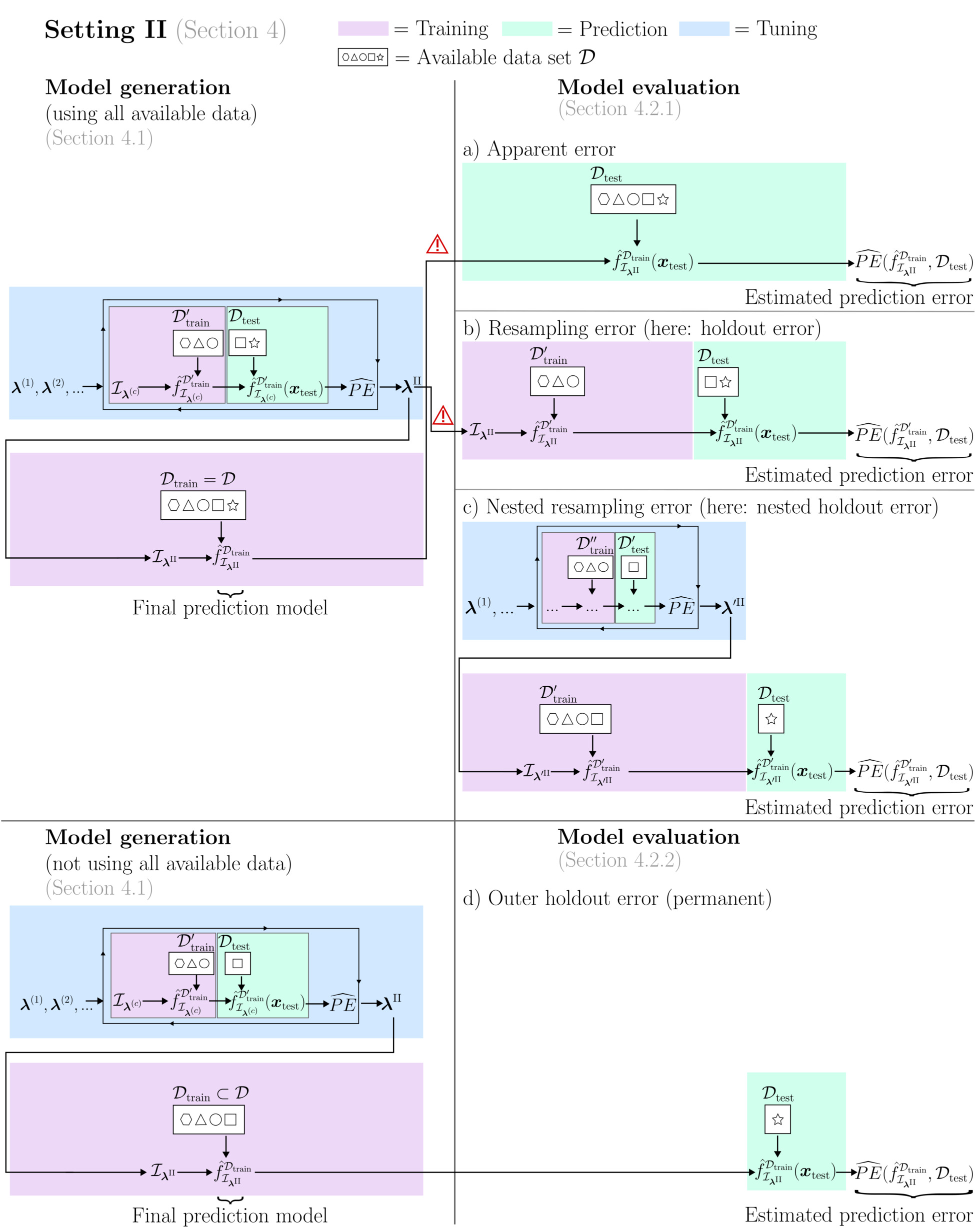}
    \caption{Overview of different model evaluation procedures and their relation to the model generation process if tuning is based on (temporary) holdout and all HPs are tuned.  Data leakage is present if any subset of $\dtest$ used for prediction error estimation has also been employed to generate the evaluated prediction model (which is not necessarily the final model). In the figure, the point at which data \enquote{leaks} into the model evaluation is marked by the red caution symbol. }
    \label{fig:setting2}
\end{figure}

As outlined in Section~\ref{sec:setting2_gen_princ}, the model generation process in Setting II results in a final prediction model $\fdliitrain$.
Evaluating this model is generally more complex than evaluating a prediction model with pre-specified HPs (Setting I),  since it must be taken into account that the model generation process involved HP tuning. Similar to Section~\ref{sec:setting1_eval}, we will in the following differentiate between cases in which the model generation (i.e.\ the HP tuning followed by a final training) is performed on the full data set (i.e.\ $\dtrain = \ds$) vs.\ a (proper) subset of the available data (i.e.\ $\dtrain \subset \ds$). A graphical overview of model evaluation in Setting II is provided in Figure~\ref{fig:setting2}.
\subsubsection{Evaluation of a model generated on all available data} \label{sec:setting2_eval_all}
\paragraph{Apparent error} As in Setting I, reporting the apparent error for model evaluation is inappropriate in Setting II 
(see Figure~\ref{fig:setting2}, model evaluation a). In this case, however, the designated test data set $\dtest=\dtrain=\ds$ is even used twice during model generation: first during the HP tuning process and then again during the final training process. Depending on the specific tuning procedure employed, this can introduce an even greater optimistic bias compared to, for example, using default HP values. Although the apparent error is generally not suitable for assessing a model's performance, some users who performed tuning via resampling may mistakenly believe it now reflects a form of resampling error. This was noted by \citet{Neunhoeffer2019}, who also reference a paper that appears to have fallen into this pitfall.

\paragraph{Resampling error} Similar to Setting I, an alternative evaluation procedure in Setting II is to employ a resampling method (see Figure~\ref{fig:setting2}, model evaluation b). In principle, the chosen resampling method is carried out as described in Section~\ref{sec:setting1_eval_all}, except that in each resampling iteration, the model is trained on $\dtraine$ and evaluated on $\dtest$ with $\hp=\hpii$ instead of $\hp = \hpi$. Unfortunately, unlike in Setting I, using resampling methods for model evaluation in Setting II results in data leakage: Although in each resampling iteration, $\dtest$ is not involved in training $\fdliitraine$ (the model trained on $\dtraine$ for evaluation purposes), it is used in the tuning process performed on $\dtrain$ (including $\dtest$) to obtain $\hpii$. Accordingly, since not every model generation step resulting in $\fdliitraine$ is conducted exclusively on $\dtraine$, information from $\dtest$ is available during the model generation process (specifically, during tuning). Based on the definition given in  Section~\ref{sec:prelim_processes_eval}, this constitutes a form of data leakage and may result in an optimistically biased resampling error \citep{Wainer2021, Hosseini2020}. 
While the inadequacy of the apparent error is widely recognized, the described pitfall associated with the resampling error is less well known and will go undetected by those not involved in model development if HP tuning is not reported \citep{Hosseini2020,Lones2024}.\\
The potential optimistic bias becomes evident when considering the following typical practice: As outlined in Section~\ref{sec:setting2_gen_princ}, the tuning process already returns a prediction error estimate for the final prediction model (the estimate based on which $\hpii$ was selected). Given that tuning was performed with a resampling method (e.g., CV), computation time can be saved by directly using this value as the resampling-based evaluation result. However, if the selected HP configuration $\hpii$ is the result of overtuning, this will not be detected in the model evaluation process, as the deflated prediction error estimate is simply adopted here. 
In principle, adopting the resampling prediction error estimate from tuning in Setting II behaves analogously to (resampling-induced) overtuning as using the apparent error does to overfitting in Setting I. This is because both procedures are unable to discern that either the selected HPs (overtuning) or the selected parameters (overfitting) have been adapted too much to the respective test data set(s) $\dtest$. \\
As stated in Section~\ref{sec:setting2_gen_procedures}, the extent to which overtuning occurs depends on the specific tuning procedure. If the HP selection is   mildly overtuned, the prediction error estimate obtained from the tuning process may only exhibit a slight optimistic bias. However, as an extreme case, we can again consider the experiments from Section~\ref{sec:setting2_gen_procedures} in which HP tuning has been performed on null data \citep{Hosseini2020,Varma2006,Boulesteix2009,Bischl2023hpo}. Here, the difference between the prediction error estimate of the selected HP configuration and the true prediction error indicating random prediction is substantial, and adopting the former as the final evaluation result for a useless prediction model is clearly a biased approach.\\
Note that data leakage is also present if the specified $\dtraine$ and $\dtest$ subsets used for tuning and evaluation are not identical. This is the case if additional resampling iterations are conducted during evaluation, if different resampling methods are used during tuning and evaluation (e.g., holdout and $k$-fold CV), or if the apparent error is used for tuning. 

\paragraph{Nested resampling error} The optimistic bias of the resampling error arises because, in each resampling iteration, not all steps of the model generation process are performed exclusively on $\dtraine$. A natural extension, therefore, is to ensure that the complete model generation is applied only to $\dtraine$ in every iteration (see Figure~\ref{fig:setting2}, model evaluation c). Specifically, this implies that the tuning process is not only performed once on $\dtrain$ in order to generate the final prediction model but also on every $\dtraine$ specified during resampling (for evaluation purposes). 
If the tuning process itself is based on a resampling method (i.e.\ if tuning is not performed using the apparent error, which is hardly ever the case if the currently described model evaluation procedure is employed),  this results in two nested resampling methods. Accordingly, this procedure is called nested resampling, where the resampling method that initially splits $\dtrain$ into $\dtraine$ and $\dtest$ is the outer resampling loop and the resampling method creating additional splits within each $\dtraine$ (resulting in subsets denoted as $\dtrainee$ and $\dteste$) is the inner resampling loop \citep[e.g.,][]{Bischl2023hpo,Hosseini2020,Wainer2021}. To distinguish nested resampling from the resampling methods discussed above and in Section~\ref{sec:setting1_eval_all}, we will refer to the latter as simple resampling where necessary.\\
The most straightforward form of nested resampling is the nested holdout method, where $\dtrain$ is split once into $\dtraine$ and $\dtest$, and $\dtraine$ is further divided into $\dtrainee$ and $\dteste$. In this setup, the best HP configuration for $\dtraine$ is determined by training and evaluating a model for each candidate HP configuration on $\dtrainee$ (for training) and $\dteste$ (for prediction error estimation). We denote this configuration as $\hpiie$, as it may differ from the final prediction model's configuration, $\hpii$, which has been obtained by tuning the model on $\dtrain$ rather than $\dtraine$. Using the HP configuration $\hpiie$, the model is then trained on $\dtraine$ and evaluated on $\dtest$, which has remained unseen throughout the entire model generation process. Note that nested holdout is commonly referred to as train-validation-test split  \citep{Bischl2023hpo}, which, using the notation above, could also be referred to as $\dtrainee$-$\dteste$-$\dtest$-split. Instead of holdout, any other resampling method can be used for inner and outer resampling, and it is also possible to combine different resampling methods. For example, $k$-fold CV can be used for outer resampling and holdout for inner resampling, since in the inner resampling, precise prediction error estimation is less critical as long as a sufficiently good $\hpiie$ is selected in each iteration \citep{Bischl2023hpo,Hosseini2020}. \\
While nested resampling prevents data leakage, it also has several disadvantages. First, it can be very computationally expensive, since the tuning process, which can already be time-consuming when conducted once, has to be repeated for each $\dtraine$ specified by the outer resampling loop \citep{Wainer2021, Bischl2023hpo}. 
Second, it is usually not feasible to conduct nested resampling with manual tuning. Apart from being even more time-demanding than nested resampling with automated tuning, it is often not possible to repeat the same tuning procedure more than once due to the informal nature of manual tuning (e.g., the user might not remember which candidate HP configurations have been evaluated during tuning). Third, like simple resampling, nested resampling does not provide an estimate of the prediction error for the final model $\fdliitrain$. However, while both methods evaluate models trained on $\dtraine$ rather than $\dtrain$ (with $\ntraine < \ntrain$), simple resampling at least uses the same HP configuration $\hpii$ as the final prediction model. In contrast, nested resampling does not necessarily evaluate models with the same HP configuration, as each inner resampling loop may select a different configuration (see the nested holdout example above, which evaluates a model based on $\hpiie$ instead of $\hpii$). This makes the nested resampling result more difficult to interpret \citep{Hosseini2020}. 
The described disadvantages could explain why nested resampling estimates are not commonly reported in studies presenting new prediction models, as indicated by a recent systematic review on clinical prediction models \citep{Navarro2023_systemreview}.
 
\subsubsection{Evaluation of a model generated on a subset of the available data}\label{sec:setting2_eval_subset}
As in Setting I (see Section~\ref{sec:setting1_eval_subset}), it is also possible in Setting II to use only a subset of the available data for model generation (i.e.\ $\dtrain \subset \ds$) and reserve the remaining observations exclusively for evaluation (i.e.\ $\dtest = \ds \setminus \dtrain$; see Figure~\ref{fig:setting2}, model evaluation d; \citealp{Hosseini2020}). This approach essentially corresponds to nested resampling with holdout as the outer resampling method, except that the holdout is permanent, meaning that the prediction model generated on $\dtrain$ (equivalent to $\dtraine$ in the previous section) serves as the final prediction model. Similar to Setting I, we thus distinguish the two evaluation procedures by referring to them as temporary outer holdout (described in Section~\ref{sec:setting2_eval_all}) and permanent outer holdout (described here). We also again note that there might be some confusion in the terminology, as a permanent outer holdout combined with a (temporary) inner holdout can, just like its temporary counterpart, also be referred to as a train-validation-test split. \\
The statements regarding the temporary vs.\ permanent holdout in Setting I also apply to Setting II: Compared to the temporary outer holdout, the permanent outer holdout does not exhibit a pessimistic bias as it actually evaluates the final prediction model. However, this comes at the cost of not using all available data for model generation. Accordingly, the same recommendation as in Section~\ref{sec:setting1_eval_subset} applies: a permanent outer holdout should only be employed if the number of observations in $\ds$ is sufficiently large or if it is computationally expensive or practically infeasible to repeat the model generation process. Note that the second point is particularly relevant in Setting II due to the increased effort of model generation \citep{Collins2024}.

\section{Empirical illustration of different model generation and evaluation procedures}\label{sec:illustration}
In this section, we illustrate different procedures for model generation and evaluation and assess their impact on prediction error estimates from available vs. new data. We specifically focus on the selection of HPs and the potential for data leakage.        

\subsection{Real-world prediction problem} \label{sec:companion}
Our illustration is based on a real-world prediction problem from the COMPANION study \citep{Hodiamont2022}. This study aimed to develop a casemix classification for adult palliative care patients in Germany that considers the complexity of each patient’s palliative care situation to assign them to a class reflecting their resource needs.  A casemix classification for palliative care patients has been deemed necessary, as the differentiation of patients based on their diagnosis, which corresponds to the current practice in Germany, has been found to be inappropriate for predicting resource needs in the context of palliative care. Despite yielding many important insights, the COMPANION project was ultimately unable to develop a prediction model with sufficient predictive performance, even after exploring various model generation approaches. However, this makes it a good example to illustrate how optimistically biased evaluation procedures can present prediction models in a more favorable light. \\
To develop a casemix classification that relates patients' resource needs to the complexity of their palliative care situation, the COMPANION team formulated a prediction problem where each observation represents a patient's palliative care phase. The outcome $y^{(i)}$, defined as the average cost per day in palliative care phase $i$, serves as an empirical proxy for resource needs in the corresponding phase. The set of features $\bs{x}^{(i)}$ intended to reflect the palliative care situation of each phase consists of (i) the type of palliative care phase (categorical), (ii) patient age (integer-valued), (iii) two cognitive features (confusion and agitation; both ordinal), (iv) the Australia-modified Karnofsky Performance Status (AKPS; \citealp{Abernethy2005akps}) that measures the patients' functional status (ordinal), and (v) the Integrated Palliative care Outcome Scale (IPOS; \citealp{Murtagh2019ipos}), which is a score that is based on 17 ordinal variables covering physical symptoms, psycho-social burden, family needs, and practical problems. Accordingly, the number of features provided to the learning algorithm is $p=6$. All types of data were collected by the clinical staff of participating palliative care teams.\\
It is important to note that although the study aimed to identify a casemix classification, the continuous nature of the specified outcome variable (i.e.\ average cost per day) inherently makes the prediction problem a regression task. To ensure that the obtained prediction model still produces classes that are also interpretable and can be implemented in practice, a decision tree approach was chosen (e.g., using the CART algorithm, discussed in Sections~\ref{sec:prelim}-\ref{sec:setting2}), despite potential limitations on predictive performance. In the resulting decision tree, each terminal node represents a casemix class (defined by the features that capture the complexity of the palliative care situation) and predicts the average cost per day for that class. Notably, decision trees were also used in the casemix classifications developed for palliative care patients in Australia  \citep{Eagar2004technical} and the UK \citep{Murtagh2023}, which served as the basis for many decisions in the development of the German casemix classification.  \\
The COMPANION study collected data from three palliative care settings (specialist palliative care units, palliative care advisory teams, and specialist palliative home care), with a casemix classification to be developed for each setting. In our illustration, we only consider the data from the specialist palliative home care setting. We apply several parameterless preprocessing steps to the raw data set, which correspond to those used in the COMPANION study and are considered as pre-specified in our illustration (e.g., the removal of dead patients; more details can be found in Supplementary Section~B.2.1). The resulting data set contains 1{,}449 palliative care phases; descriptive statistics are provided in Table~S1.\\
Note that while our experimental setup described in the following section is based on the COMPANION study, not all aspects align with how the actual study was conducted, as some elements have been simplified or modified for illustrative purposes.

\subsection{Experimental setup}\label{sec:setup}
\subsubsection{Overview} \label{sec:setup_overview}
The aim of our study is to illustrate different model generation and evaluation procedures and examine their impact on prediction error estimates derived from available data compared to those obtained from new data.
Additionally, we examine how these estimates are affected by performance measure, sample size, and learning algorithm, resulting in a total of 96 distinct analysis settings. Before providing more details on these, we first outline the general procedure that is carried out for each analysis setting: 
\begin{enumerate}[label=(\roman*)]
\item The COMPANION data set with 1{,}449 observations (i.e.\ palliative care phases) introduced above is randomly split into two subsets of equal size, which we denote as $\dtrain$ and $\dnew$ (with $n_{\text{train}} = 724$ and $n_{\text{new}} = 725$). We assume that $\dtrain$ is the only data set available for both model generation and evaluation. Consistent with the notation used in previous sections, this implies $\dtrain = \ds$. The desired output is a prediction model as described above (i.e.\ a decision tree that predicts the average patient costs based on several features reflecting the palliative care situation). 
\item We use $\dtrain$ exclusively to generate and evaluate a prediction model. Although the specific procedure is determined by the analysis setting, each model is generated using all available data (which is already implied by referring to the available data as $\dtrain$). The learning pipeline used for each training process and its HPs are described in Section~\ref{sec:setup_lp}. Since the HP selection in the considered analysis settings can be either data-independent or achieved through tuning, we refer to the chosen HP configuration as $\hp$ rather than $\hpi$ or $\hpii$ in the following to keep the notation general. Step (ii) results in a model $\fdlhptrain$ and an associated prediction error estimate, which we denote as $\ped$. In an ML application, $\ped$ would be the reported error. 
\item The prediction model $\fdlhptrain$ is evaluated on the second data set $\dnew$, which represents observations that are drawn from the same distribution as the observations in $\dtrain$ but were unseen during the generation of $\fdlhptrain$.  This step should therefore yield an unbiased estimate of the model's prediction error, denoted as $\pednew$ (however, see the note on clustering
in Section~\ref{sec:results} and Supplementary Section~B.5). Note that, in principle, the estimation of $\pednew$ resembles a permanent holdout approach, where $\dnew$ is held out during model generation. However, it is not truly a holdout, as $\dnew$ is unavailable during model evaluation. This is also why $\dnew$ is not referred to as $\dtest$; throughout the paper, the notation $\dtest$ is used exclusively for subsets of the available data.
\end{enumerate}
Performing steps (i) to (iii) results in a vector  $(\ped, \pednew)$, which includes the prediction error estimates derived from available and new data, respectively. By comparing these estimates, we can determine whether $\ped$ correctly reflects the predictive performance of the model or if it is affected by any form of bias. Ideally, $\ped$ should be equal to $\pednew$, indicating that the model evaluation conducted on $\dtrain$ yields an unbiased estimate prediction error estimate (although small differences do not necessarily indicate bias, as $\pednew$ is also an estimate). To ensure that the difference between the two prediction error estimates is not driven by a specific data split, steps (i) to (iii) are repeated 50 times for each analysis setting (using the same 50 splits for each analysis setting). Since we consider 96 analysis settings and 50 repetitions of splitting the initial COMPANION data set, our illustration generates $96 \times 50 = 4{,}800$ vectors of  $(\ped, \pednew)$. Note
that each analysis setting may produce 50 different prediction models, as in each repetition, $\dtrain$ contains different observations. \\
The described setup is implemented in the software environment \texttt{R} \citep{rsoftware} using the \texttt{mlr3} package framework \citep{mlr3_package}. While the COMPANION data set cannot be made publicly available, the \texttt{R} code and the individual prediction error estimates can be found at \url{https://github.com/NiesslC/overoptimistic\_trees}.\\
As stated above, we consider a total of 96 analysis settings. These result from a full factorial variation of four factors: two performance measures, two sample sizes, two learning algorithms, and twelve combinations of model generation and evaluation procedures (yielding the total of $2 \times 2 \times 2 \times 12 = 96$ analysis settings). 
The two considered sample sizes are (i) $n_{\text{train}} = 724$ (the sample size of $\dtrain$ after splitting the original data set) and (ii) $n_{\text{train}} = 362$ (half of the observations in $\dtrain$ being randomly deleted).  Note that $\dnew$ is not affected by this variation and still has $n_{\text{new}} = 725$ observations.
The two performance measures considered in our illustration are the Root Mean Squared Error (RMSE) and the coefficient of determination ($R^2$), which are commonly used performance measures and have also been employed to evaluate other decision-tree-based prediction models for palliative care patients (\citealp{Eagar2004technical,Murtagh2023}; see Supplementary Section~B.3 for more information on both performance measures).  Note that in each analysis setting, we use the same performance measure for both the model evaluations performed during model generation (i.e.\ tuning) and the evaluation of the final prediction model. The two learning algorithms and twelve combinations of model generation and evaluation procedures are described in Sections~\ref{sec:setup_lp} and \ref{sec:setup_procedures}, respectively. 

\subsubsection{Learning pipeline and HPs}\label{sec:setup_lp}
\begin{figure}[!t]
    \centering
    \includegraphics[width=1\linewidth]{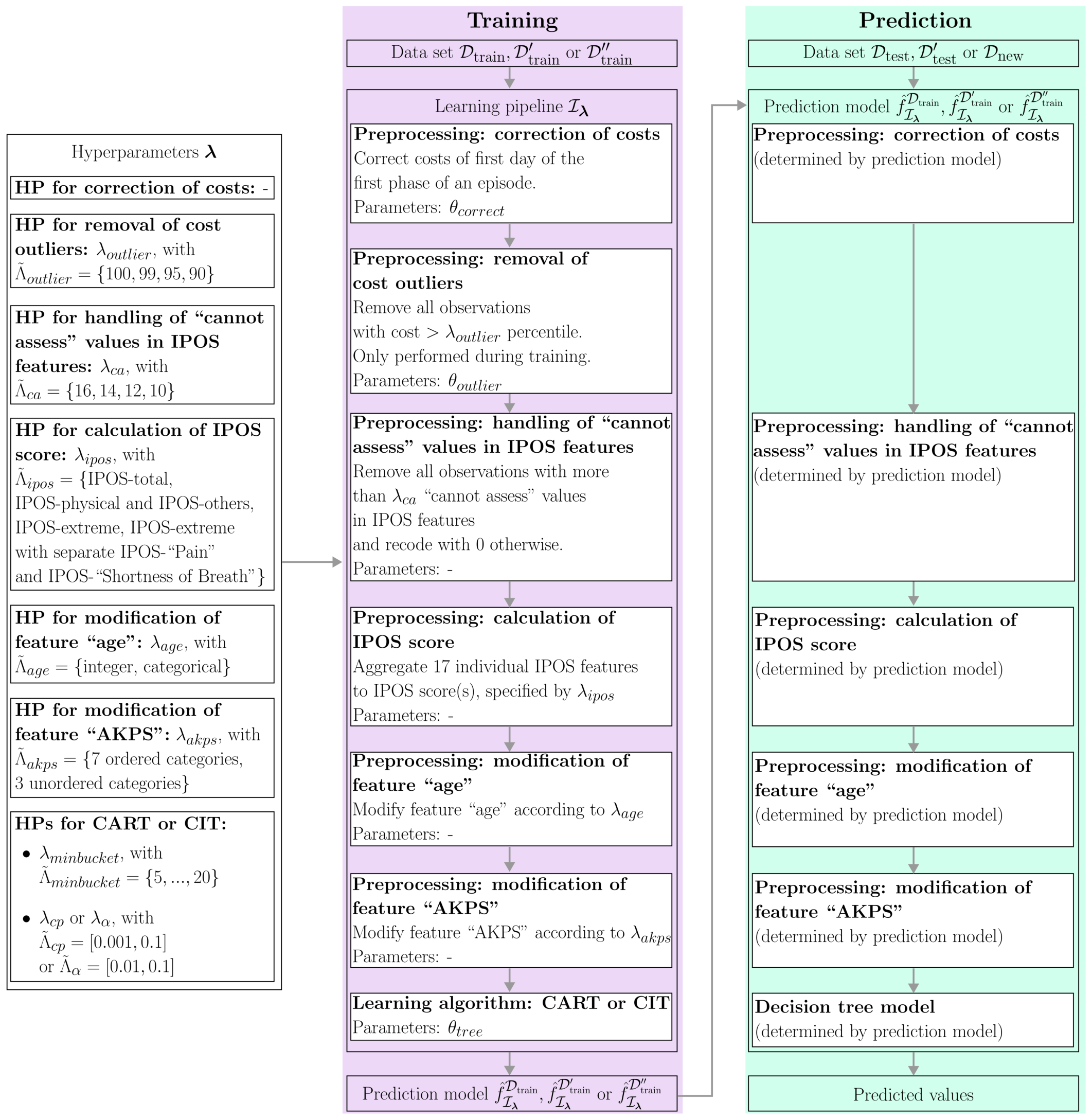}
    \caption{Overview of the learning pipeline $\lp$ used in the illustration (middle panel). In addition, the considered HPs, their search spaces (left panel), and the steps applied during prediction (right panel) are shown.}
    \label{fig:lp_companion}
\end{figure}

The learning pipeline $\lp$ applied in each training process consists of six preprocessing steps, followed by a learning algorithm (see Figure~\ref{fig:lp_companion} for an overview). While the full learning pipeline actually consists of more preprocessing steps (referred to in Section~\ref{sec:companion} and detailed in Supplementary Section~B.2.1), we will, for simplicity, not further consider them in the illustration, as they are considered as pre-specified (i.e.\ have no HPs that are relevant for tuning) and  are both parameterless and precede the first parameterized preprocessing step in the learning pipeline (i.e.\ can safely be applied to the full data set).

\paragraph{Preprocessing steps} Here, we provide a brief overview of the six preprocessing steps in $\lp$ applied during each training process and outline their associated HPs. 
Additional details can be found in Figure~\ref{fig:lp_companion}, and a comprehensive description is available in Supplementary Section~B.2.2. \\
The six preprocessing steps serve one of three purposes: (i) correction of the outcome variable (correction of costs), (ii) handling of problematic observations (removal of cost outliers and handling of \enquote{cannot assess} values in IPOS features), and (iii) calculation or modification of features (calculation of the IPOS score, modification of the feature \enquote{age}, and modification of the feature \enquote{AKPS}). As discussed in Section~\ref{sec:prelim_lp_preproc}, preprocessing steps can be distinguished based on different characteristics, which also applies to the six preprocessing steps considered in this section.
Two of the six steps have parameters: the correction of costs (with $\theta_{correct}$) and the removal of cost outliers (with $\theta_{outlier}$). These two steps, along with another step (handling of \enquote{cannot assess} values in IPOS features), alter the outcome distribution, but the removal of cost outliers is not applied during prediction.  \\
All preprocessing steps, except for the correction of costs, include HPs: $\hpoutlier$, $\hpca$, $\hpipos$, $\hpage$, and $\hpakps$. Consistent with the notation introduced in Section~\ref{sec:prelim_hp_notation}, we collectively refer to them as $\hpp$. For these HPs, it is not possible to define a HP domain $\Lambda_\hpindsmall$ that contains all possible configurations; therefore, we only specify a search space $\tilde{\Lambda}_\hpindsmall$ for each HP (see Figure~\ref{fig:lp_companion}). Each search space is categorical, offering 2 or 4 values, all of which have been discussed and deemed reasonable during the COMPANION project. 
The first HP value in each search space is set as the default and corresponds to the value ultimately selected for the COMPANION project.

\paragraph{Learning algorithm} After applying all preprocessing steps to the data, it is provided to the learning algorithm, which then yields a prediction model (i.e.\ a decision tree). We consider two learning algorithms: (i) the CART algorithm (introduced in Section~\ref{sec:prelim_lp_algo}; \texttt{R} package \texttt{rpart}; \citealp{rpart_package}), and (ii) the Conditional Inference Tree algorithm (CIT; \texttt{R} package \texttt{partykit}; \citealp{partykit_package1,partykit_package2,partykit_package3}). As stated in Sections~\ref{sec:prelim_lp_algo} and \ref{sec:setting1_gen}, the CART algorithm builds a decision tree model by partitioning the feature space $\mathcal{X}$ into terminal nodes using a sequence of binary splits. Since we are considering a regression problem, the splitting rules are determined by minimizing the sum of squared errors, and the prediction value $\hat{f}(\bs{x})$  for each terminal node is the mean of all outcome values (here: costs) in that node \citep{Breiman1984}. The CIT algorithm also employs recursive binary partitioning, but instead of minimizing a simple loss function that represents node impurity (here: the sum of squared errors), it uses statistical test procedures to find the optimal splits. This approach has the advantage that, unlike the CART algorithm, the CIT algorithm is not affected by selection bias toward features with many possible splits or missing values \citep{partykit_package2}.\\
For both algorithms, we consider two HPs for tuning that determine when the algorithm stops splitting.  The first HP is $\hpminb$, which specifies the minimum number of observations in any terminal node. The smaller $\hpminb$, the larger the number of terminal nodes in the resulting decision tree and the higher the risk of overfitting. We set the search space of $\hpminb$ to $\{{5,\ldots,20}\}$ for tuning. If $\hpminb$ is not tuned, we set the HP to its default, $\hpminb$ = 7. The second HP is either $\hpcp$ (for CART) or $\hpalpha$ (for CIT). Both HPs serve a similar purpose: $\hpcp$ determines the factor by which a split must improve the overall lack of fit to be attempted (which, in case of a regression problem, corresponds to improving the overall $R^2$ of the model by at least $\hpcp$). The HP $\hpalpha$ is the numerical significance level that must be met in the statistical testing procedure conducted by CIT to implement a split. Accordingly, the smaller $\hpcp$ or the higher $\hpalpha$, the higher the risk of overfitting. We specify the search space for $\hpcp$ and $\hpalpha$ as $[0.001,0.1]$ and $[0.01, 0.1]$, respectively. If $\hpcp$ and $\hpalpha$ are not tuned, we use their default values of $\lambda_{cp} = 0.01$ and $\lambda_{\alpha} = 0.05$.\\
All other HPs of CART and CIT are not tuned and, except for one HP, follow the default values from their corresponding implementation in the \texttt{mlr3} package \citep{mlr3_package}, which largely align with the defaults of the underlying packages (i.e.\ \texttt{rpart} and \texttt{partykit}; \citealp{mlr3book_2databasic}). The exception is $\lambda_{maxdepth}$, which we set to 4 to align with the COMPANION project, where this value was chosen to ensure that the resulting decision tree model would be useful in clinical practice.\\
We refer to the algorithm HPs that are considered for tuning (i.e.\ $\hpminb$ and $\hpcp$ or $\hpalpha$) as $\hpa$. The remaining algorithm HPs that are not tuned in any of the analysis settings will not be considered further for simplicity.

\subsubsection{Model generation and evaluation procedures}\label{sec:setup_procedures}
We consider twelve different combinations of model generation and evaluation procedures that could be employed in step (ii) of our illustration (see Section~\ref{sec:setup_overview}) to obtain a prediction model with associated $\ped$. They represent an exemplary yet non-exhaustive selection of procedures that are used in ML applications. The twelve combinations are based on five model generation procedures, where for three of them, we apply two different procedures to evaluate the final prediction model, and for the other two, we use three different evaluation procedures (resulting in a total of $3 \times 2 + 2 \times 3 = 12$ combinations). \\
Before describing the procedures in more detail, there are a few general points to consider. First, as already stated in Section~\ref{sec:setup_overview}, all model generation procedures use the full data set $\dtrain$ that was created by the respective repetition, i.e.\ we do not consider the permanent holdout evaluation procedures introduced in Sections~\ref{sec:setting1_eval_subset} and \ref{sec:setting2_eval_subset} (which would imply $\dtrain \subset \ds$). Second, since the prediction model used in this illustration is a decision tree, it is theoretically possible to manually assess the plausibility of the generated models in addition to estimating their prediction error. However, in addition to not being feasible for all $96 \times 50$ generated models, this step is also often not part of the evaluation process in practice, as many ML-based prediction models are not interpretable by humans without additional tools. Therefore, we do not perform this assessment. Third, whenever $\dtrain$ is (temporarily) split as part of a resampling method (either during model generation or evaluation), we use the same splits (e.g., the same 10 CV folds) across all procedures to ensure that differences in prediction error estimates are not due to variations in the data splits of $\dtrain$. \\
We now present the procedures in more detail, first describing the model generation procedure and then the associated evaluation procedures to estimate the prediction error of the resulting model.  The following paragraph titles refer to the model generation procedures and can be read as \enquote{Setting - Tuning Procedure (- HPs tuned)}. An overview of all generation and evaluation procedures is provided in Table~\ref{tab:procedures}. 

\begin{sidewaystable}
\caption{Overview of the twelve combinations of model generation and evaluation procedures examined in the illustration. They result from five model generation procedures, each paired with two or three evaluation procedures.}
\label{tab:procedures}
\resizebox{\textwidth}{!}{%
\begin{tabular}{lllllllllll}
\hline
\multicolumn{1}{c}{\multirow{5}{*}{\textbf{Setting}}} &
  \multicolumn{8}{c}{\textbf{Model generation on $\dtrain$}} &
  \multicolumn{2}{c}{\textbf{Model evaluation on $\dtrain$}} \\ \cline{2-11} 
\multicolumn{1}{c}{} &
  \multirow{4}{*}{\textbf{\begin{tabular}[c]{@{}l@{}}Model\\ generation\\ name\end{tabular}}} &
  \multirow{4}{*}{\textbf{\begin{tabular}[c]{@{}l@{}}Pre-\\ specified\\ HPs\end{tabular}}} &
  \multirow{4}{*}{\textbf{\begin{tabular}[c]{@{}l@{}}Tuned\\ HPs\end{tabular}}} &
  \multicolumn{5}{c}{\textbf{Tuning procedure}} &
  \multirow{4}{*}{\textbf{\begin{tabular}[c]{@{}l@{}}Prediction\\ error\\ estimation\end{tabular}}} &
  \multirow{4}{*}{\textbf{\begin{tabular}[c]{@{}l@{}}Data\\ leakage\\ possible\end{tabular}}} \\ \cline{5-9}
\multicolumn{1}{c}{} &
   &
   &
   &
  \textbf{\begin{tabular}[c]{@{}l@{}}Search\\ space\end{tabular}} &
  \textbf{\begin{tabular}[c]{@{}l@{}}Termination\\ criterion\end{tabular}} &
  \textbf{\begin{tabular}[c]{@{}l@{}}Search\\ strategy\end{tabular}} &
  \textbf{\begin{tabular}[c]{@{}l@{}}Joint vs.\\ sequential\\ tuning\end{tabular}} &
  \textbf{\begin{tabular}[c]{@{}l@{}}Prediction\\ error\\ estimation\end{tabular}} &
   &
   \\ \hline
\multirow{2}{*}{I} &
  \multirow{2}{*}{I-no tuning} &
  \multirow{2}{*}{\begin{tabular}[c]{@{}l@{}}$\hpp$,\\ $\hpa$\end{tabular}} &
  \multirow{2}{*}{-} &
  \multirow{2}{*}{-} &
  \multirow{2}{*}{-} &
  \multirow{2}{*}{-} &
  \multirow{2}{*}{-} &
  \multirow{2}{*}{-} &
  Apparent &
  Yes \\ \cline{10-11} 
 &
   &
   &
   &
   &
   &
   &
   &
   &
  10-fold CV &
  No \\ \hline
\multirow{2}{*}{II} &
  \multirow{2}{*}{II-manual-P} &
  \multirow{2}{*}{$\hpa$} &
  \multirow{2}{*}{$\hpp$} &
  \multirow{2}{*}{\begin{tabular}[c]{@{}l@{}}See\\ Figure ~\ref{fig:lp_companion}\end{tabular}} &
  \multirow{2}{*}{None} &
  \multirow{2}{*}{\begin{tabular}[c]{@{}l@{}}Exhaustive\\ search\end{tabular}} &
  \multirow{2}{*}{Sequential} &
  \multirow{2}{*}{Apparent} &
  Apparent &
  Yes \\ \cline{10-11} 
 &
   &
   &
   &
   &
   &
   &
   &
   &
  10-fold CV &
  Yes \\ \hline
\multirow{4}{*}{II} &
  \multirow{4}{*}{II-automated-A} &
  \multirow{4}{*}{$\hpp$} &
  \multirow{4}{*}{$\hpa$} &
  \multirow{4}{*}{\begin{tabular}[c]{@{}l@{}}See\\ Figure ~\ref{fig:lp_companion}\end{tabular}} &
  \multirow{4}{*}{\begin{tabular}[c]{@{}l@{}}60\\ evaluations\end{tabular}} &
  \multirow{4}{*}{\begin{tabular}[c]{@{}l@{}}Random\\ search\end{tabular}} &
  \multirow{4}{*}{Joint} &
  \multirow{4}{*}{10-fold CV} &
  Apparent &
  Yes \\ \cline{10-11} 
 &
   &
   &
   &
   &
   &
   &
   &
   &
  10-fold CV &
  Yes \\ \cline{10-11} 
 &
   &
   &
   &
   &
   &
   &
   &
   &
  \begin{tabular}[c]{@{}l@{}}10-2-fold\\ nested CV\end{tabular} &
  No \\ \hline
\multirow{2}{*}{II} &
  \multirow{2}{*}{II-combined-PA} &
  \multirow{2}{*}{-} &
  \multirow{2}{*}{\begin{tabular}[c]{@{}l@{}}$\hpp$,\\ $\hpa$\end{tabular}} &
  \multicolumn{5}{c}{\multirow{2}{*}{\begin{tabular}[c]{@{}c@{}}II-manual-P for $\hpp$ and II-automated-A for $\hpa$\\ (for each configuration of $\hpp$)\end{tabular}}} &
  Apparent &
  Yes \\ \cline{10-11} 
 &
   &
   &
   &
  \multicolumn{5}{c}{} &
  10-fold CV &
  Yes \\ \hline
\multirow{4}{*}{II} &
  \multirow{4}{*}{II-automated-PA} &
  \multirow{4}{*}{-} &
  \multirow{4}{*}{\begin{tabular}[c]{@{}l@{}}$\hpp$,\\ $\hpa$\end{tabular}} &
  \multirow{4}{*}{\begin{tabular}[c]{@{}l@{}}See\\ Figure ~\ref{fig:lp_companion}\end{tabular}} &
  \multirow{4}{*}{\begin{tabular}[c]{@{}l@{}}210\\ evaluations\end{tabular}} &
  \multirow{4}{*}{\begin{tabular}[c]{@{}l@{}}Random\\ search\end{tabular}} &
  \multirow{4}{*}{Joint} &
  \multirow{4}{*}{10-fold CV} &
  Apparent &
  Yes \\ \cline{10-11} 
 &
   &
   &
   &
   &
   &
   &
   &
   &
  10-fold CV &
  Yes \\ \cline{10-11} 
 &
   &
   &
   &
   &
   &
   &
   &
   &
  \begin{tabular}[c]{@{}l@{}}10-2-fold\\ nested CV\end{tabular} &
  No \\ \hline
\end{tabular}%
}
%{\raggedright (*) overlap-induced \par}
\end{sidewaystable}

\paragraph{\pdefault}
The simplest model generation procedure corresponds to Setting I, where all HPs are set to their default values (i.e.\ no tuning is performed), and the learning pipeline only needs to be trained once on the data set $\dtrain$. \\
For this model generation procedure, we evaluate the resulting model by (i) the apparent error and (ii) the 10-fold CV error. The former is affected by data leakage and may thus exhibit a substantial optimistic bias (see Section~\ref{sec:setting1_eval_all}).

\paragraph{\pM}
In this model generation procedure, the preprocessing HPs ($\hpp$) are tuned, while the algorithm HPs ($\hpa$) are set to their default values. It aims to represent inexperienced users who either lack the confidence or the programming skills to tune algorithm HPs but manually experiment with different preprocessing options, without realizing that this is a form of HP tuning. As discussed in Sections~\ref{sec:setting2_gen_am} and \ref{sec:setting2_gen_procedures}, manual tuning procedures typically differ from automated tuning procedures, which is reflected by the procedure {\pM}. First, the HPs are tuned sequentially (i.e.\ each HP is tuned individually, with previously tuned HPs set to their selected values and subsequently tuned HPs set to their default values).
Second, during the tuning of each HP, the apparent error is used to estimate the prediction error of each candidate HP configuration. The order in which the HPs are tuned sequentially is $\hpipos$, $\hpage$, $\hpakps$, $\hpoutlier$, $\hpca$ (which reflects a user who first experiments with variations in the features before removing observations, though any other order is also possible). If more than one HP value yields the same prediction error estimate, the first value that was evaluated is selected. Since the preprocessing HPs are tuned sequentially (i.e.\ one at a time), and only two ($\hpage$, $\hpakps$) or four ($\hpipos$, $\hpoutlier$, $\hpca$) values per HP are available, only 16 ($=2\times2+4\times3$) configurations of $\hpp$ need to be evaluated during tuning. Therefore, no criterion is specified to terminate tuning before all configurations are evaluated.\\
Similar to the first model generation procedure (\pdefault), we consider the apparent error and the 10-fold CV error to evaluate the final prediction model. However, the 10-fold CV error is now affected by data leakage, potentially leading to an optimistic bias due to (apparent error-induced) overtuning (see Section~\ref{sec:setting2_eval_all}). Note that we do not consider evaluation procedures involving nested resampling for {\pM}, as this is typically not feasible if manual tuning was used for model generation (see Section~\ref{sec:setting2_eval_all}).

\paragraph{\pA}
This model generation procedure represents a standard procedure in many ML applications, where the algorithm HPs $\hpa$ are selected through automated tuning, while the preprocessing HPs $\hpp$ are set to their default values (e.g., because users are not aware that they can be tuned). Even when tuning is fully automated, the procedures used in practice are often simple and based on rules of thumb \citep{Bischl2023hpo}, which we aim to reflect in our illustration: we employ a random search algorithm, terminate the tuning after 60 evaluations (which corresponds to 30 times the dimension of the search space, as there are 2 HPs in $\hpa$), and use 10-fold CV for prediction error estimation. The tuning procedure is performed jointly for all HPs, which is the standard practice for automated tuning.\\
As with the previous model generation procedures, we report both the apparent error and the 10-fold CV error. Note that, since the 10-fold CV error for the selected HP configuration, $\hpaii$, has already been calculated during tuning, we use this value as the 10-fold CV error estimate of the final prediction model to avoid performing additional resampling iterations. Similar to the procedure {\pM}, data leakage is present in both evaluation procedures and may result in optimistically biased prediction error estimates. Specifically, the optimistic bias in the 10-fold CV error would arise from (resampling-induced) overtuning. 
Since the procedure {\pA} is fully automated, we additionally estimate the prediction error using nested CV. Here, we use 10 folds for the outer resampling loop and 2 folds for the inner resampling loop (the small number of inner folds saves computation time, and we only need to achieve correct HP selection rather than precise error estimation here; this is also recommended by \citealp{Bischl2023hpo}). As discussed in Section~\ref{sec:setting2_eval_all}, this evaluation procedure is not affected by data leakage.

\paragraph{\pC}
As a fourth model generation procedure, we tune both preprocessing and algorithm HPs (i.e.\ $\hpp$ and $\hpa$), but with two different tuning procedures. More specifically, the preprocessing HPs are tuned as in {\pM}, and for each candidate configuration of the preprocessing HPs, the algorithm HPs are tuned as in {\pA}.  Although this procedure might initially seem unintuitive and overly complex, it actually mirrors a realistic scenario for users who can tune algorithm HPs but may not be aware of or able to tune preprocessing HPs: Consider a user who has programmed three functions: (i) \mbox{\texttt{preprocess\_data}}, which takes the raw data set as input and returns the preprocessed data set; (ii) \mbox{\texttt{tune\_algorithm}}, which tunes the algorithm HPs as specified in {\pA} based on the preprocessed data set and returns the selected HPs $\hpaii$; and (iii) \mbox{\texttt{get\_apparent\_error}}, which takes the preprocessed data set and a learning algorithm with HPs $\hpaii$ as input and returns the apparent error of the resulting model. Suppose the user initially plans to run these three functions once but is dissatisfied with the apparent error reported by \mbox{\texttt{get\_apparent\_error}}. They would then modify \mbox{\texttt{preprocess\_data}} to try, for example, a different way of aggregating the IPOS score (i.e.\ using a different $\hpipos$) and rerun \mbox{\texttt{tune\_algorithm}} and \mbox{\texttt{get\_apparent\_error}}. After testing all values for $\hpipos$, they would proceed to adjust $\hpage$, $\hpakps$, and so forth, updating the algorithm HPs by running \mbox{\texttt{tune\_algorithm}} before calling \mbox{\texttt{get\_apparent\_error}} for each tried preprocessing configuration $\hpp$. Note that since 16 configurations for $\hpp$ are tried (see {\pM}), and for each configuration of $\hpp$, 60 candidate configurations for $\hpa$ are evaluated (see {\pA}), $60\times16 = 960$ HP configurations are assessed in total. The user would ultimately select the preprocessing HPs $\hppii$ that yield the best apparent error and the algorithm HPs $\hpaii$ returned by \mbox{\texttt{tune\_algorithm}} after setting $\hppii$ in \mbox{\texttt{preprocess\_data}}.\\
For this model generation procedure, we again consider the apparent error and the 10-fold CV error to evaluate the resulting prediction model. Note that the apparent error estimate corresponds to the best apparent error achieved during tuning and can therefore be directly adopted for evaluation. More specifically, it is the output of \mbox{\texttt{get\_apparent\_error}} after running \mbox{\texttt{preprocess\_data}} with $\hppii$ and then \mbox{\texttt{tune\_algorithm}}. The 10-fold CV error estimate can also directly be taken from the tuning procedure and corresponds to the 10-fold CV estimate which was calculated during the execution of \mbox{\texttt{tune\_algorithm}} after running \mbox{\texttt{preprocess\_data}} with $\hppii$. For the reasons discussed in the previous model generation procedures, both the apparent error and the 10-fold CV error estimates are subject to data leakage.

\paragraph{\pAii}
The final model generation procedure is similar to the procedure {\pA} described above, except that the set of jointly tuned HPs now also includes the five preprocessing HPs, $\hpp$, and the number of evaluations is increased to 210. As in {\pA}, this corresponds to 30 times the dimension of the search space, as there are now 7 tuned HPs. This procedure represents a conceptually simple way to incorporate preprocessing HPs into the tuning process and is recommended by \cite{Bischl2023hpo}. However, as noted in Section~\ref{sec:setting2_gen_am}, integrating preprocessing HPs into an automated tuning procedure requires advanced programming expertise, which may explain why this procedure is not standard practice yet.\\
We use the same three model evaluation procedures as in {\pA}, with the same considerations discussed in {\pA} also applying here.

\subsection{Results}\label{sec:results}
\begin{figure}[t]
    \centering
    \includegraphics[width=1\linewidth]{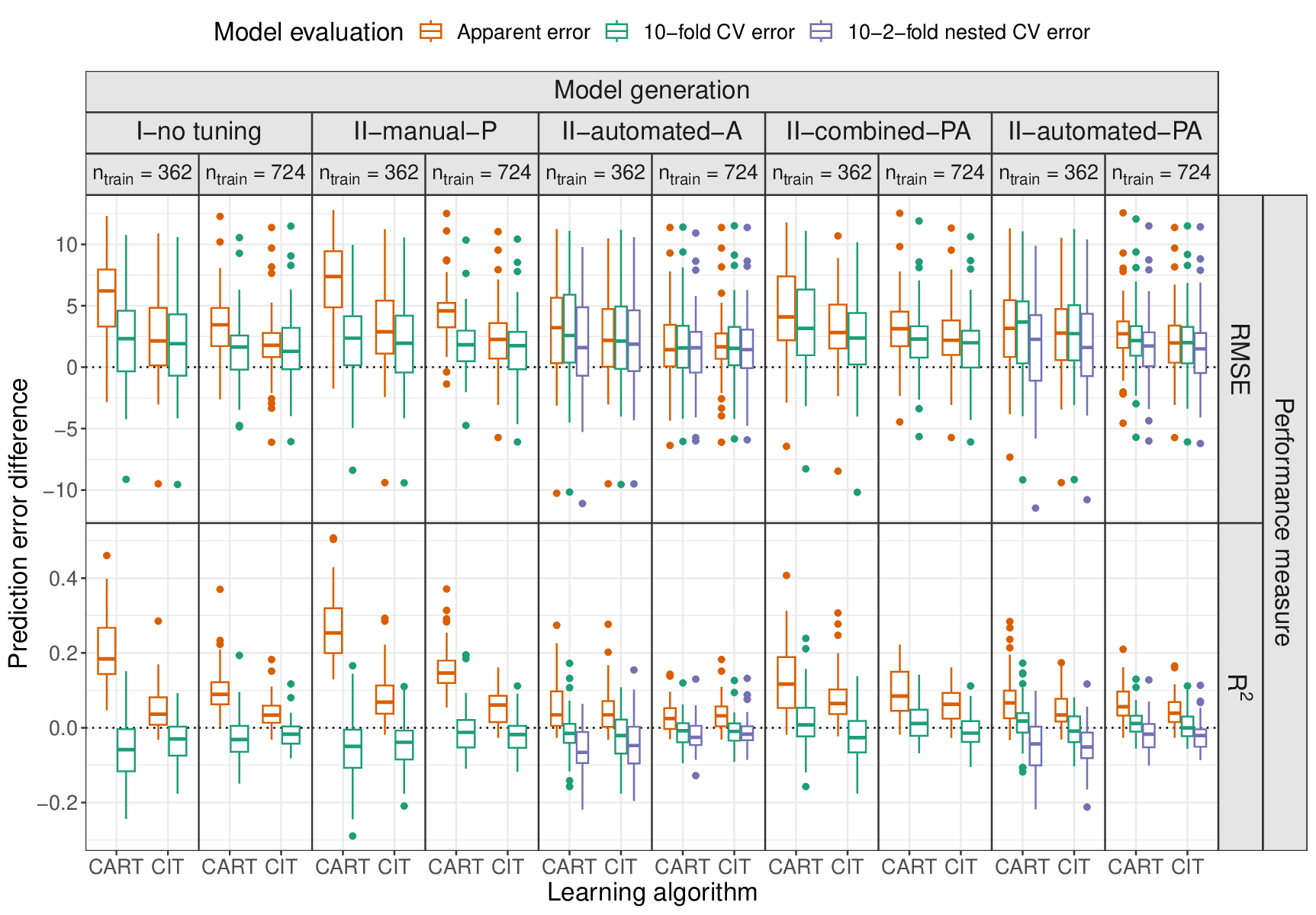}
    \caption{Resulting prediction error differences for 96 analysis settings, with each boxplot summarizing 50 repetitions of a specific setting. The prediction error differences are calculated as $\pednew - \ped$ for RMSE and $\ped - \pednew$ for $R^2$. For both performance measures, a positive median difference (averaged over the 50 repetitions) indicates that $\ped$ is optimistically biased, while a negative median difference suggests a pessimistic bias.}
    \label{fig:results_a_naive}
\end{figure}

Figure~\ref{fig:results_a_naive} 
illustrates the differences between $\ped$ and $\pednew$ for each of the 96 analysis settings (with 50 repetitions per setting). Additionally, the absolute values of $\ped$ and $\pednew$, as well as the selected HPs (for analysis settings where HPs are tuned), are presented in Figures~S2 to S6.\\
Before examining the prediction error differences in more detail, we first consider the absolute values of $\pednew$ (displayed in Figure~S2). Here, the general observation can be made that across all analysis settings, none of the generated models demonstrates sufficient predictive performance, which was expected and aligns with the findings of the COMPANION project. Of course, this result does not imply that HP tuning is generally not useful; rather, it demonstrates that tuning alone is not a guaranteed solution for obtaining a well-performing model for any prediction problem. Even in the analysis settings with the best median prediction errors (averaged across 50 repetitions), the median $\pednew$ reaches only 0.074 for $R^2$  ($\ntrain = 724$, CIT, \pM) and 42.1 for RMSE ($\ntrain = 724$, CIT, \pAii). For reference, the median $\pednew$ for RMSE using a naive model that predicts the mean of $\dtrain$ on $\dnew$ is 44.0 for the smaller sample size and 43.5 for the larger sample size, which is only slightly worse than the result from the decision tree models. While small effects of sample size and learning algorithm on $\pednew$ can be observed (with larger sample sizes and using the CIT instead of the CART algorithm resulting in smaller prediction errors), no clear pattern emerges for the model generation procedure.\\ 
We will now analyze the differences between $\ped$ and $\pednew$. To ensure consistent interpretation of their signs across both performance measures, the prediction error differences in Figure~\ref{fig:results_a_naive} are presented as $\pednew - \ped$ for RMSE and $\ped - \pednew$ for $R^2$. With this definition, a positive median difference indicates that the prediction error estimate $\ped$ is optimistically biased, while a negative median difference suggests a pessimistic bias.\\
As stated in Section~\ref{sec:setup_procedures}, depending on the model evaluation procedure, $\ped$ corresponds to one of three prediction error estimates: (i) the apparent error, (ii) the 10-fold CV error, or (iii) the 2-fold-within-10-fold CV error. We structure the reporting of the results according to these three evaluation procedures.
\paragraph{Apparent error}
Figure~\ref{fig:results_a_naive} shows that, across the considered model generation procedures, the median prediction error differences vary the most for the apparent error. Despite this variation, the median differences are consistently positive in all analysis settings. Although there are individual repetitions with negative differences, these results clearly indicate that the apparent error is optimistically biased. As discussed in Section~\ref{sec:setting1_eval_all}, this problem arises due to data leakage, or more specifically, the fact that this evaluation procedure uses observations for prediction error estimation that were already seen during model generation, which in turn allows potential overfitting and overtuning (if HPs are tuned) of the model to go undetected.\\
The optimistic bias of the apparent error is most pronounced in analysis settings where the preprocessing HPs $\hpp$ are tuned manually (\pM). This is not surprising, as this procedure specifically selects the HP values that optimize the apparent error. Here, the bias is largest when the smaller sample size and the CART algorithm are used for model generation, resulting in a median difference of 7.39 for RMSE and 0.253 for $R^2$. Note that while the absolute values of $\ped$ still do not indicate good predictive performance in these analysis settings (see Figure~S2), the median $R^2$ values resulting from the CART algorithm (0.234 and 0.176 for the two sample sizes) are comparable to the prediction errors reported for the Australian and UK decision tree models (0.17 and 0.27), which were generally deemed viable \citep{Eagar2004technical, Murtagh2023}. Regarding the selected HPs, particularly for $\hpipos$ (which specifies how the IPOS score is calculated) and $\hpca$ (which determines how \enquote{cannot assess} values in IPOS features are handled), alternative values are frequently chosen instead of the defaults (see Figures~S3a to S6a). This suggests that these alternative values may present a high potential for overfitting, thereby improving the apparent error.\\
In the analysis settings where both the preprocessing and the algorithm HPs are tuned using different procedures (\pC), the optimistic bias of the apparent error is similar for the CIT algorithm or slightly smaller for the CART algorithm compared to the {\pM} procedure. Again, the optimistic bias is largest in the analysis settings where a smaller sample size and the CART algorithm are considered, resulting in a median difference of 4.09 for RMSE and 0.117 for $R^2$. The slight decrease in optimistic bias can be attributed to the fact that, across all analysis settings using the {\pC} procedure, the algorithm HP $\hpminb$ is set to a higher value than its default of $\hpminb=7$, which results in a reduced risk of overfitting (see Figures~S3b to S6b). In the analysis settings where no HPs are tuned (\pdefault), the optimistic bias of the apparent error is also reduced slightly compared to the {\pM} procedure. For the smaller sample size combined with the CART algorithm, the observed median difference is 6.21 for RMSE and 0.184 for $R^2$. The reduction in optimistic bias compared to {\pM} is expected, as {\pdefault} does not involve HP tuning. \\
The lowest optimistic bias for the apparent error is observed in the analysis settings where either only $\hpa$ (\pA) or both $\hpp$ and $\hpa$ (\pAii) are tuned automatically, with the largest median difference being 3.22 for RMSE and 0.035 for $R^2$. This is not surprising, as in these procedures, all HPs are selected based on their associated CV error estimate rather than the apparent error. Notably, across all analysis settings, the HP values for $\hpp$ selected by the {\pAii} procedure differ from those chosen by the  {\pM} and {\pC} procedures (see  Figures~S3a to S6a).

\paragraph{CV error} If $\ped$ corresponds to the CV error, the resulting median prediction error differences indicate that this error is, as expected, generally less optimistic than the apparent error. The only exception occurs in a few analysis settings using RMSE as performance measure, where the apparent error differences are close to zero; here, the median differences of apparent error and CV error are approximately equal. \\
In the analysis settings without HP tuning, the $R^2$ differences exhibit a negative median difference, with the median difference closest to zero, -0.059, observed for the smaller sample size combined with the CART algorithm. This pessimistic bias is an expected result, as CV evaluates models trained on fewer observations than the final prediction model (see Section~\ref{sec:setting1_eval_all}). In contrast to $R^2$, the prediction error differences for RMSE in the analysis settings without tuning are mostly positive. Although the median differences are small (with the largest median difference being 2.32 in the analysis setting where both the smaller sample size and the CART algorithm are considered), the overall distribution of the prediction error differences in each setting suggests the presence of an optimistic bias. This finding is unexpected, as prediction errors estimated by CV in a setting where no HPs are tuned should not exhibit an optimistic bias but rather a  pessimistic bias (as observed for $R^2$). However, this can be attributed to the fact that both $\ped$ based on CV and $\pednew$ are affected by data leakage stemming from a violation of the assumption that all observations are independently drawn from the same distribution (see Section~\ref{sec:prelim_processes_eval} and Supplementary Section~A). This type of data leakage is distinct from the leakage caused by the overlap between the data used for model generation and evaluation, which is the primary focus of this paper. Specifically, the COMPANION data set exhibits a clustering structure that is not accounted for during the split into $\dtrain$ and $\dnew$ or during the creation of CV splits on $\dtrain$, resulting in a potential optimistic bias for both $\pednew$ (due to the initial split) and $\ped$ (due to the CV splits). As $\ped$ is also subject to a larger clustering-induced optimistic bias than $\pednew$, the bias does not cancel out when taking their difference and is therefore evident in Figure~\ref{fig:results_a_naive}. Notably, the different levels of clustering-induced optimistic bias in $\ped$ and $\pednew$ appear to have less impact on $R^2$, where, as described above, the prediction error differences are mostly negative. Further details on the impact of the clustering structure on the results, including an explanation of why it was not considered when performing the splits, are provided in Supplementary Section~B.5.\\
The additional source of optimistic bias introduced by the clustering structure of the data is also relevant when interpreting the prediction error differences in the analysis settings with HP tuning. While our primary focus here is on overlap-induced data leakage that arises since the observations used for the CV-based error estimation have already been seen during HP tuning (thus hindering the detection of potential overtuning), we have to consider that any observed optimistic bias may as well stem from clustering-induced data leakage. Consequently, we compare the prediction error differences in analysis settings with HP tuning to those in settings without tuning (where only clustering-induced data leakage is present) rather than directly comparing them to zero. Based on this assessment, the impact of overlap-induced data leakage on $\ped$ appears to be limited. This is particularly true for RMSE, where the CV error differences are generally comparable to those resulting from the {\pdefault} procedure.  For $R^2$, the median differences tend to be closer to zero compared to the {\pdefault} procedure. In some analysis settings involving the smaller sample size and the CART algorithm, there is even a positive median difference (with the largest median difference of 0.018 observed in the setting where {\pAii} is used in combination with the smaller sample size and the CART algorithm). Consequently, there appears to be a small overtuning effect that is not detected by the CV error due to overlap-induced data leakage. However, the median differences are too close to zero, and the variation within each analysis setting is too large to definitively determine which bias ultimately predominates, i.e.\ whether the CV error is overall optimistic or pessimistic in these settings.

\paragraph{Nested CV error} In the analysis settings using the {\pA} or {\pAii} procedures for model generation, the prediction error differences of the nested CV error can also be analyzed. As expected, we observe the tendency for the nested CV error to be more pessimistic than the simple CV error (indicated by the smaller differences compared to the CV error; however, in some settings, the median differences for simple and nested CV errors are approximately equal). Although the nested CV error is not affected by the optimistic bias that may result from undetected overtuning effects (see Section~\ref{sec:setting2_eval_all}), the median differences for RMSE are positive, indicating the presence of an optimistic bias. As discussed above for the simple CV error, this is due to the clustering-induced optimistic bias, which appears to outweigh the pessimistic bias typically associated with nested resampling. In the analysis settings using $R^2$ as performance measure, the distribution of the prediction error differences indicates that the nested CV error is pessimistically biased.\\

To summarize, the choice of model generation and evaluation procedure generally affects the difference between the prediction error estimates derived from available data and new data. As expected, when the evaluation procedure is based on the apparent error, the resulting estimate exhibits an optimistic bias, which varies depending on the model generation procedure.
As likewise expected, the simple CV error is less optimistic than the apparent error, while the nested CV error is even less optimistic. The corresponding prediction error differences are less variable across model generation procedures compared to the apparent error. For simple CV, this indicates that, in the considered experimental setup, the tuning procedures do not introduce relevant overtuning effects on error estimation. Instead, the main source of bias for simple CV is either the clustering-induced optimistic bias (or, more precisely, the different bias level relative to $\pednew$) or the pessimistic bias arising from the use of fewer observations during evaluation. This also holds true for the nested CV error.

\section{Discussion and conclusion}\label{sec:discussion}
This paper reviewed and empirically demonstrated the implications and potential pitfalls of HP tuning in the generation and evaluation of prediction models from the perspective of applied ML users, with a specific focus on the distinction between preprocessing and algorithm HPs.\\
While HP tuning is generally a powerful tool for improving model performance, it also introduces potential sources of error. In the model generation process, failing to select an adequate tuning procedure can result in a prediction model that performs no better, or even worse, than a model using default HP settings. During model evaluation, failing to properly account for HP tuning can lead to optimistically biased prediction error estimates. The risk of such errors is especially high for preprocessing HPs, as they are often tuned subconsciously. \\
To provide different examples of model generation and evaluation procedures in the context of HP tuning and to examine their impact on the difference between prediction error estimates from available and new data, we conducted an illustrative study using a real-world prediction problem from palliative care medicine. Although both the apparent error and CV error can, in theory, be optimistically biased when HPs are tuned, this was consistently true only for the apparent error (with the highest optimistic bias occurring in analysis settings that imitated manual tuning of preprocessing HPs without considering algorithm HPs). In contrast, the prediction error differences for the CV error appeared not to be considerably compromised by data leakage, as these differences were comparable to the analysis settings without HP tuning.\\ 
In addition to explicitly considering preprocessing HPs and manual tuning procedures, our illustrative study stands out from other investigations on HP tuning by not only using real data but also building most of the setup (including the learning pipeline, HPs, and performance measures) on a real-world project. While this ensures that the observed results are realistic and not derived from overly simplified or extreme setups, they are not generalizable beyond this specific context because the considered real-world project and the derived setup are not representative of other ML applications. By using real data, our illustration was also limited in that we could only compare the prediction error estimates from the available data set to those from a new data set (which, due to the clustering structure, was also over-optimistic) instead of comparing it to the true prediction errors. Nevertheless, it was still possible to compare differences across analysis settings and derive tendencies. Finally, the illustration could have been extended by treating the learning algorithm as a tunable HP. However, with the given setup, doing so would offer limited insights, as it is reasonably predictable that the resampling-based tuning procedures would select the CIT algorithm, while the tuning procedures based on the apparent error would favor the CART algorithm.\\
Based on these conceptual and empirical insights, it is clear that to ensure HP tuning becomes a benefit rather than a pitfall, applied ML users must take care throughout the entire model development process. First, they should thoroughly consider which HPs (including preprocessing HPs) are to be tuned and which are not. An adequate tuning procedure that fits the specific prediction problem should then be specified. Unfortunately, this is typically non-trivial, as it depends on various factors such as sample size and the specific HPs to be tuned. More research is needed to better guide users in this respect (see \citealp{Bischl2023hpo}, for an overview of current recommendations). In general, it is recommended to use automated tuning procedures instead of manual ones (see again \citealp{Bischl2023hpo}, for automated tuning implementations in \texttt{R} and \texttt{Python}). If automated tuning is not feasible, users should at least ensure that the manual tuning procedure is error-free, reproducible, and resampling-based. For model evaluation, only two evaluation procedures are guaranteed to be unaffected by data leakage caused by HP tuning: (i) nested resampling (if the entire data set is used for model generation) or (ii) a permanent (outer) holdout (if only a subset of the available data is used for model generation). However, similar to the tuning procedure, there is a lack of guidance on how to choose between these approaches and how to specify them (e.g., which resampling methods to use for nested resampling).
Although simple resampling may turn out to be a viable option in some applications (including our example), this can generally not be known in advance. Therefore, we discourage its use in settings involving HP tuning, as well as any other evaluation procedures that could result in data leakage. \\ 
Regardless of how model generation and evaluation are performed, it is essential that they and all other relevant details (e.g., the complete learning pipeline and its HPs) are transparently reported in both code and text form. For this purpose, users may rely on checklists such as REFORMS (\citealp{Kapoor2024reforms}; intended for all applied research fields using ML) or TRIPOD+AI (\citealp{Collins2024tripodai}; intended for clinical prediction models). While transparency does not imply correctness, it allows readers to identify potential issues, such as data leakage, and to critically interpret the claimed model performance.  Moreover, it emphasizes the existence and importance of preprocessing and its HPs, while the current lack of transparency can create the impression that the data were not preprocessed at all or that no alternative preprocessing options were explored. To further enhance transparency and encourage applied ML users to be more intentional about their choices, it is also possible to preregister the entire model development process, for example, by using the template proposed by \cite{Hofman2023}.\\
In conclusion, by addressing the implications and pitfalls of HP tuning from an applied perspective and emphasizing often-overlooked aspects, we hope that this review can further enhance the quality of ML-based predictive modeling.

\section*{Funding Information}
This work was supported by the German Research Foundation (BO3139/9-1, BO3139/7) to ALB. The authors of this work take full responsibility for its content.

\section*{Acknowledgments}
The authors thank Patrick Callahan for language corrections and Julian Lange for useful literature input.

\section*{Conflicting interests}
The authors have declared no conflicts of interest for this article.

\FloatBarrier

\printbibliography

\newpage
\section*{Supplementary Material}
\appendix
\beginsupplement

\section{Other leakage types} \label{sec:supp_leakage}
As stated in Section~2.4.2, \cite{Kapoor2023leakage} identify 
three general types of data leakage, which may arise from: (i) overlap between the data used for model generation and evaluation, (ii) violation of the assumption that all observations are independently drawn from the same distribution, or (iii) use of illegitimate features. While our paper primarily addresses overlap-induced data leakage, we will now provide additional details on the other two types.

\subsection{Violation of the i.i.d.\ assumption}\label{sec:supp_leakage_iid}
In the following, we first consider the case of Setting I with $\dtrain = \ds$ and  discuss the implications for $\dtrain \subset \ds$ and Setting II afterwards. \\
Even with a strict separation between the data used for model generation and evaluation, achieved through the use of resampling methods, data leakage can still occur if the assumption that all observations in $\dtrain$ are independently drawn from the same distribution is violated. This assumption, also known as the i.i.d.\ assumption, was stated in Section~2.1. Non-i.i.d.\ settings may, for example, arise when $\dtrain$ is a clustered data set, i.e.\ when the observations originate from different clusters (e.g., study centers). Observations within clusters are typically more similar than observations between clusters, where similarity can refer to both the feature vector $\bs{x}^{(i)}$ or the outcome $y^{(i)}$ \citep{Hornung2023}. If the prediction model is intended to be applied to observations from other clusters than those present in $\dtrain$ in the future, resampling methods that are based on random sampling (i.e.\ ignoring the cluster structure) will be optimistically biased since in each resampling iteration, the observations in $\dtest$  are more similar to $\dtraine$ than observations originating from new clusters 
\citep{Hornung2023,Kapoor2023leakage,Rosenblatt2024}. Although the level of optimistic bias depends on the specific clustering structure (e.g., cluster size and correlation within clusters), it is generally recommended to perform grouped resampling at cluster level, where all observations in a cluster are either assigned to $\dtraine$ or $\dtest$ in each resampling iteration \citep{Hornung2023,Bischl2023hpo}. In the context of healthcare research, this type of resampling is referred to as internal-external validation \citep{Collins2024, Debray2023}. For other examples of non-i.i.d.\ settings and corresponding resampling methods, see \citet{Hornung2023} and the references therein.
\\
Our elaborations also apply to the case of Setting I with $\dtrain \subset \ds$, with a permanent holdout used instead of a (temporary) resampling method; here, one simply replaces $\dtrain$ with $\ds$ and $\dtraine$ with $\dtrain$.\\
In Setting II, where resampling is typically used for both model generation (tuning) and evaluation, data leakage due to the violation of the i.i.d.\ assumption biases the prediction error estimate of the final model only when the non-i.i.d.\ data structure is ignored during model evaluation. This occurs specifically in the outer resampling loop of nested resampling (for $\dtrain = \ds$) or in the permanent outer holdout (for $\dtrain \subset \ds$). However, it is recommended to also take into account the non-i.i.d.\ data structure during tuning, both for the final prediction model and, if nested resampling is used, within the inner resampling loop, to ensure consistency  \citep{Hornung2023}.

\subsection{Use of illegitimate features}
If $\dtrain$ and $\dtest$ include features that are generally not available for new observations to which the model will be applied in practice, these features can be considered illegitimate, and if included in the final prediction model, constitute another type of data leakage. An example raised by \cite{Kapoor2023leakage} is the use of anti-hypertensive drugs as a feature for predicting hypertension. Note that this type of data leakage is conceptually different from the other two types, as it stems from a design issue that is independent of the model evaluation procedure.

\section{Additional information on the empirical illustration}
\subsection{Descriptive statistics}
Table~\ref{tab:companion_descr} provides descriptive statistics of the COMPANION data set used in the empirical illustration.

\begin{table}[!htbp]   
    \caption{Distribution of the outcome variable and features in the COMPANION data set %(specialist palliative home care setting) 
    after applying the initial preprocessing steps (described in Supplementary Section~\ref{sec:supp_preproc1}). 
    In addition, two preprocessing steps from the learning pipeline $\lp$ (see Section~5.2.2 and Supplementary Section~\ref{sec:supp_preproc2}) have been performed: the correction of costs and the aggregation of the IPOS score (default version). }
    \label{tab:companion_descr}
    \centering
   % \small
\begin{tabular}[t]{lr}
\toprule
%& Overall\\
& $n= $1{,}449\\
\midrule
\addlinespace[0.3em]
\multicolumn{2}{l}{\textbf{Average cost per day per palliative care phase (\euro)}}\\
\hspace{1em}Mean (SD) & 49.0 (43.1)\\
\hspace{1em}Median [Min, Max] & 35.9 [0.315, 357]\\
\addlinespace[0.3em]
\multicolumn{2}{l}{\textbf{Palliative care phase}}\\
\hspace{1em}stable & 453 (31.3\%)\\
\hspace{1em}unstable & 281 (19.4\%)\\
\hspace{1em}deteriorating & 486 (33.5\%)\\
\hspace{1em}terminal & 229 (15.8\%)\\
\addlinespace[0.3em]
\multicolumn{2}{l}{\textbf{Age (years)}}\\
\hspace{1em}Mean (SD) & 74.7 (12.2)\\
\hspace{1em}Median [Min, Max] & 76.0 [23, 102]\\
\addlinespace[0.3em]
\multicolumn{2}{l}{\textbf{Confusion}}\\
\hspace{1em}absent & 950 (65.6\%)\\
\hspace{1em}mild & 248 (17.1\%)\\
\hspace{1em}moderate & 144 (9.9\%)\\
\hspace{1em}severe & 107 (7.4\%)\\
\addlinespace[0.3em]
\multicolumn{2}{l}{\textbf{Agitation}}\\
\hspace{1em}absent & 837 (57.8\%)\\
\hspace{1em}mild & 306 (21.1\%)\\
\hspace{1em}moderate & 217 (15.0\%)\\
\hspace{1em}severe & 89 (6.1\%)\\
\addlinespace[0.3em]
\multicolumn{2}{l}{\textbf{AKPS}}\\
\hspace{1em}(10) comatose or barely rousable & 79 (5.5\%)\\
\hspace{1em}(20) totally bedfast and requiring extensive nursing care 
\\\hspace{1em}by professionals and/or family & 381 (26.3\%)\\
\hspace{1em}(30) almost completely bedfast & 242 (16.7\%)\\
\hspace{1em}(40) in bed more than 50\% of the time & 270 (18.6\%)\\
\hspace{1em}(50) considerable assistance and frequent medical care required & 265 (18.3\%)\\
\hspace{1em}(60) able to care for most needs; but requires occasional assistance & 151 (10.4\%)\\
\hspace{1em}(70) cares for self; unable to carry on normal activity or 
\\\hspace{1em}to do active work & 38 (2.6\%)\\
\hspace{1em}(80) normal activity with effort; some signs or symptoms of disease & 14 (1.0\%)\\
\hspace{1em}(90) able to carry on normal activity; minor sign of symptoms 
\\\hspace{1em}of disease & 9 (0.6\%)\\
\addlinespace[0.3em]
\multicolumn{2}{l}{\textbf{IPOS total score}}\\
\hspace{1em}Mean (SD) & 24.8 (7.98)\\
\hspace{1em}Median [Min, Max] & 25.0 [2.00, 55.0]\\
\bottomrule
\end{tabular}

\end{table}

\subsection{Preprocessing steps} 

\subsubsection{Initial preprocessing steps}\label{sec:supp_preproc1}
In the following, we describe the parameterless and pre-specified preprocessing steps that are applied to the full COMPANION data set in its rawest version available. Note that the raw data set is on patient contact level, which was the unit for data collection \citep{Hodiamont2022}. The initial preprocessing steps are:
\begin{enumerate}[label=(\roman*)]
    \item data cleaning steps (e.g., correct variable types and labels),
    \item the removal of contacts with palliative care phase \enquote{bereavement}, AKPS $= 0$ (\enquote{dead}), or costs $= 0$, 
    \item  the aggregation of the contact level data into palliative care phase level data  (the outcome is constructed by summing the costs of all patient contacts and dividing by the number of days in the corresponding phase; for features that may vary during a phase, the highest value of the first day is used),
    \item the removal of palliative care phases (one phase with an extreme and implausible cost value is removed; phases with \enquote{missing} values in either one or both cognitive features or in one of the individual IPOS features are removed; phases with \enquote{missing} or \enquote{cannot assess} in the AKPS feature are removed), and
    \item the replacement of \enquote{cannot assess} values with \enquote{absent} in the two cognitive features.
\end{enumerate}
These preprocessing steps yield a data set with 1{,}449 observations.

\subsubsection{Preprocessing steps in the learning pipeline}\label{sec:supp_preproc2}
In this section, we detail the six preprocessing steps of the learning pipeline $\lp$ that is applied in each training process, including their associated HPs. An overview of these preprocessing steps is given in Figure~4. 

\paragraph{Correction of costs} As stated in Section~5.1, the outcome variable $y^{(i)}$ is defined as the average cost per day in palliative care phase $i$, which is intended to reflect the resource needs in that phase. This variable is calculated based on the staff time used to care for a patient and their relatives on each day of the corresponding palliative care phase. However, analyses have shown that if a palliative care phase is the first phase in an episode of care (see Supplementary Section~\ref{sec:supp_cluster} for more information on episodes of care), the staff time and thus the costs of the first day are increased regardless of the complexity of the palliative care situation (e.g., due to time-consuming admission interviews). For this reason, the first-day costs of the first phase of an episode are adjusted using a factor based on comparisons with the costs of the first days in later phases of an episode. This factor is initially calculated for each palliative care team and then averaged to obtain a single overall correction factor, denoted as $\theta_{correct}$. This preprocessing step accordingly includes a parameter that must be estimated from the data set, though it does not involve any HPs in our illustration. Moreover, it is a step that modifies the outcome (albeit slightly), not for compatibility with the learning algorithm, but to change the interpretation of the prediction model, which now intends to predict a corrected version of the outcome. Accordingly, this step is also applied during prediction.

\paragraph{Removal of cost outliers} The distribution of the outcome variable in the COMPANION data set is right skewed, i.e.\  some palliative care phases have exceptionally high costs (see Table~\ref{tab:companion_descr}). Since it is not possible to definitively attribute these values to data entry errors, they are not permanently removed from the data set. 
However, since the prediction values calculated by the corresponding decision tree algorithm in each terminal node can be sensitive to outliers, removing cost outliers during the training process could improve model performance. Importantly, this preprocessing step is only applied during training and not during prediction, i.e.\ when the final prediction model is used to make predictions on a data set, no cost outliers are removed. Removing them during prediction could artificially improve the model's performance, as cost outliers are typically difficult to predict correctly \citep[see also][]{Kapoor2023leakage}. \\
The definition of outliers is generally not straightforward, as many possible options exist \citep{Kuhn2013,Steyerberg2019}. We denote the corresponding HP as $\hpoutlier$. In our illustration, we define all cost values higher than the $\hpoutlier$th cost percentile as outliers, with $\hpoutlier \in \{100,99,95,90\}$. If $\hpoutlier=100$ (the default value), no outliers are removed. Note that this preprocessing step includes the parameter $\theta_{outlier}$, which corresponds to the percentile calculated according to $\hpoutlier$.

\begin{figure}[!ht]
    \centering
    \includegraphics[width=1\linewidth]{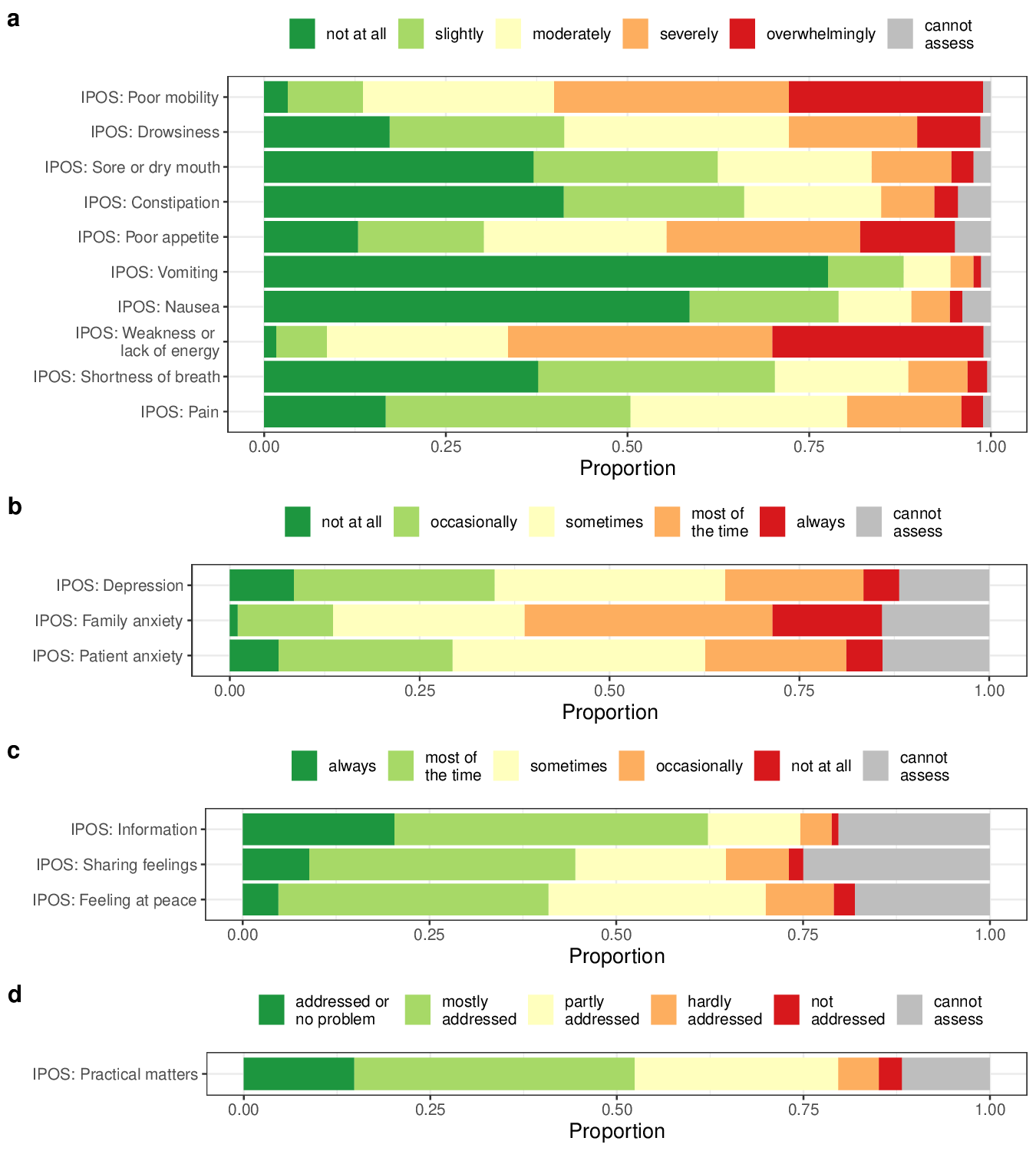}
    \caption{Distribution of the 17 individual IPOS features in the COMPANION data set after applying the initial preprocessing steps (described in Supplementary Section~\ref{sec:supp_preproc1}). a: Physical symptoms. b: Emotional symptoms. c: Communication issues. d: Practical issues.}
    \label{fig:ipos}
\end{figure}

\paragraph{Handling of \enquote{cannot assess} values in IPOS features} As outlined in Section~5.1, the set of features  to generate the prediction model includes the Integrated Palliative care Outcome Scale (IPOS; \citealp{Murtagh2019ipos}), which is a score based on 17 individual features covering physical symptoms, psycho-social burden, family needs, and practical problems. Each of the 17 features is ordinal and can take values from 0 to 4, where 0 and 4 correspond to the least and highest symptom or concern severity, respectively. For example, for the features IPOS-\enquote{Pain} and IPOS-\enquote{Shortness of Breath}, a value of 0 corresponds to ``not at all'' and a value of 4 corresponds to ``overwhelmingly'' (see Figure~\ref{fig:ipos} for an overview of all 17 features). 
In its default version (see the next preprocessing step), the IPOS score is constructed by summing all 17 features, resulting in a score that ranges from 0 to 68. However, each IPOS feature also includes missing values, which are either due to missing data entries (coded as ``missing'') or because the response option \enquote{cannot assess} was selected during the IPOS assessment.
For example, assessing whether a patient is burdened by pain (IPOS-\enquote{Pain}) can be challenging for clinical staff if the patient is comatose. \\
While observations affected by the first type of missing values (``missing'') do not occur often and are removed as part of the initial preprocessing steps described in Supplementary Section~\ref{sec:supp_preproc1}, handling the \enquote{cannot assess} values is more challenging. If all observations with at least one \enquote{cannot assess} response were removed, almost half of the COMPANION data set would be discarded (see Table~\ref{tab:companion_ca}; this would also apply approximately to any subset $\dtrain$ or $\dnew$ of the  COMPANION data set). To avoid the loss of valuable information, an alternative approach is to treat \enquote{cannot assess} values as 0 (i.e.\  least symptom or concern severity), based on the assumption that an unobserved burden does not initiate a care mandate and therefore does not result in costs. However, it is not clear whether this assumption is valid for observations where many or even all IPOS features are recorded as \enquote{cannot assess} (e.g., if 15 out of 17 IPOS features are recorded as \enquote{cannot assess}, these features might not have been assessed at all). It could thus be a reasonable approach to set \enquote{cannot assess} values to 0 but exclude observations with many \enquote{cannot assess} values, as they potentially result in incorrect IPOS scores. %% and can be viewed as a form of measurement error.
Specifying the exact threshold for the maximum number of \enquote{cannot assess} values is, however, not straightforward. It can be denoted as HP $\hpca$, and ranges from 0 to 17  (observations with more than $\hpca$ \enquote{cannot assess} values are removed; if $\hpca$ = 17, no observations are removed). In our illustration, we consider the values $\{16, 14, 12, 10\}$ for $\hpca$, with $\hpca =16$ being the default. \\
This preprocessing step does not have any parameters. Since it removes observations, it modifies the distribution of the outcome variable. We argue that if observations with more than $\hpca$ \enquote{cannot assess} values are found to yield unreliable IPOS scores, the resulting prediction model should not be used for future observations where this criterion applies, implying that the corresponding preprocessing step alters the scope of the model (such that it cannot be used for observations with more than $\hpca$ IPOS features recorded as \enquote{cannot assess}). Accordingly, this step is also applied during the prediction process. As shown in Table~\ref{tab:companion_ca}, the change in the outcome distribution is, however, minimal because the values considered for $\hpca$ remove only a small number of observations (9 observations for $\hpca = 10$ and 0 observations for $\hpca = 16$) from the full COMPANION data set with 1{,}449 observations. As discussed in Section~2.3.4, it is recommended to specify HPs of preprocessing steps that affect the outcome distribution based on user expertise rather than tuning. However, given that this step only removes a few observations and because specifying $\hpca$ based on user expertise is challenging, we argue that $\hpca$ can be tuned. 

\begin{table}[ht]
    \centering
        \caption{Outcome distribution (average cost per day per palliative care phase) in the full COMPANION data set  (after applying the initial preprocessing steps described in Supplementary Section~\ref{sec:supp_preproc1}) if observations with more than $\lambda_{ca} \in \{0,10,12,14,16\}$ ``cannot assess'' values in the 17 individual IPOS features are removed. The minimum and maximum number of ``cannot assess'' values are 0 and 17, respectively.}
    \label{tab:companion_ca}

\begin{tabular}{ll}
\toprule
\addlinespace[0.3em]
\multicolumn{2}{l}{\textbf{$\lambda_{ca} = 0$  }}\\
\hspace{1em}Mean (SD) & 48.62 (45.12)\\
\hspace{1em}Median [Min, Max] & 34.96 [1.11, 356.70]\\
\hspace{1em}Missing & 662 (45.7\%)\\
\addlinespace[0.3em]
\multicolumn{2}{l}{\textbf{$\lambda_{ca} = 10$  }}\\
\hspace{1em}Mean (SD) & 49.03 (43.14)\\
\hspace{1em}Median [Min, Max] & 35.91 [0.32, \vphantom{1} 356.70]\\
\hspace{1em}Missing & 9 (0.6\%)\\
\addlinespace[0.3em]
\multicolumn{2}{l}{\textbf{$\lambda_{ca} = 12$  }}\\
\hspace{1em}Mean (SD) & 48.98 (43.09)\\
\hspace{1em}Median [Min, Max] & 35.91 [0.32, 356.70]\\
\hspace{1em}Missing & 3 (0.2\%)\\
\addlinespace[0.3em]
\multicolumn{2}{l}{\textbf{$\lambda_{ca} = 14$  }}\\
\hspace{1em}Mean (SD) & 48.99 (43.07)\\
\hspace{1em}Median [Min, Max] & 35.92 [0.32, \vphantom{1} 356.70]\\
\hspace{1em}Missing & 2 (0.1\%)\\
\addlinespace[0.3em]
\multicolumn{2}{l}{\textbf{$\lambda_{ca} = 16$  }}\\
\hspace{1em}Mean (SD) & 48.98 (43.05)\\
\hspace{1em}Median [Min, Max] & 35.92 [0.32, 356.70]\\
\hspace{1em}Missing & 0 (0.0\%)\\
\bottomrule
\end{tabular}

\end{table}

\paragraph{Calculation of IPOS score} After removing observations based on their individual IPOS feature values, the next preprocessing step is to construct the IPOS score from these features. Aggregating the individual IPOS features into an IPOS score can be done in several ways, and we denote the corresponding HP as $\hpipos$.
A straightforward and commonly used option is to simply sum the values of all 17 IPOS features, which we denote as IPOS-total (the default of $\hpipos$). \\
Instead of aggregating all 17 IPOS features into one score, it is also possible to generate multiple IPOS scores based on the subscales in which the features can be divided \citep{Murtagh2019ipos}. These subscales are: (i) physical symptoms (10 features), (ii) emotional symptoms (4 features), and (iii) communication/practical issues (3 features) (see Figure \ref{fig:ipos}). In our illustration, we consider the generation of two subscale scores: one score that sums the  features corresponding to the physical symptoms (IPOS-physical; $[0, 40]$) and one score that sums the remaining features (IPOS-others; $[0,28]$). Note that in this case, the number of features provided to the learning algorithm increases from $p=6$ to $p=7$.\\
A third option to construct the IPOS score is to sum all 17 IPOS features as in the IPOS-total score, but  recode them (before summing) as 1 if their value is $\in \{3,4\}$ (i.e.\  takes one of the two most extreme values), and 0 otherwise.
This score will be referred to as the IPOS-extreme score and ranges from 0 to 17. It was developed by the COMPANION team and was motivated by the possibly too strict assumption made by the previous preprocessing step, namely that \enquote{cannot assess} values are equivalent to a value of 0. This assumption is relaxed by the IPOS-extreme score, which only requires assuming that the true value of an IPOS feature recorded as \enquote{cannot assess} is  $\in \{0, 1, 2\}$ and not necessarily equal to 0.\\
The fourth considered IPOS score option is similar to the IPOS-extreme score, except that the features IPOS-\enquote{Pain} and IPOS-\enquote{Shortness of Breath} are excluded from the score (which now ranges from 0 to 15) and are instead provided separately on their original ordinal scale to the learning algorithm. The motivation for this version is that pain and shortness of breath may be strong predictors of the costs associated with a palliative care phase. Therefore, model performance might be improved by including IPOS-\enquote{Pain} and IPOS-\enquote{Shortness of Breath}  as individual features rather than aggregating them into the IPOS-extreme score.
If this IPOS option is used, the number of features provided to the learning algorithm increases from $p=6$ to $p=8$.\\
This preprocessing step does not have any parameters. Moreover, it does not alter the outcome distribution, which is why it is applied during both training and prediction.

\paragraph{Modification of feature \enquote{age}} In the COMPANION data set, age is measured on an integer scale and ranges from 23 to 102 years  (see Table~\ref{tab:companion_descr}). In its default configuration, this feature is provided to the learning algorithm on its original integer scale, without any preprocessing. Alternatively, it could be transformed into a categorical feature with six categories, using the years 50, 60, 70, 80, and 90 as cutpoints. This option could improve the model's prediction error, as, for example, the CART algorithm suffers from a selection bias towards features with many possible splits \citep{partykit_package2}. We refer to the HP that specifies the used option as $\hpage$, with no modification of age as default. This preprocessing step has the same characteristics as the aggregation of individual IPOS features into a score (i.e.\ no parameters, applied during training and prediction). 

\paragraph{Modification of feature \enquote{AKPS}}  The Australia-modified Karnofsky Performance Status (AKPS; \citealp{Abernethy2005akps}), which measures patients' functional status on an ordinal scale, takes values of $\{10, 20, ..., 90\}$ in the COMPANION data set, with AKPS $= 10$ corresponding to ``comatose or barely rousable'' and AKPS $= 90$ to ``able to carry on normal activity; minor sign of symptoms of disease'' (see Table~\ref{tab:companion_descr}). In its default configuration, AKPS is considered ordinal, with the three highest categories, 70, 80, and 90, merged due to their low frequency. However, it might also be reasonable to transform AKPS into an unordered categorical variable, as costs may not monotonically decrease or increase with AKPS, but could be highest when the patient has, for example, an AKPS of 50, which corresponds to \enquote{considerable assistance and frequent medical care required}.  In this case, we collapse the AKPS categories even further to avoid overfitting, resulting in AKPS $\in \{10\textnormal{-}20,30\textnormal{-}50,60\textnormal{-}90\}$. We refer to the corresponding HP as $\hpakps$, with the ordered AKPS variable as default. This preprocessing step has the same characteristics as the two previous preprocessing steps (i.e.\ no parameters, applied during training and prediction).\\

Note that for the preprocessing steps estimating parameters from the available observations (i.e.\  correction of costs, with  $\theta_{correct}$,  and removal of cost outliers, with $\theta_{outlier}$), their position in the preprocessing pipeline in relation to the steps where observations are removed (i.e.\  removal of outliers and handling of \enquote{cannot assess} values) is of relevance since a different set of observations might yield a different parameter estimate. Accordingly,  performing the preprocessing steps in a different order could lead to (slightly) different results. \\
Moreover, during the execution of the illustration as described in Section~5.2.1, in some resampling iterations performed during model generation and evaluation (particularly for nested CV), it occasionally happens that certain ordinal or categorical features in the data subset for which predictions are being made contain new values that were not encountered during training. This issue occurs exclusively with the highest and/or lowest values of these features, which are less frequent in the original COMPANION data set and thus more likely to be absent in the training set. Specifically, this affects the highest value of (cognitive) agitation, the highest and lowest values of AKPS (if AKPS is not collapsed into three unordered categories), the lowest value of age (if age is transformed into a categorical feature), and the highest values of \enquote{Pain} and IPOS-\enquote{Shortness of Breath} (if the fourth option for aggregating the IPOS score is selected). In these cases, we collapse the highest and second highest and/or lowest and second lowest values when making predictions.

\subsection{Performance measures}\label{sec:supp_performance}
In the illustration, two performance measures are considered: RMSE and $R^2$. The RMSE is obtained by taking the square root of the MSE (see Section~3.1) and is expressed in the same units as the outcome variable (i.e.\ costs in \euro). It ranges from 0 to $\infty$, where $\text{RMSE} = 0$ indicates perfect prediction. The $R^2$ performance measure is calculated by dividing the squared error of the prediction model by the squared error of a naive model that predicts the mean and then subtracting this ratio from 1. It is a relative measure that can be interpreted as the proportion of variance in the outcome variable explained by the prediction model. The range of $R^2$ is $(-\infty, 1]$, with $R^2= 1$ indicating perfect prediction and a $R^2$ value of 0 or less indicating that a model performs no better or worse than the naive model, respectively.  In this context, a lower prediction error corresponds to a higher $R^2$ value. See, e.g., \cite{Kuhn2013} for more details on both performance measures.
\newpage

\subsection{Absolute prediction error estimates and selected HPs}
\begin{figure}[h!]
    \centering
    \includegraphics[width=1\linewidth]{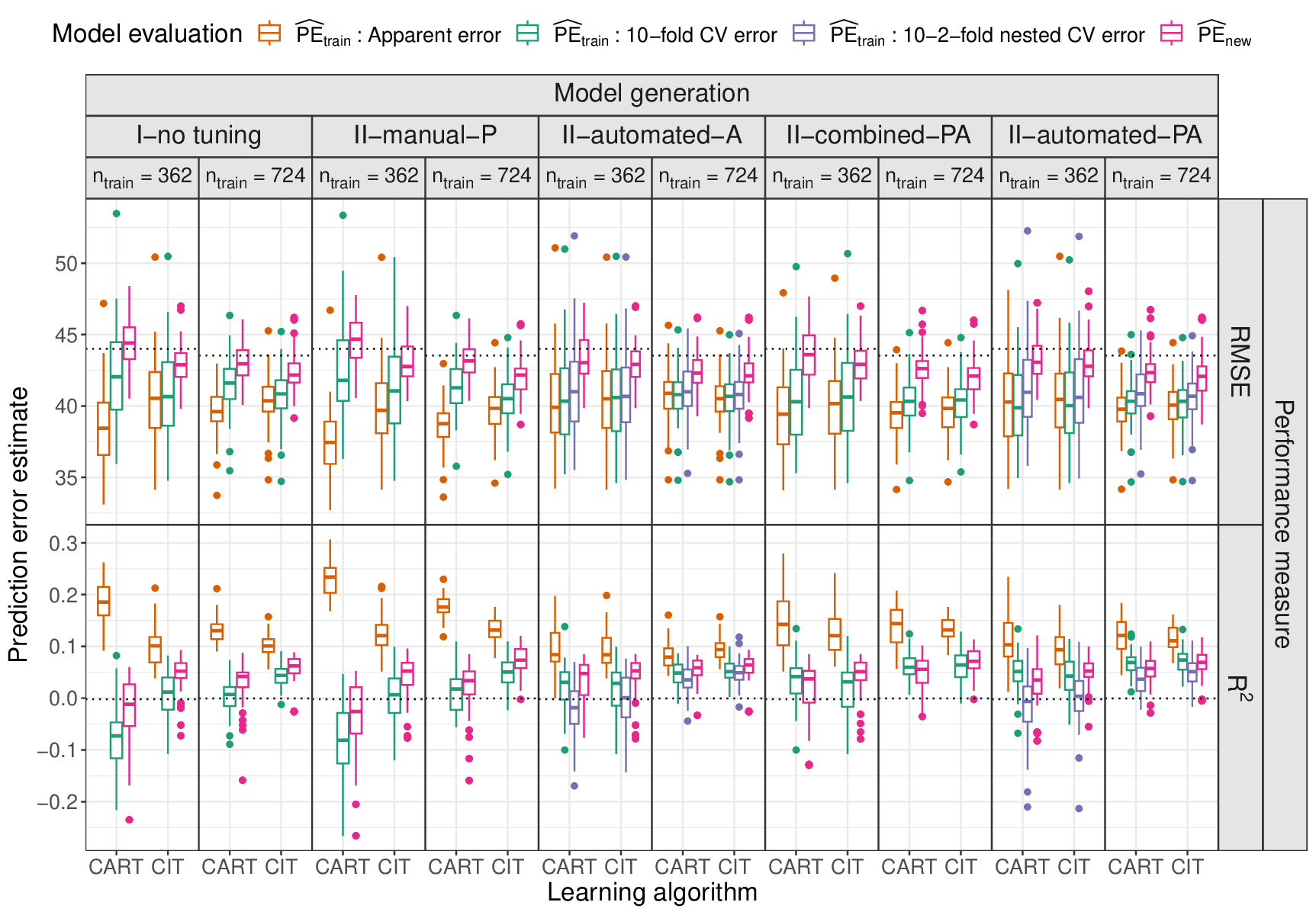}
    \caption{
Absolute prediction error estimates $\ped$ across 96 analysis settings, with each boxplot summarizing 50 repetitions of a specific setting. Additionally, absolute prediction error estimates $\pednew$ are shown. Importantly, $\pednew$ is independent of the model evaluation procedure performed on $\dtrain$ and is therefore shown only for the 40 settings formed by all possible combinations of model generation procedures, performance measures, sample sizes, and learning algorithms ($5 \times 2 \times 2 \times 2 = 40$), where each boxplot again represents 50 repetitions. For reference, the dotted line represents the median prediction error estimate on $\dnew$ (averaged over the 50 repetitions) for a featureless learning algorithm, which naively predicts the mean. Taking the difference between $\ped$ and $\pednew$ for each repetition results in Figure~5 in the main text.}
\label{fig:results_b_naive}
\end{figure}

\begin{figure}
    \centering
    \includegraphics[width=1\linewidth]{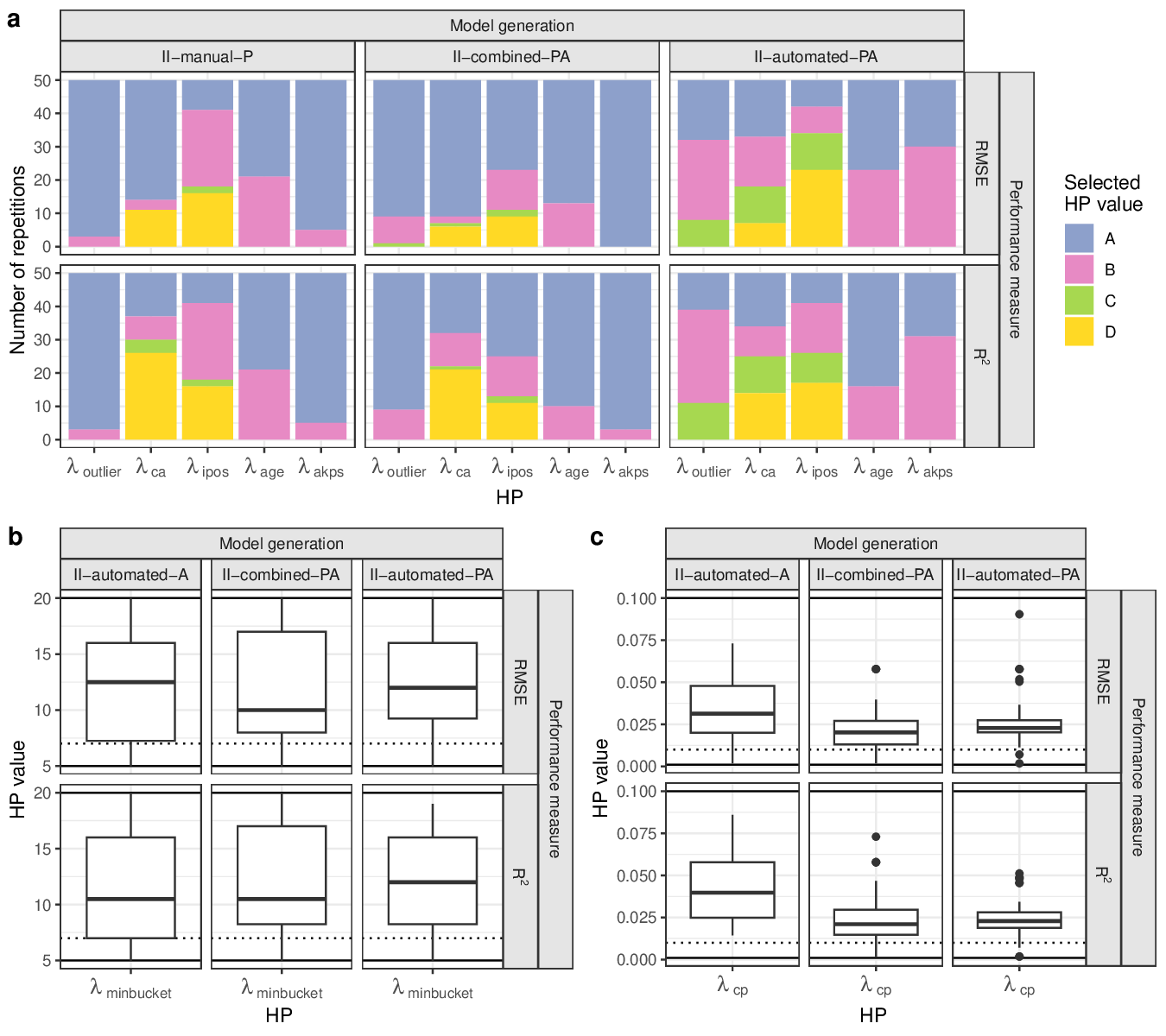}
    \caption{Selected HPs for the analysis settings where CART is used as the learning algorithm and $n_{\text{train}} = 362$. Only model generation procedures that involve tuning the corresponding HP type are shown. a: Preprocessing HPs. The labels A, B, C, and D correspond to the first, second, and, if present, subsequent values in the corresponding search space (with A being the default value). b and c: Algorithm HPs. Each boxplot represents 50 repetitions. The solid and dashed lines indicate the range of the considered search space and the default value, respectively.}
    \label{fig:results_c_cart_nsmall}
\end{figure}
\begin{figure}
    \centering
    \includegraphics[width=1\linewidth]{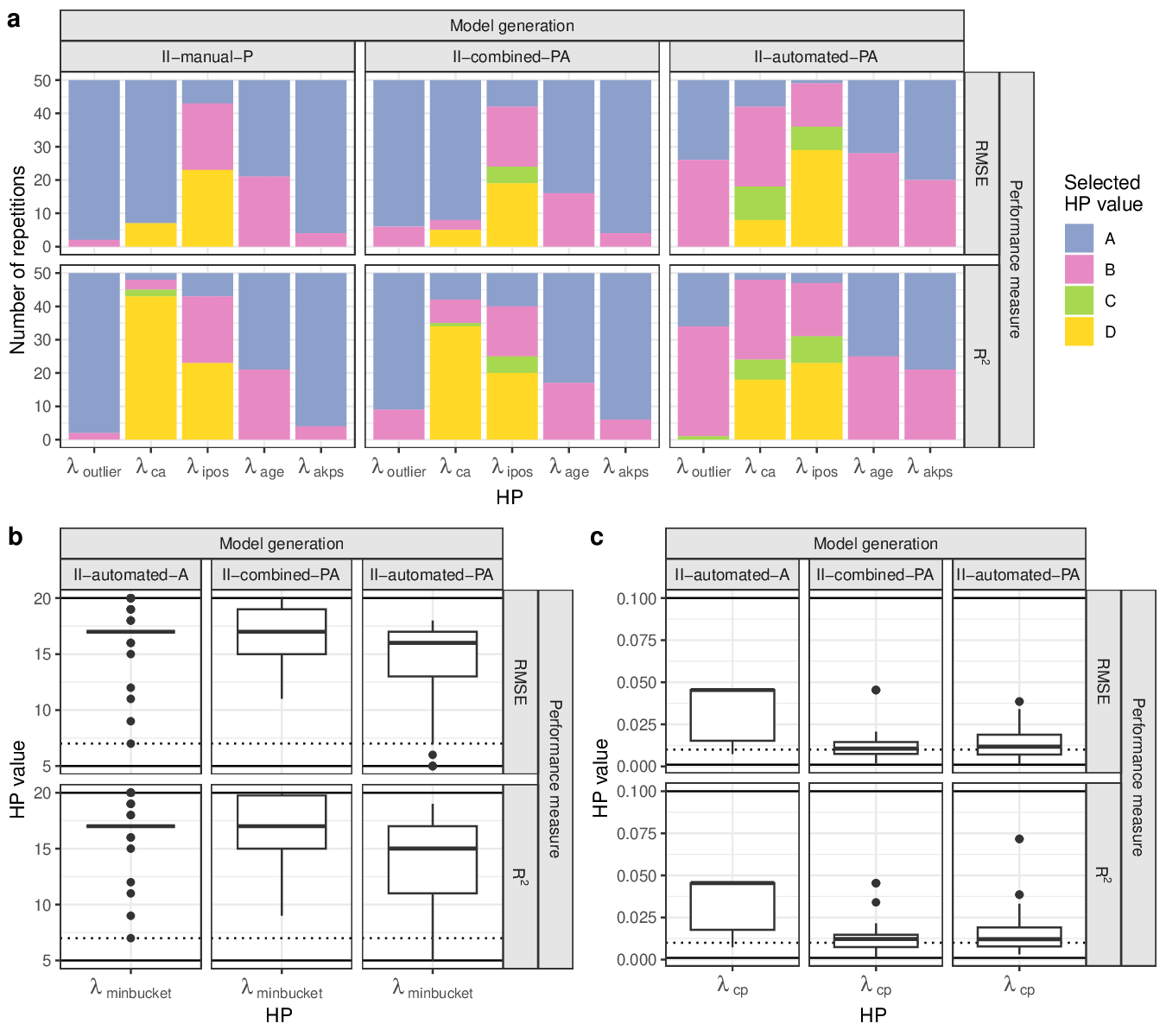}
    \caption{Selected HPs for the analysis settings where CART is used as the learning algorithm and $n_{\text{train}} = 724$. Only model generation procedures that involve tuning the corresponding HP type are shown. a: Preprocessing HPs. The labels A, B, C, and D correspond to the first, second, and, if present, subsequent values in the corresponding search space (with A being the default value). b and c: Algorithm HPs. Each boxplot represents 50 repetitions. The solid and dashed lines indicate the range of the considered search space and the default value, respectively.}
    \label{fig:results_c_cart_nlarge}
\end{figure}
\begin{figure}
    \centering
    \includegraphics[width=1\linewidth]{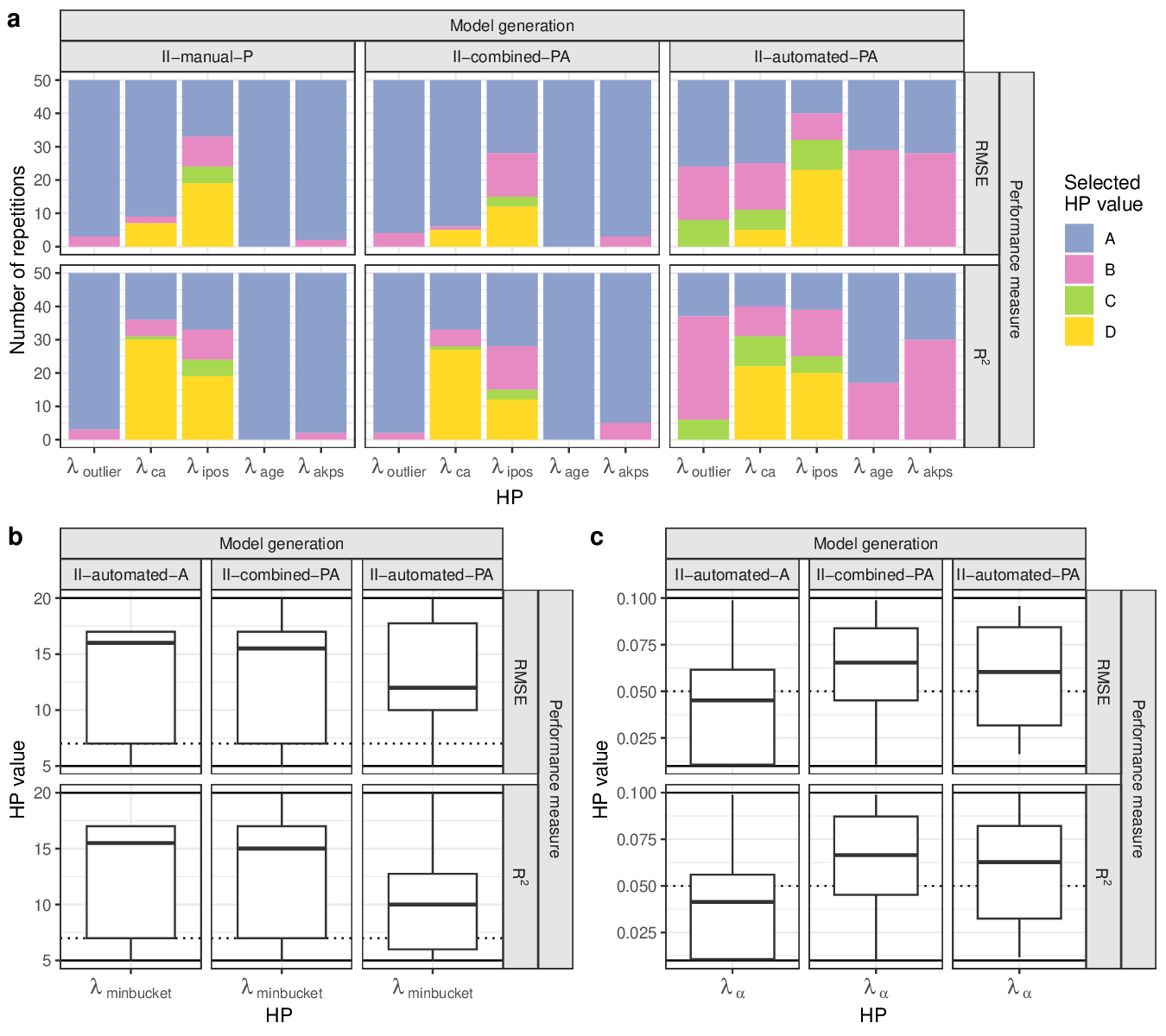}
    \caption{Selected HPs for the analysis settings where CIT is used as the learning algorithm and $n_{\text{train}} = 362$. Only model generation procedures that involve tuning the corresponding HP type are shown. a: Preprocessing HPs. The labels A, B, C, and D correspond to the first, second, and, if present, subsequent values in the corresponding search space (with A being the default value). b and c: Algorithm HPs. Each boxplot represents 50 repetitions. The solid and dashed lines indicate the range of the considered search space and the default value, respectively.}
    \label{fig:results_c_cit_nsmall}
\end{figure}
\begin{figure}
    \centering
    \includegraphics[width=1\linewidth]{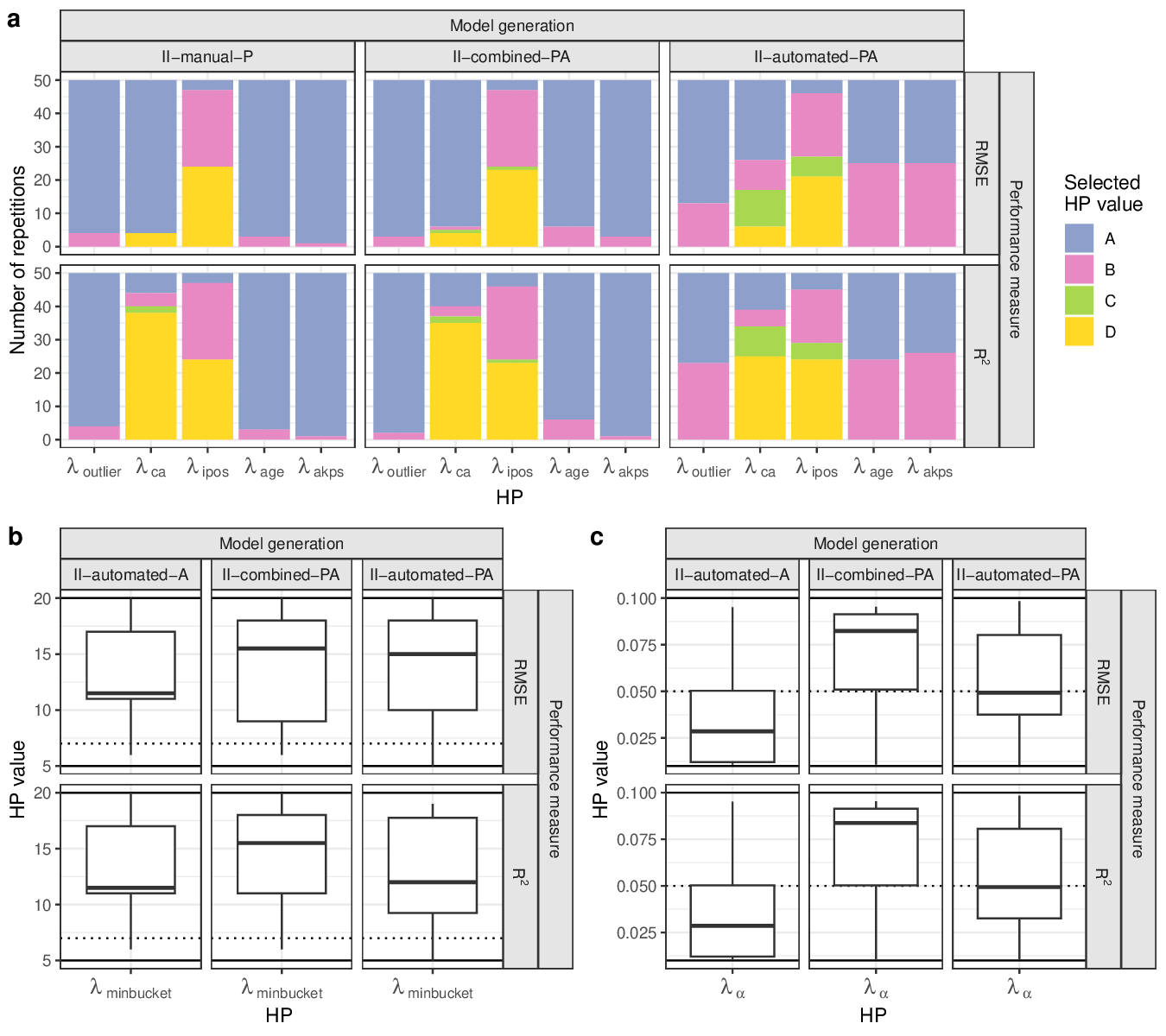}
    \caption{Selected HPs for the analysis settings where CIT is used as the learning algorithm and $n_{\text{train}} = 724$. Only model generation procedures that involve tuning the corresponding HP type are shown. a: Preprocessing HPs. The labels A, B, C, and D correspond to the first, second, and, if present, subsequent values in the corresponding search space (with A being the default value). b and c: Algorithm HPs. Each boxplot represents 50 repetitions. The solid and dashed lines indicate the range of the considered search space and the default value, respectively.}
    \label{fig:results_c_cit_nlarge}
\end{figure}

\clearpage
\subsection{Clustering structure}\label{sec:supp_cluster}
In Figure~5 (Section~5.3), which presents the prediction error differences for 96 analysis settings, it can be seen that the CV error unexpectedly exhibits an optimistic bias in settings without HP tuning. The same observation applies to the nested CV error in analysis settings with HP tuning. These results can be attributed to the clustering structure of the COMPANION data set, and we will explain this in more detail below. Specifically, we will describe the clustering structure (Supplementary Section \ref{sec:supp_cluster1}), explain how it impacts the estimated prediction errors (Supplementary Section \ref{sec:supp_cluster2}), discuss why the experimental setup was not adapted to account for this clustering (Supplementary Section \ref{sec:supp_cluster3}), and present an additional extension of the experimental setup with respect to clustering (Supplementary Section~\ref{sec:supp_cluster4}).
\subsubsection{Clustering in the COMPANION data set}\label{sec:supp_cluster1}
The COMPANION data set exhibits a nested clustering structure. At the first level, clustering arises because several palliative care phases may originate from the same episode of care of a patient. An episode of care is defined as the period between admission to a specific specialist palliative care setting and the termination of care in that same setting. At the second level, clustering occurs because the episodes of care in the data were collected from different palliative care teams. Episodes within the same team are typically more similar to one another than to episodes from different teams. Since no episode of care is associated with more than one palliative care team, the clustering follows a nested structure.\\
As a result, the 1{,}449 palliative care phases reported for the COMPANION data set in Section~5.1 originate from 705 episodes of care, which in turn are collected from 9 specialist palliative home care teams. 
\begin{figure}[!htpb]
    \centering
    \includegraphics[width=0.8\linewidth]{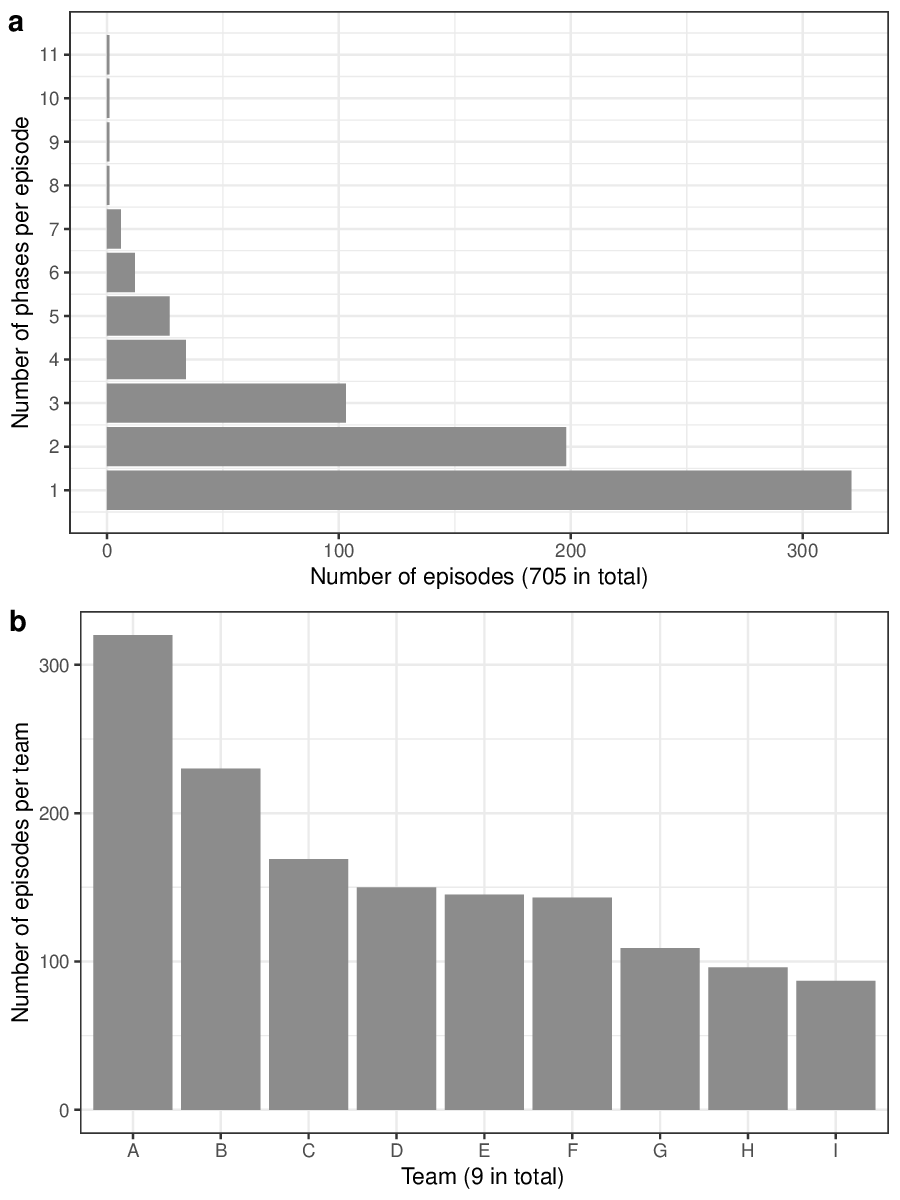}
    \caption{Overview of the nested clustering structure in the COMPANION data set. The x-axis represents the clusters, and the y-axis indicates the cluster size. a: Phases within episodes (first-level clustering). b: Episodes within teams (second-level clustering). The labeling of the teams (A, B, C, etc.) is specific to this plot and reflects the teams' ordering based on the number of episodes, with `A' representing the team with the most episodes.}
    \label{fig:cluster}
\end{figure}
A more detailed depiction of this nested clustering structure is provided in Figure~\ref{fig:cluster}.

\subsubsection{Impact on prediction error estimates}\label{sec:supp_cluster2}
While our empirical illustration and the paper as a whole focus on overlap-induced data leakage, the clustering structure of the COMPANION data set introduces another form of leakage that generally occurs when the assumption of independent and identically distributed (i.i.d.) observations is violated and the violation is not accounted for during model evaluation. This type of leakage is briefly mentioned in Section~2.4.2 of the main paper and described in more detail in Supplementary Section~\ref{sec:supp_leakage_iid}. As a result, the prediction error estimates can be optimistically biased, even in the absence of overlap-induced data leakage. We now explain where the clustering is not accounted for in the experimental setup and how this affects the estimated prediction errors and their differences.\\
First, the clustering structure is ignored when splitting the COMPANION data set into $\dtrain$ and $\dnew$, as the split is performed at the phase level rather than at the episode or team level. Consequently, if the prediction model is intended to be applied to new episodes and teams not present in the COMPANION data set, $\pednew$ is optimistically biased, as it has an unfair advantage compared to other data sets with new episodes and teams.  A more precise statement in step (iii) in Section~5.2.1 would thus be that $\pednew$ is unbiased except for a potential optimistic bias caused by clustering-induced data leakage. Second, if $\ped$ is estimated via simple or nested CV, the clustering structure is also ignored when creating the CV splits. Accordingly, as with $\pednew$, this leads to an optimistic bias in $\ped$ due to data leakage induced by clustering (although in contrast to $\pednew$, $\ped$ may also be affected by other biases). Note that for nested CV, it is only the ignoring of the clustering in the outer CV loop that results in the optimistic bias, as the inner splits are only used for tuning. \\
For the difference between $\ped$ and $\pednew$, which is the focus of our illustration, this has two key implications:
If $\ped$ results from an analysis setting where the apparent error was used to evaluate the final prediction model, the difference between $\ped$ and $\pednew$ may underestimate the optimistic bias that would arise if $\dnew$ contained exclusively observations from new episodes and teams not present in $\dtrain$. 
If $\ped$ corresponds to the simple or nested CV error, the clustering-induced optimistic bias would, under the assumption that $\ped$ and $\pednew$ are subject to the same level of bias, effectively cancel out when considering the difference between $\ped$ and $\pednew$. However, as shown in Figure~5, this is not the case. 
Further analysis (not shown) reveals that the observed differences arise from the slightly higher proportion of patient episodes present in both $\dtraine$ and $\dtest$ during resampling, compared to the proportion of episodes present in both $\dtrain$ and $\dnew$ during the initial split. As a result, $\ped$ is affected by a larger optimistic bias than $\pednew$, which manifests in Figure~5, where their difference is examined.

\subsubsection{Splits on cluster level}\label{sec:supp_cluster3}
\begin{figure}[t]
    \centering
    \includegraphics[width=1\linewidth]{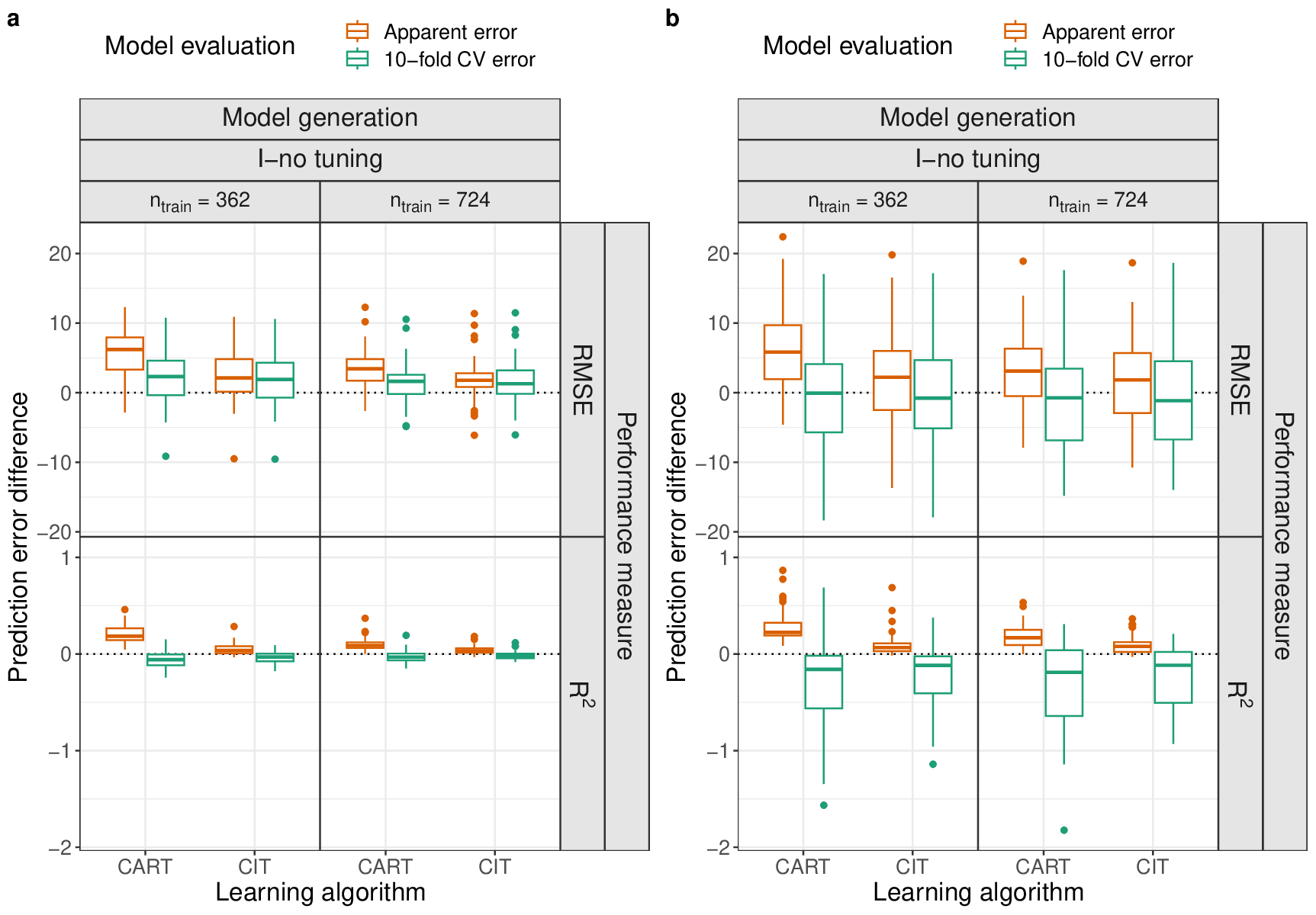}
    \caption{Comparison of prediction error differences when clustering is ignored vs.\ accounted for. Both subfigures present the prediction error differences for all considered analysis settings without HP tuning, with each boxplot summarizing 50 repetitions of a specific setting.  The prediction error differences are calculated as $\pednew - \ped$ for RMSE and $\ped - \pednew$ for $R^2$. a: Naive setup, where clustering is ignored during splitting. Results are adapted from Figure~5, with extended y-axis limits. b: Cluster setup, where clustering is accounted for by performing splits at the team level.}
    \label{fig:results_a_teams}
\end{figure}

To prevent data leakage due to clustering, both the initial split into $\dtrain$ and $\dnew$, as well as any resampling method applied to $\dtrain$, must be performed at the team level. With a total of 9 teams, this means that in each repetition of every analysis setting, $\dtrain$ consists of either 4 or 5 teams. Furthermore, when performing CV on $\dtrain$ at the team level, it is not possible to create 10 folds. Instead, each team forms a fold, and CV is carried out in a leave-one-out manner. Figure~\ref{fig:results_a_teams} presents the resulting prediction error differences for all analysis settings where no HPs are tuned, alongside the corresponding results from the original setup with naive splits (i.e.\ splits that ignore clustering) for comparison. 
First, it can be observed that if $\ped$ corresponds to the CV error, the differences are smaller than or equal to zero for RMSE. This confirms that the optimistic bias found for the CV error in the corresponding naive setup is caused by the clustering structure of the data. However, Figure~\ref{fig:results_a_teams}b also reveals that performing CV at the team level leads to highly variable prediction error differences, which is not surprising given the limited number of teams, each varying in the number of episodes and phases they contain. Since we argue that, under these circumstances, it is not reasonable to perform HP tuning, we decided to ignore the clustering structure in the setup of our main analysis. Additionally, in the interest of computational resources, we did not conduct the team-level analysis for the remaining analysis settings involving tuning. However, this should clearly not be taken as a standard for applications beyond illustrative purposes. 

\subsubsection{Learning algorithms for clustered data}\label{sec:supp_cluster4}
\begin{figure}
    \centering
    \includegraphics[width=0.88\linewidth]{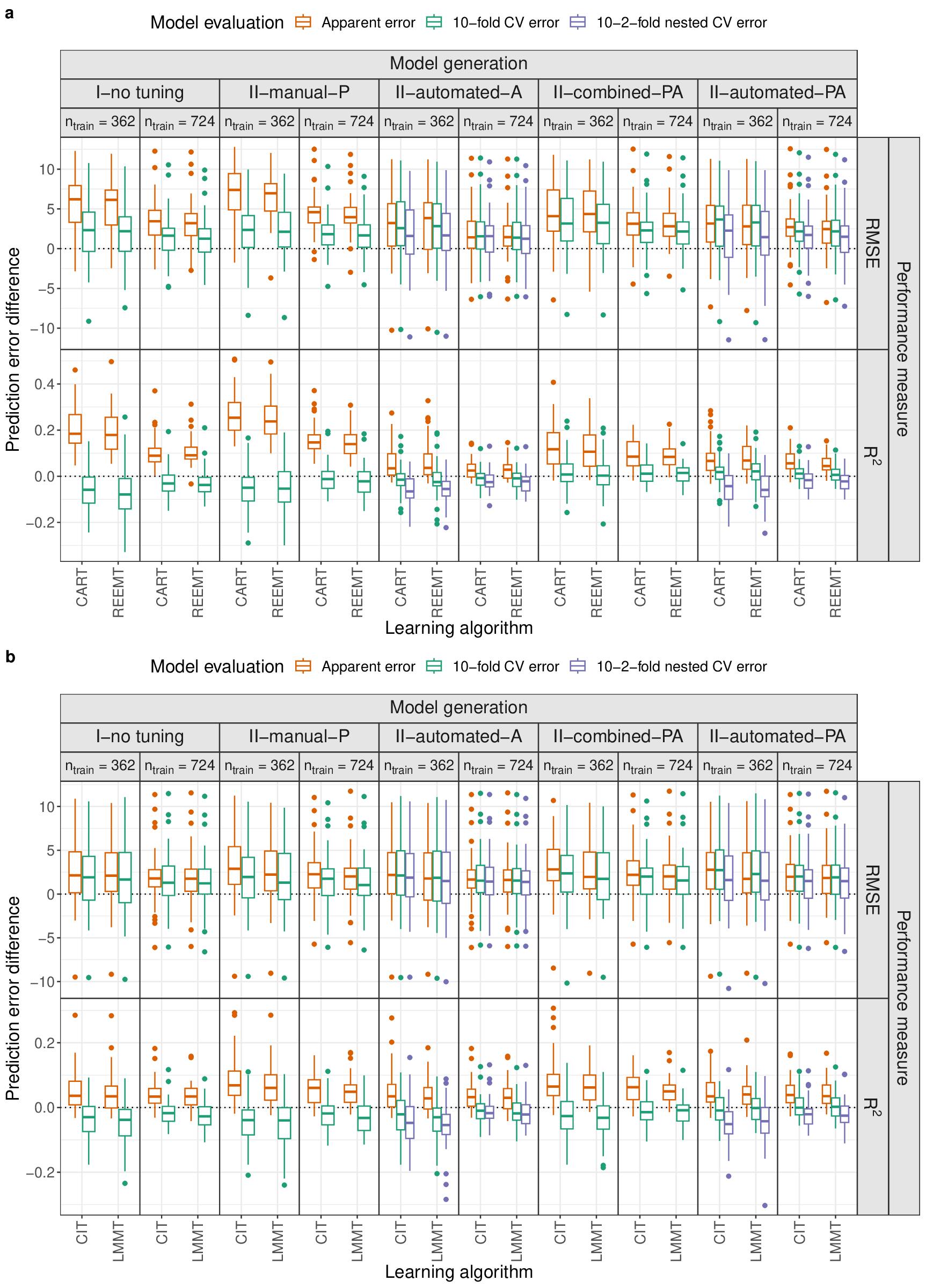}
    \caption{Comparison of prediction error differences between CART and CIT and their counterparts that include random intercepts, REEMT and LMMT, respectively. The same model generation and evaluation procedures, performance measures, and sample sizes as in the main setup are included. Each boxplot summarizes results from 50 repetitions of a specific setting. The prediction error differences are calculated as $\pednew - \ped$ for RMSE and $\ped - \pednew$ for $R^2$. a: CART vs. REEMT. b: CIT vs. LMMT.}
    \label{fig:results_a_re}
\end{figure}

In addition to performing splits at the cluster level, we also extended the main experimental setup by including additional learning algorithms specifically designed for clustered data. These are the Random Effects/Expectation-Maximization Tree algorithm (REEMT; \texttt{R} package \texttt{REEMtree}; \citealp{Sela2011}), and the Linear Mixed-Effects Model Tree algorithm (LMMT; \texttt{R} package \texttt{glmertree}; \citealp{Fokkema2018}). In the implementation used for our illustration, both algorithms take into account the clustering structure by iterating between two steps: (i) fitting a decision tree using the CART algorithm for REEMT or the CIT algorithm for LMMT and (ii) estimating random intercepts via a linear mixed model, which are subtracted from the outcome variable in the subsequent tree-fitting iteration. To ensure model stability, random effects are only included for each palliative care team, rather than for each individual episode, as more than 300 episodes consist of only a single palliative care phase (Figure~\ref{fig:cluster}a). Including REEMT and LMMT in the analysis, however, does not yield new insights. Their results closely resemble those of CART and CIT, as demonstrated in Figure~\ref{fig:results_a_re}, which compares the prediction error differences of the algorithms.

\end{document}